\documentclass[journal]{IEEEtran}

\usepackage{amsmath,amsfonts}
\usepackage{algorithmic}
\usepackage{algorithm}
\usepackage{array}
\usepackage{textcomp}
\usepackage{stfloats}
\usepackage{url}
\usepackage{verbatim}
\usepackage{graphicx}
\usepackage{cite}

\usepackage{caption}
\usepackage{subcaption}
\usepackage{makecell}
\usepackage{tablefootnote}

\usepackage[pagebackref=false,breaklinks=true,letterpaper=true,colorlinks,citecolor=black,linkcolor=black,bookmarks=false]{hyperref}
\usepackage{amsmath}
\usepackage{amssymb}
\usepackage{color}
\usepackage{multirow}

\def\eg{\emph{e.g.}}

\def\etal{{\em et al.~}}

\newcommand{\figref}[1]{Fig.~\ref{#1}}
\newcommand{\tabref}[1]{Table~\ref{#1}}

\begin{document}

\title{AdapterShadow: Adapting Segment Anything Model for Shadow Detection}

\author{
Leiping Jie, and Hui Zhang, \textit{Senior Member}\\
\IEEEcompsocitemizethanks{
\IEEEcompsocthanksitem Leiping Jie is with the Department of Computer Science, Hong Kong Baptist University, Hong Kong, China (17482305@life.hkbu.edu.hk).
\IEEEcompsocthanksitem Hui Zhang is with the Department of Computer Science and Technology, BNU-HKBU United International College, Zhuhai, China (amyzhang@uic.edu.cn).
}
}

\markboth{Journal of \LaTeX\ Class Files,~Vol.~14, No.~8, August~2021}%
{Shell \MakeLowercase{\textit{et al.}}: A Sample Article Using IEEEtran.cls for IEEE Journals}


\IEEEpubid{
\begin{minipage}{\textwidth} \ \\[30pt]
\centering
1520-9210 \copyright 2023 IEEE. Personal use is permitted, but republication/redistribution requires IEEE permission. \\
See https://www.ieee.org/publications/rights/index.html for more information.
\end{minipage}
}


\maketitle

\begin{abstract}
Segment anything model (SAM) has shown its spectacular performance in segmenting universal objects, especially when elaborate prompts are provided. However, the drawback of SAM is twofold. On the first hand, it fails to segment specific targets, \eg, shadow images or lesions in medical images. On the other hand, manually specifying prompts is extremely time-consuming. To overcome the problems, we propose AdapterShadow, which adapts SAM model for shadow detection. To adapt SAM for shadow images, trainable adapters are inserted into the frozen image encoder of SAM, since the training of the full SAM model is both time and memory consuming. Moreover, we introduce a novel grid sampling method to generate dense point prompts, which helps to automatically segment shadows without any manual interventions. Extensive experiments are conducted on four widely used benchmark datasets to demonstrate the superior performance of our proposed method. Codes will are publicly available at \url{https://github.com/LeipingJie/AdapterShadow}.

\end{abstract}


\IEEEpeerreviewmaketitle 

\section{Introduction}
\IEEEPARstart{S}{hadows} are formed when light is blocked by objects. Although shadows reflect the structure of 3D world, they bring troubles to many computer vision tasks, \eg, object tracking and detection. However, high performance shadow detection is a challenging problem, since shadows can appear anywhere in any shape, size, and intensity. 

\begin{figure}[tbh!]
  \begin{center} 
  \includegraphics[width=0.45\textwidth]{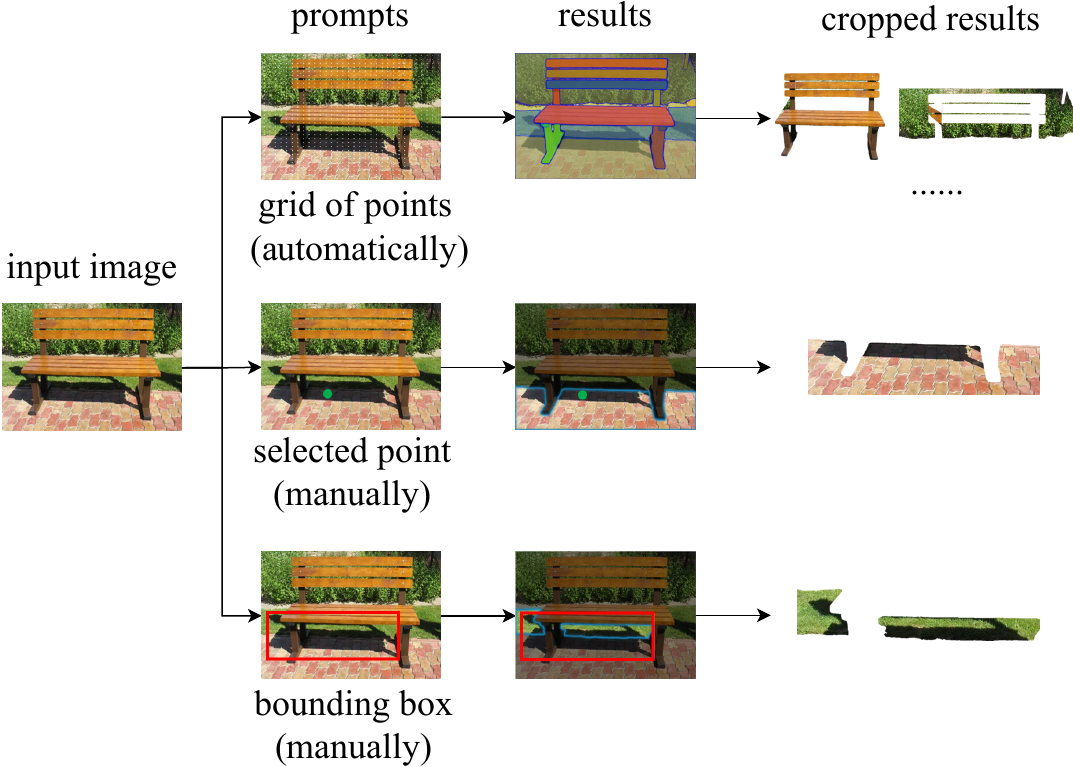}
  \end{center}
  \caption{Illustration of two different kinds of prompts (point, bounding box) that are supported by SAM project (\url{https://segment-anything.com/}). Specifically, the first row shows the automatic way of generating grid of points, which will produce multiple segmentation results. The second and the third row indicate the results of manually specifying a single point prompt or a bounding box. As shown, they all fail to segment the shadow regions.}
  \label{fig_SMLFA}
\end{figure}

Earlier methods~\cite{finlayson2009entropy, huang2009moving} designed hand-craft shadow features to train machine learning models. Despite the progress made, hand-craft features are still fragile at detecting truly complex shadows. Like almost every other computer vision task, shadow detection is now dominated by deep neural networks (DNN). Current shadow detection approaches are built using carefully designed DNNs and trained with annotated benchmark datasets~\cite{sbu_dataset, ucf_dataset, istd_dataset, gy_tip2021}, resulting in excellent performance. Nevertheless, the scales of the annotated datasets are relatively too small compared to those of other tasks, such as ImageNet~\cite{deng2009imagenet} for image classification. Specifically, ImageNet has over 14 million images while the largest shadow detection dataset, the CUHK dataset~\cite{gy_tip2021}, has only 10K images. Constructing large-scale shadow detection datasets may be a promising way to improve the performance of shadow detection. However, it is quite expensive and time-consuming, especially for those complex and soft shadows.

Recently, large language models (LLMs) have become increasingly popular, especially after the popular ChatGPT demonstrated powerful performance. Meanwhile, large-scale vision-based models such as the segment anything model (SAM) for image segmentation have also been proposed. SAM is trained on $11$ million images with over $1B$ masks, and shows appealing performance. A question then is: \textit{Can we leverage the powerful feature representation learned in SAM to improve the performance of shadow detection?} Direct use of SAM for shadow detection leads to unsatisfactory results~\cite{jie2023sam}, which has also been reported in other articles for tasks such as medical images~\cite{he2023accuracy}. Given the large model size of SAM, fine-tuning it is unaffordable for most modern GPUs. In this paper, we investigate the possibility of leveraging SAM for shadow detection via adapters. Specifically, the parameters in the image encoder and prompt encoder are frozen, while those of the adapter inserted into the image encoder and the mask decoder are trainable. This has two advantages. First, the number of trainable parameters is relatively small, which means that high GPU memory consumption is not required for training. Secondly, due to the smaller scale of existing shadow benchmarks, fewer parameters can effectively avoid overfitting and match the scale of training data.

Currently, SAM supports two sets of prompts: sparse (points, boxes, and text) and dense (masks). In general, designating high-quality prompts will produce better results than using the automatically generated prompts provided by SAM, such as dense grid points. But what's Worse is that using grid points as prompts produces multiple outputs, which requires additional strategies~\cite{jie2023sam} to select suitable candidate points and is therefore not suitable for shadow detection. Therefore, another question is: \textit{Can we automatically generate high-quality prompts? } To this end, we introduce an auxiliary network that first generates a coarse shadow mask and then selects points from it as point prompts for SAM. In this way, our method can generate high-quality point prompts without manually specifying them. We devised several different methods for selecting point prompts and choosing the best design in our experiments.

In summary, the contributions can be summarized as follows:
\begin{itemize}
    \item We introduce the first SAM-based shadow detection method through the use of adapters.
    \item We propose a novel sampling method which can produces high-quality prompt points.
    \item We conduct extensive experiments on popular benchmarks to demonstrate the superior performance of our network.
\end{itemize}

\IEEEpubidadjcol
\section{Related Work}
\subsection{Traditional Shadow Detection Methods}
Early straightforward approaches used color information to identify shadow areas, based on the assumption that shadow areas are usually visually darker and the chromaticity remains unchanged. Color spaces, such as HSV~\cite{cucchiara2003detecting}, YUV~\cite{chen2010enhanced}, were introduced to detect shadows. Later, physical-based color features~\cite{finlayson2009entropy, huang2009moving}, shadow edge features~\cite{huang2011characterizes}, and combination of various features~\cite{martel2005moving, zhu2010learning} were designed and learned using statistical machine learning algorithms, \eg, Gaussian Mixture Model (GMM)~\cite{martel2005moving}, Conditional Random Filed (CRF)~\cite{lalonde2010detecting}, Decision Tress(DT)~\cite{zhu2010learning} or Support Vector Machine(SVM)~\cite{huang2011characterizes}. The biggest issue with traditional methods is their poor robustness. Handcrafted features are unable to differentiate between shadow-like objects and shadows cast on complex backgrounds.

\subsection{Deep Shadow Detection Methods}
Along with the rapid development of deep learning, various shadow detection methods based on deep neural networks have been proposed~\cite{khan14_cvpr, shen16_cvpr, chen_cvpr2021, Lu_2022_CVPR, ding_eccv2022, Wang_2020_CVPR, wang_tpami2022}. Hu~\etal~\cite{hu18_cvpr, hu18_tpami} proposed to learn direction-aware context features by using spatial recurrent neural network (RNN). Zhu \etal~\cite{zhu18_eccv} introduced a bidirectional feature pyramid network which can capture both local and global information. Zheng~\etal~\cite{zheng19_cvpr} designed a differentiable distraction-aware shadow (DS) module to extract and integrate the semantics of visual distraction regions. Considering that collecting and annotating large-scale datasets is expensive and time-consuming, Chen~\etal~\cite{chen20_cvpr} presented a multitask mean teacher model that was trained using unlabeled data in a semi-supervised manner. Meanwhile, Naoto also~\etal ~\cite{naoto20_tcsvt} showed the possibility of detecting shadows by training on synthetic data. However, the performance remains unsatisfactory due to the inherent domain gap between the synthetic and the real data. More recently, ViT~\cite{dosovitskiy20_iclr} based shadow detection methods have also been proposed. Jie~\etal~\cite{jie2022icassp, jie2022icme, jie2023rmlanet} introduced an optimized transformer-based network for shadow detection. In addition, Liao~\etal~\cite{liao2021mm} leveraged ensemble learning of multiple existing methods using predicted depth maps. Deep learning shadow detection methods have achieved promising results on multiple benchmark datasets~\cite{sbu_dataset, ucf_dataset, istd_dataset, gy_tip2021}. However, how shadow detection benefits from large underlying vision models has not yet been studied.

\subsection{Segment Anything Model} 
Meta AI's Segment Anything Model (SAM)~\cite{kirillov2023segment} was introduced, aiming at generic image segmentation. Since it was released, many researchers have been conducted based on it. Pioneering works directly evaluated its ability on other specific tasks, \eg, camouflaged object detection~\cite{tang2023can, ji2023sam, chen2023sam}, medical image segmentation~\cite{hu2023sam, ma2023segment,he2023accuracy,cheng2023sam}, mirror and transparent objects segmentation~\cite{han2023segment}. However, SAM does not perform well, as the original SAM was trained on large scale natural images that show a significant domain gap with these specific images. In particular, Jie \etal~\cite{jie2023sam} performed evaluations on three shadow detection datasets but got worse results. They found that the grid point prompts used to automatically generate shadow masks usually produce multiple predictions. Although they used the Max F-measure or Max IOU strategy to select the best candidate from the predictions, the results were still not satisfactory. Later, different adapters were introduced to fine-tune the SAM model. Chen \etal~\cite{chen2023sam} proposed to insert adapters before and after the transformer blocks of the SAM image encoder and fine-tuned the SAM mask decoder, with other components frozen. In contrast, Zhang \etal~\cite{zhang2023customized} inserted the proposed adapters in the query and value brunches of the transformer blocks of the SAM image encoder. Dai \etal~\cite{dai2023samaug} utilized the SAM model to generate coarse prediction, where augmented points were chosen by four strategies: random point, max entropy point, max distance point and saliency point. Their method required to run the SAM model twice and failed when the coarse predictions were dissatisfied. Despite the boosting performance achieved, how to automatically generate prompts is still a question. In~\cite{wu2023medical}, the point prompt was produced by selecting a random point within the ground truth mask region, which is obviously unavailable when inference with images without ground truths provided. Towards it, Shaharabany \etal~\cite{shaharabany2023autosam} froze the whole SAM model and proposed a prompt generator network which generated encoded features to feed the SAM mask decoder. In contrast, our proposed prompt generator produces point prompts from the predicted coarse mask instead of directly using the encoded features. This design allows us to impose explicit supervision to the predicted coarse mask, thereby ensuring the high-quality of the generated point prompts.
\section{Methodology}
\subsection{Recap Segment Anything Model}
  SAM~\cite{kirillov2023segment} is a VIT-based model, which consists of three parts, an image encoder, a flexible prompt encoder and a lightweight mask decoder.
  \begin{itemize}
      \item \textbf{Image Encoder} leverages an MAE~\cite{he2022masked} pre-trained  ViT~\cite{dosovitskiy20_iclr} to handle high resolution input images.
      \item \textbf{Prompt Encoder} supports two sets of prompts: sparse (points, boxes, text) and dense (masks). Specifically, points and boxes are decoded by the summation of the positional embeddings and the learnable embeddings. Text and dense prompts are represented by the CLIP~\cite{clip2021icml}'s text encoder, and the summation of image embedding and the output of a series of convolution layers, respectively.
      \item \textbf{Mask Decoder} combines the image embedding and the prompt embedding to obtain the desired masks.
  \end{itemize}
  
\begin{figure*}[tbh!]
  \begin{center} 
  \includegraphics[width=0.9\textwidth]{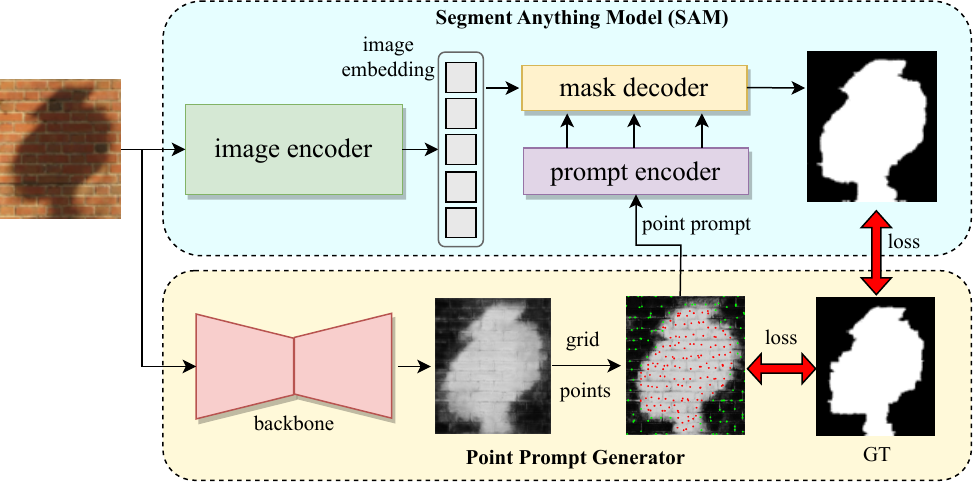}
  \end{center}
  \caption{Illustration of two different kinds of prompts (point, bounding box) that are supported by SAM. Specifically, the first row shows the automatic way of generating even grid points, which will produce multiple segmentation results. The second and the third row indicate the results of manually specifying a single point prompt or a bounding box. As shown, they all fail to segment the shadow regions.}
  \label{fig_architecture}
\end{figure*}

It is worth mentioning that the dataset used in SAM: \textbf{SA-1B} are collected from three stages: assisted-manual, semi-automatic and fully automatic stage, which results in a total of $11M$ diverse, high-resolution, licensed, and privacy protecting images and $1.1B$ high-quality segmentation masks. SA-1B is the largest segmentation dataset and has $11\times$ more images and $400\times$ more masks than the largest existing segmentation dataset Open Images~\cite{kuznetsova2020open}.

If no prompts are provided, the official SAM demo will prompt the SAM with each point from the automatically generated dense grid, determine the final points by de-duplicating predicted masks, and get all the predicted masks. 

SAM does achieve a significant performance on various segmentation benchmarks, especially its remarkable zero-shot transfer capabilities on 23 diverse segmentation datasets. However, researchers also reported its deficiencies on several tasks, such as medical image segmentation~\cite{ma2023segment, ji2023sam}, camouflaged object detection~\cite{tang2023can, ji2023sam}. It seems that the desired generic object segmentation model is still not yet implemented.

\subsection{Overview of Our Method}
As shown in Fig.~\ref{fig_architecture}, our proposed network consists of two parts: the adapted Segment Anything Model and the Point Prompt Generation module. Specifically, we add efficient and lightweight Adapters in the transform blocks inside the image encoder. During training, except for the added Adapters and the mask decoder part, the remaining parameters in SAM are all fixed. In the point prompt generator, we propose a novel grid sampling method to select high-quality point prompts.

\begin{figure}[tbh!]
  \begin{center} 
  \includegraphics[width=0.4\textwidth]{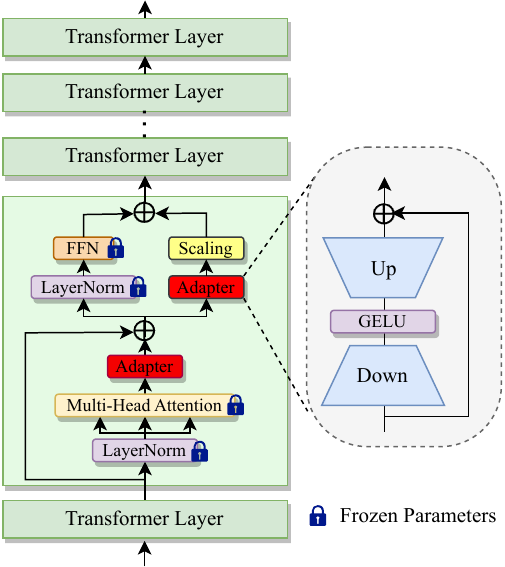}
  \end{center}
  \caption{Illustration of our modified transformer layer in the image encoder of SAM. The first adapter is added after the multi-head attention block, while the other scaled adapter replaces the shortcut connection of the feedforward block. The adapter only consists of a GELU activation function and two MLPs. Note that only the adapters are trainable while others are froze. }
  \label{fig_image_encoder}
\end{figure}

\subsection{Transformer Adapter}
Roughly, Transformer layers include two key components: the multi-head self-attention and the feed forward network. Give the input embeddings $E_i$, the corresponding attention output $E_{att}^i$ and final output $E_{out}^i$ can be obtained by:
\begin{equation}
  \begin{aligned}
    E_{att}^i &= MHA(LN(E_i)), \\
    E_{out}^i &= LN(FFN(E_{att}^i)),
  \end{aligned}
\end{equation}
where $LN$, $MHA$ and $FFN$ are LayerNorm~\cite{ba2016layer}, Multi-head Attention and Feed Forward Network, respectively. To minimize the changes to the original image encoder in SAM, we plug simple trainable Adapters in both $MHA$ and $FFN$ blocks, as shown in \figref{fig_image_encoder} (red block). Specifically, given the input embedding $E_{in}\in \mathbb{R}^{B\times N\times C}$, our adapter first downsamples the input by MLP with ratio $r$ $(0<r<1)$ to get intermediate embedding $E_{m}\in \mathbb{R}^{B\times N\times rC}$. Then, we activate $E_{m}$ using GELU~\cite{hendrycks2016gaussian} and upsamples the result by MLP with ratio $1/r$ to get the output embedding $E_{out}\in \mathbb{R}^{B\times N\times C}$. In this way, the input and output dimensions keep equal. This procedure can be formulated as:
\begin{equation}
  E_{out} = MLP_{{1}/{r}}(GELU(MLP_{r}(E_{in}))).
\end{equation}
Moreover, we also tried other Adapters that claimed improved performance on NLP tasks, \eg, LoRA~\cite{pfeiffer2020adapterfusion}, AdapterFusion~\cite{pfeiffer2020adapterfusion}. However, no such improvements were observed in our task. Actually, their performance even worse.

\subsection{Point Prompt Generation}
As we mentioned, SAM supports both sparse and dense prompts. Obviously, text prompt is not suitable for our task. Therefore, we investigate sparse sets including points, bounding boxes and masks. To unify these three types, we propose to utilize the predicted coarse shadow mask as a proxy, since both points and a bounding box can be generated from it. Considering that our goal of the point prompt generation is to provide good prompts for SAM, we design an efficient and lightweight decoder. Specifically, we first adopt an off-the-shelf feature extractor (\eg, Efficient-Net~\cite{tan19_icml}, ResNext101~\cite{xie2017aggregated}) as our encoder to extract feature pyramids. Given an input shadow image $I_s\in \mathbb{R}^{h\times w\times 3}$, $N$-level of features $F_e^{i=1\to N}\in \mathbb{R}^{\frac{h}{i}\times \frac{w}{i}\times C_i}$ ($N=5$) are extracted. Note that $C_i$ varies for different backbones, \eg, for Efficient-Net B1, $C_i=[16,24,40,112,1280]$, $C_i=[24,40,64,176,2048]$ for Efficient-Net B5, while $C_i=[64, 256, 512, 1024, 2048]$ for ResNext101. Give the input feature $F_e^i$, we formulate the decoder blocks which produce the decoder features $F_d^i$ as follows:

\begin{equation}
  \begin{aligned}
    F_{u}^{i} &= Upsample(F_{d}^{i}), \\
    F_{c}^{i-1} &= Concat(F_{u}^{i}, F_{e}^{i-1}), \\
    F_{d}^{i-1} &= ConvBlock(F_{c}^{i-1}),
  \end{aligned}
\end{equation}
where $Upsample$, $Concat$, $ConvBlock$ are the differentiable interpolation operator, the concatenation operator along the channel dimension and two $3\times 3$ 2D convolution layers with BatchNorm and LeakyReLU, respectively. $F_{d}^{i}\in \mathbb{R}^{\frac{h}{i}\times \frac{w}{i}\times \frac{r}{N-i+1}}$, $F_{u}^{i}\in \mathbb{R}^{\frac{h}{i-1}\times \frac{w}{i-1}\times \frac{r}{N-i+1}}$, $F_{e}^{i-1}\in \mathbb{R}^{\frac{h}{i-1}\times \frac{w}{i-1}\times C_{i-1}}$, $F_{c}^{i-1}\in \mathbb{R}^{\frac{h}{i-1}\times \frac{w}{i-1}\times (\frac{r}{N-i+1}+C_{i-1})}$, $F_{d}^{i-1}\in \mathbb{R}^{\frac{h}{i-1}\times \frac{w}{i-1}\times \frac{r}{N-i+2}}$. Note that, when $i=N$, $F_{d}^N$ is obtained by applying a $1\times 1$ convolution layer over $F_{e}^N$. To be more effective, the channel dimension of $F_{e}^N$ is set to $r (r\ll C_N)$. For example, when using EfficientNet-B1 as our backbone, $r$ is set to $128$ with $C_5=1280$.

\begin{figure}[tbh!]
  \begin{center} 
  \includegraphics[width=0.5\textwidth]{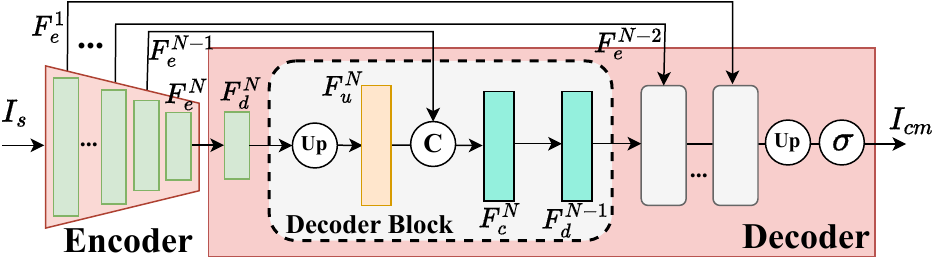}
  \end{center}
  \caption{Illustration of our lightweight network for point prompt generation. $Up$, $C$ and $\sigma$ represents the differentiable upsampling operator, the concatenation operator along the channel dimension and the Sigmoid function, respectively. We adopt EfficientNet~\cite{tan19_icml} as our encoder to extract multi-level features, which are further utilized by the decoder to predict the corresponding coarse shadow mask in a progressive manner.}
  \label{fig_pointprompt}
\end{figure}

Given the predicted coarse shadow mask $I_{cm}$, we can obtain three types of prompts $P_j (j = 0, 1, 2)$ as follows: 
\begin{equation}
  P_j=\left\{
	\begin{aligned}
		&Point\_k(M_c), & j = 0,\\
		&BBox(M_c), & j = 1,\\
		&I_{cm}, & j = 2.
	\end{aligned}
	\right.
\end{equation}
Specifically, $Point\_k$ means choosing $k$ points from $M_c$, while $BBox$ calculates the bounding box of $M_c$. Our method of generating various types of prompts automatically fully aligns with the ways of specifying prompts in original SAM. Empirically, we found that point prompts perform much better than using the bounding box or the coarse shadow mask. Based on this observation, we propose two different strategies of point sampling, as follows:
\begin{equation}
  Point\_k(M_c)=\left\{
	\begin{aligned}
		&Top_k(M_c), \\
		&Grid_{g\times g, k}(M_c),\\
	\end{aligned}
	\right.
\end{equation}
where $Top_k$ means select top $k$ points from the coarse shadow mask as prompts, while $Grid_{g\times g,k}$ represents splitting the coarse shadow mask into $g\times g$ blocks and select $k$ points from each block. For points selected from every single block, we also use top-k algorithms to choose k candidate points. Then, we define a threshold value $\tau$ to classify them into positive or negative point prompts as follows:
\begin{equation}
  Grid^i_{g\times g, k}(M_c)=\left\{
	\begin{aligned}
		&0, & M_c(i)<\tau,  \\
		&1, & M_c(i)\ge\tau, \\
	\end{aligned}
	\right.
\end{equation}
where $M_c(i)$ means the value of the $i$th point in the coarse shadow mask $M_c$.

\subsection{Loss Function}
Considering the number of shadow pixels and the number of non-shadow pixels are imbalanced, we follow~\cite{jie2022icme, jie2023rmlanet} to employ focal loss~\cite{lin2017focal} to compensate the imbalanced distribution and focus more on hard samples. The focal loss for the $i$-th pixel can be formulated as:
\begin{equation}
    \mathcal{L}_{s}^i= \left \{
    \begin{aligned}
      &-\alpha{(1-\hat{y_i})}^\gamma log\hat{y_i},     &y=1 \\
      &-(1-\alpha){\hat{y_i}}^\gamma log(1-\hat{y_i}), &y=0
    \end{aligned}  
    \right.
\end{equation}
where $\alpha$, $\gamma$, $\hat{y_i}$ and $y_i$ are the weight factor of the unbalanced distribution, the tunable focus parameter of the modulating factor ${(1-\hat{y_i})}^\gamma$, the predicted value and the ground truth, respectively. Empirically, we set $\alpha$ to ${8}/{9}$ and $\gamma$ to $2.0$ when training with the SBU and ISTD dataset. While for the CUHK dataset, $\alpha$ becomes $0.7$ and $\gamma$ becomes $2.0$ . 
\section{Experiments}
\label{sec:experiment}

\begin{table*}[htbp!]
    \begin{center}
    \caption{Quantitative comparison with the state-of-the-art methods for shadow detection on the SBU~\cite{sbu_dataset}, UCF~\cite{ucf_dataset} and ISTD~\cite{istd_dataset} benchmark dataset. The best and the second best results are marked in bold and underlined, respectively.}
    \begin{tabular}{|c|ccc|ccc|ccc|}
    \hline 
    & \multicolumn{3}{c|}{SBU~\cite{sbu_dataset}} & \multicolumn{3}{c|}{UCF~\cite{ucf_dataset}} & \multicolumn{3}{c|}{ISTD~\cite{istd_dataset}} \\
    \hline 
    \hline 
    Method & $BER$ $\downarrow$ & BER$_{S}$ $\downarrow$ & BER$_{NS}$ $\downarrow$ & $BER$ $\downarrow$ & BER$_{S}$ $\downarrow$ & BER$_{NS}$ $\downarrow$ & $BER$ $\downarrow$ & BER$_{S}$ $\downarrow$ & BER$_{NS}$ $\downarrow$ \\
    \hline 
    \hline
    Unary-Pairwise~\cite{guo11_cvpr} (2011) & 25.03 & 36.26 & 13.80 & - & - & - & - & - & - \\
    stacked-CNN~\cite{sbu_dataset} (2016) & 11.00 & 8.84 & 12.76 & 13.00 & 9.00 & 17.10 & 8.60 & 7.69 & 9.23 \\
    scGAN~\cite{nguyen17_iccv} (2017) & 9.10 & 8.39 & 9.69 & 11.50 & 7.74 & 15.30 & 4.70 & 3.22 & 6.18 \\
    patched-CNN~\cite{hs18_iros} (2018) & 11.56 & 15.60 & 7.52 & - & - & - & - & - & - \\
    ST-CGAN~\cite{istd_dataset} (2018) & 8.14 & 3.75 & 12.53 & 11.23 & \textbf{4.94} & 17.52 & 3.85 & 2.14 & 5.55 \\
    DSC~\cite{hu18_cvpr} (2018) & 5.59 & 9.76 & \textbf{1.42} & 10.54 & 18.08 & \textbf{3.00} & 3.42 & 3.85 & 3.00 \\
    ADNet~\cite{le18_eccv} (2018) & 5.37 & 4.45 & 6.30 & 9.25 & 8.37 & 10.14 & - & - & - \\
    BDRAR~\cite{zhu18_eccv} (2018) & 3.64 & 3.40 & 3.89 & 7.81 & 9.69 & 5.94 & 2.69 & \textbf{0.50} & 4.87 \\
    DC-DSPF~\cite{wang18_ijcai} (2018) & 4.90 & 4.70 & 5.10 & 7.90 & 6.50 & 9.30 & - & - & - \\
    DSDNet~\cite{zheng19_cvpr} (2019) & 3.45 & 3.33 & 3.58 & 7.59 & 9.74 & 5.44 & 2.17 & 1.36 & 2.98 \\
    MTMT-Net~\cite{chen20_cvpr} (2020) & 3.15 & 3.73 & 2.57 & 7.47 & 10.31 & 4.63 & 1.72 & 1.36 & 2.08 \\
    RCMPNet~\cite{liao2021mm} (2021) & 3.13 & 2.98 & 3.28 & 6.71 & 7.66 & 5.76 & 1.61 & 1.23 & 2.00 \\
    FDRNet~\cite{Zhu_2021_ICCV} (2021) & 3.04 & 2.91 & 3.18 & 7.28 & 8.31 & 6.26 & 1.55 & 1.22 & 1.88 \\
    SDCM ~\cite{zhu_mm2022} (2022) & 3.02 & - & - & 6.69 & - & - & 1.41 & - & -\\
    TranShadow~\cite{jie2022icassp} (2022) & 3.17 & - & - & 6.95 & - & - & 1.73 & - & - \\
    FCSD-Net~\cite{jose_wacv2023} (2023) & 3.15 & 2.74 & 2.56 & 6.96 & 8.32 & 5.60 & 1.69 & \underline{0.59} & 2.79 \\
    RMLANet~\cite{jie2022icme, jie2023rmlanet} (2023) & 2.97 & \underline{2.53} & 3.42 & \underline{6.41} & 6.69 & 6.14 & \underline{1.01} & 0.68 & 1.34 \\
    SDDNet~\cite{cong2023sddnet} (2023) & 2.94 & 3.23 & 2.64 & 6.59 & 7.89 & 5.29 & 1.27 & 1.01 & 1.52 \\
    SARA~\cite{Sun_2023_ICCV} (2023) & \underline{2.87} & 3.64 & \underline{2.10} & 7.01 & 9.43 & \underline{4.61} & 1.18 & 1.05 & \underline{1.31} \\
    \hline 
    \hline 
    Ours & \textbf{2.75} & \textbf{2.46} & 3.05 & \textbf{6.35} & \underline{6.11} & 6.60 & \textbf{0.86} & 0.65 & \textbf{1.07} \\
    \hline 
    \end{tabular}
    \label{table_detection_quantitative}
    \end{center}
  \end{table*}







\subsection{Datasets and Evaluation Metric}
\label{detection_dataset_metrics}
\noindent\textbf{Benchmark Dataset.} We employ four widely used benchmark datasets: SBU~\cite{sbu_dataset}, UCF~\cite{ucf_dataset}, ISTD~\cite{istd_dataset} and CUHK~\cite{gy_tip2021}, to fully evaluate the effectiveness of our method. Specifically, $4,089$ and $638$ pairs of the shadow images and shadow masks are provided in SBU for training and testing. However, as pointed out in~\cite{Yang_2023_ICCV}, original annotations in SBU are noisy and contain annotation errors. Therefore, we also report our performance on the relabeled test set of SBU~\cite{Yang_2023_ICCV} (named $SBUTestNew$ here). Besides, UCF contains only $145$ training and $76$ testing pairs, while ISTD consists of $1,330$ and $540$ triplets of shadow images, shadow masks and shadow-free images for training and testing. The CUHK dataset provides $7,350$, $1,050$ and $2,100$ pairs of shadow images and shadow masks for training, validation and testing. In practice, the model trained on the SBU training split is used to evaluate the performance on the SBU testing split, the relabeled $SBUTestNew$ and the UCF dataset, while the model trained on the ISTD training split or the CUHK training split are adopted to evaluate the performance on each corresponding testing split, respectively. \\

\noindent\textbf{Evaluation Metric.} For fair comparison, we choose Balance Error Rate (BER) as our evaluation metric. BER can be calculated as follows:
\begin{equation}
    BER = \left(1-\frac{1}{2}\left(\frac{T_p}{N_p} + \frac{T_n}{N_n}\right) \right )\times 100\ ,
\end{equation}
where $T_p$, $T_n$, $N_p$ and $N_n$ are the number of true shadow pixels, the number of true non-shadow pixels, the total number of shadow pixels and the total number of non-shadow pixels respectively. For $BER$, the smaller its value, the better the performance. We report three different $BER$ values: $BER$, ${BER}_S$ and ${BER}_{NS}$ which correspond to the whole image, the shadow regions and the non-shadow regions, respectively.

\begin{figure*}[!htb]
	\centering
	\vspace*{1.3mm}
	\begin{subfigure}{0.085\textwidth}
		\includegraphics[width=\textwidth]{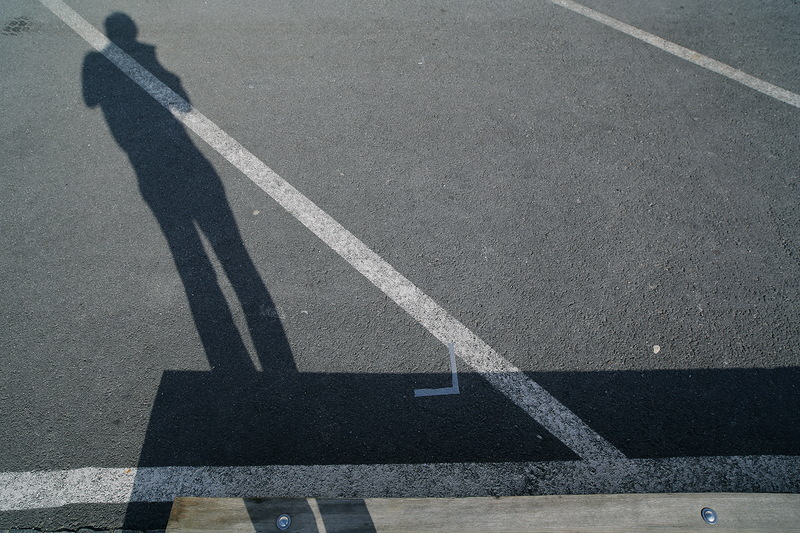}
	\end{subfigure}
	\begin{subfigure}{0.085\textwidth}
		\includegraphics[width=\textwidth]{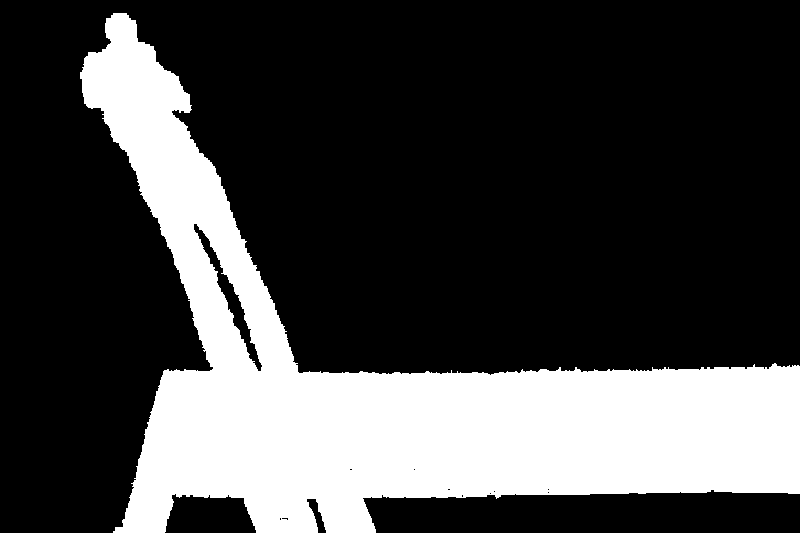}
	\end{subfigure}
	\begin{subfigure}{0.085\textwidth}
		\includegraphics[width=\textwidth]{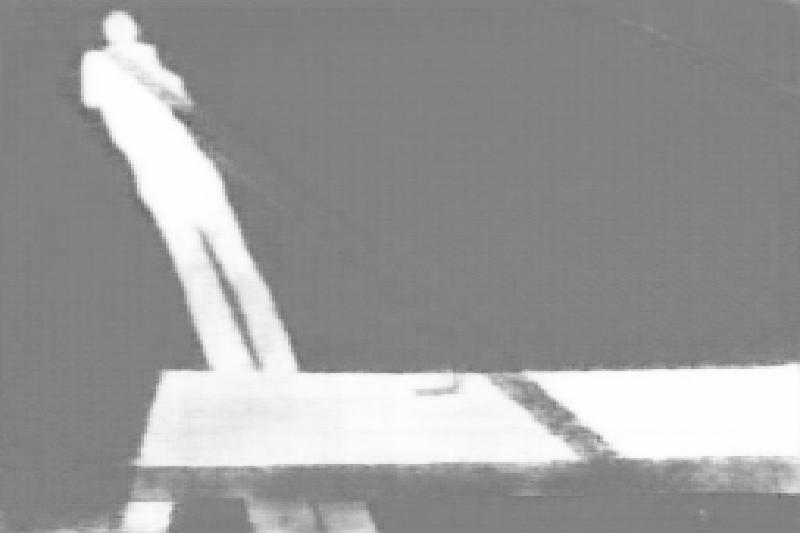}
	\end{subfigure}
	\begin{subfigure}{0.085\textwidth}
		\includegraphics[width=\textwidth]{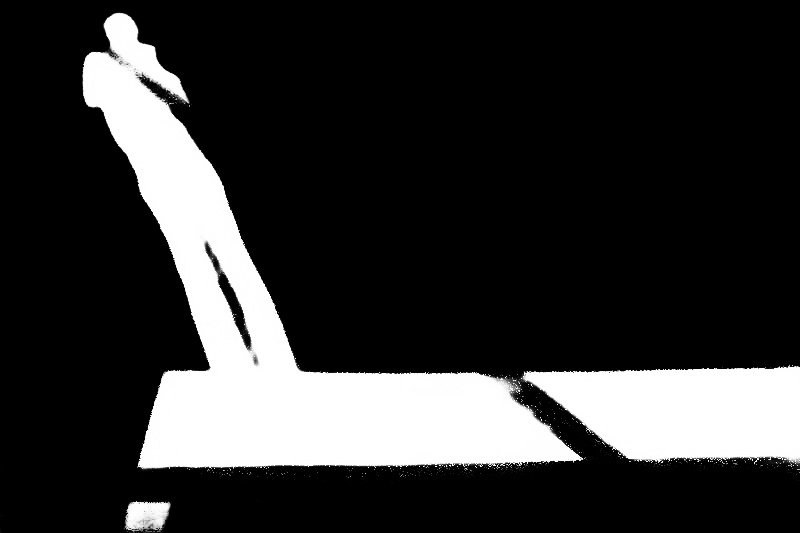}
	\end{subfigure}
	\begin{subfigure}{0.085\textwidth}
		\includegraphics[width=\textwidth]{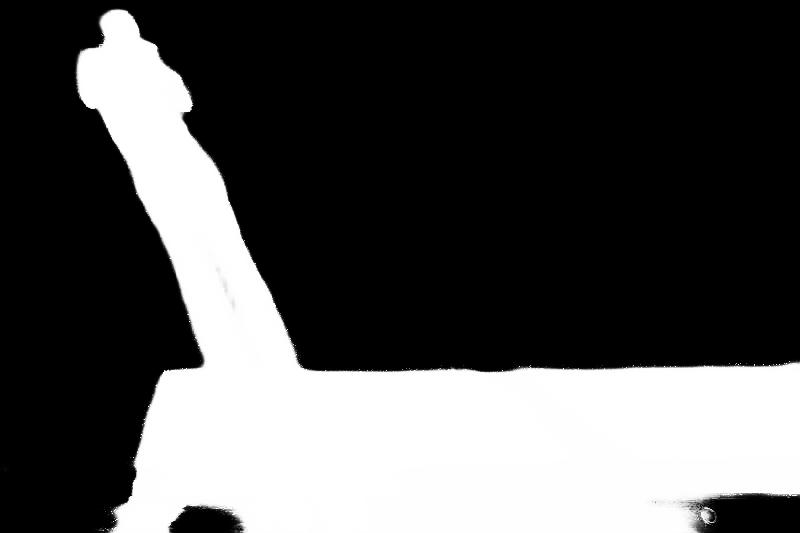}
	\end{subfigure}
	\begin{subfigure}{0.085\textwidth}
		\includegraphics[width=\textwidth]{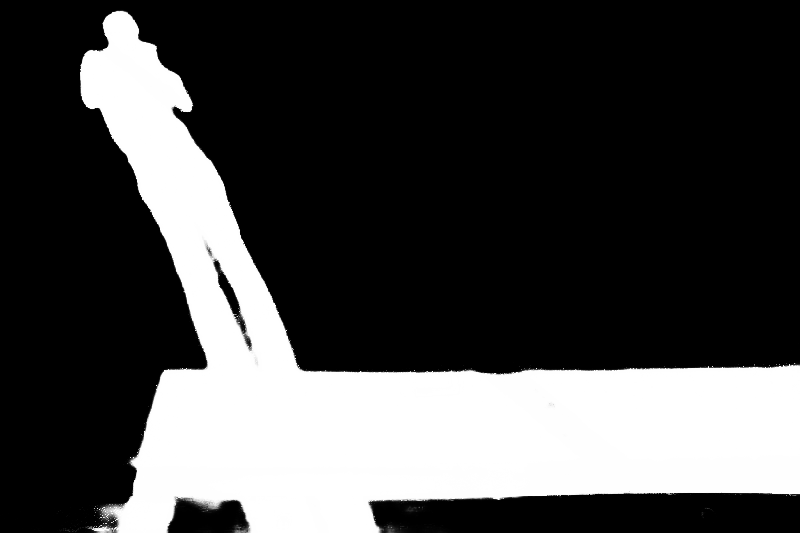}
	\end{subfigure}
	\begin{subfigure}{0.085\textwidth}
		\includegraphics[width=\textwidth]{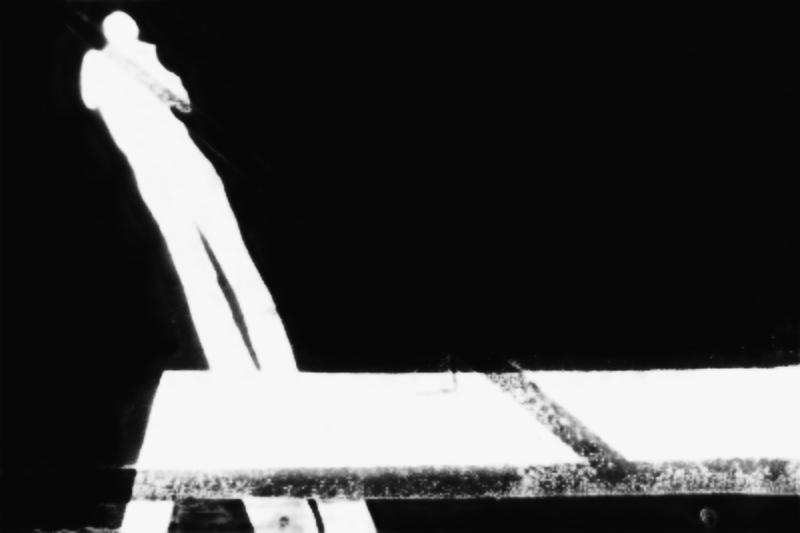}
	\end{subfigure}
	\begin{subfigure}{0.085\textwidth}
		\includegraphics[width=\textwidth]{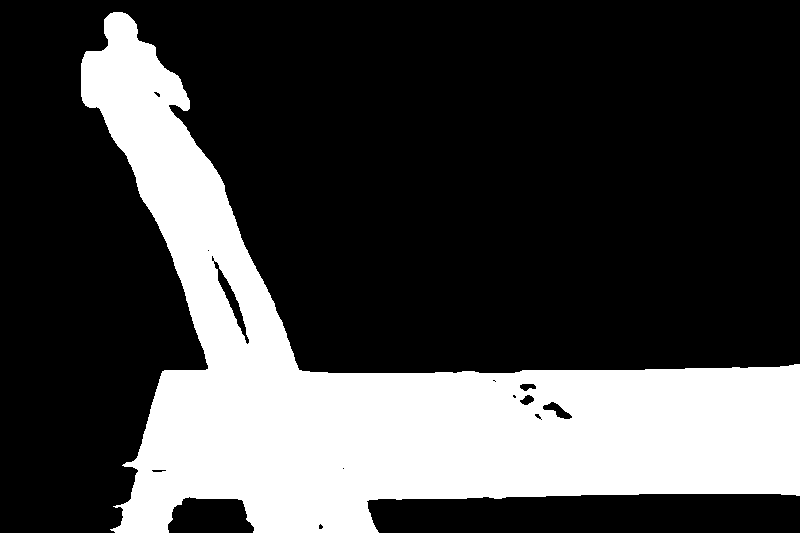}
	\end{subfigure}
	\begin{subfigure}{0.085\textwidth}
		\includegraphics[width=\textwidth]{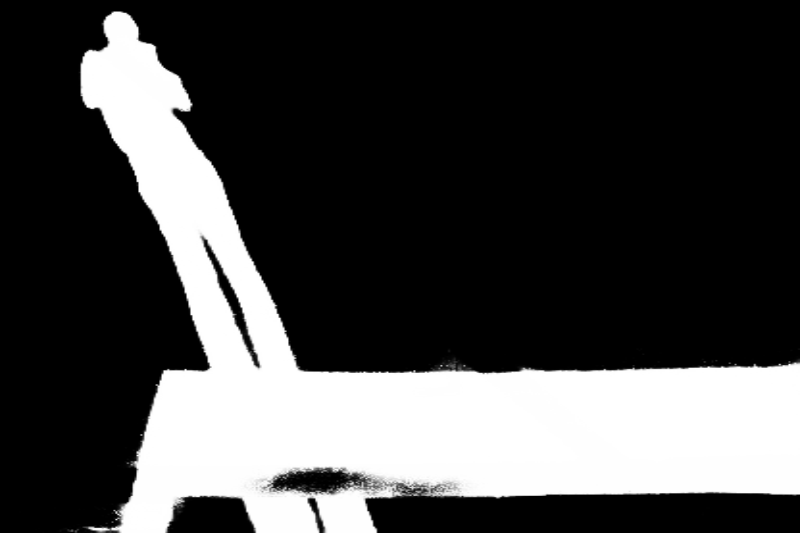}
	\end{subfigure}
	\begin{subfigure}{0.085\textwidth}
		\includegraphics[width=\textwidth]{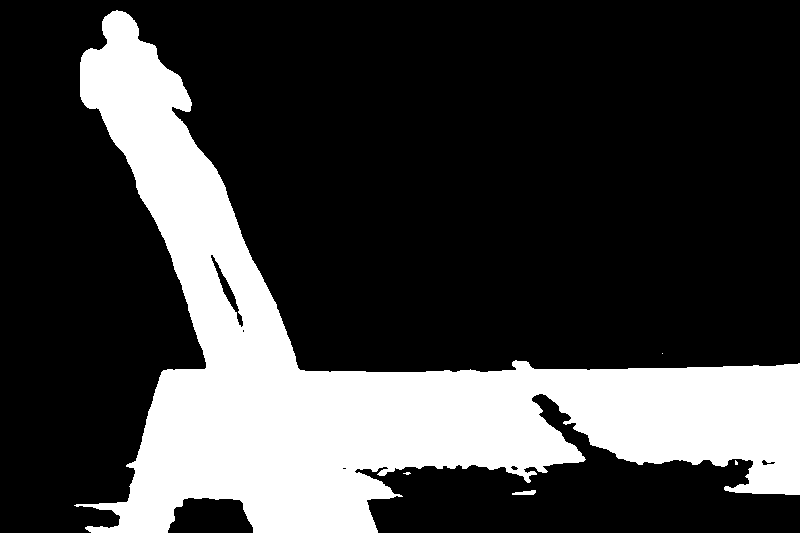}
	\end{subfigure}
	\begin{subfigure}{0.085\textwidth}
		\includegraphics[width=\textwidth]{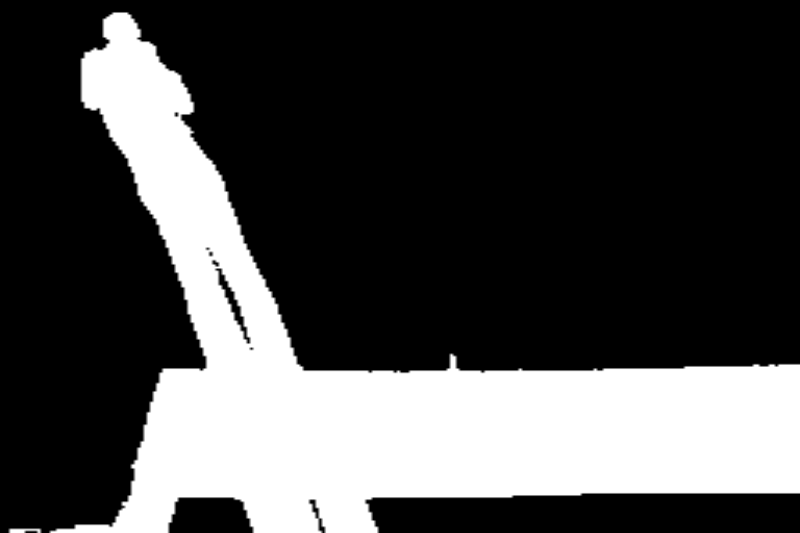}
	\end{subfigure}

	\vspace*{1.3mm}
	\begin{subfigure}{0.085\textwidth}
		\includegraphics[width=\textwidth]{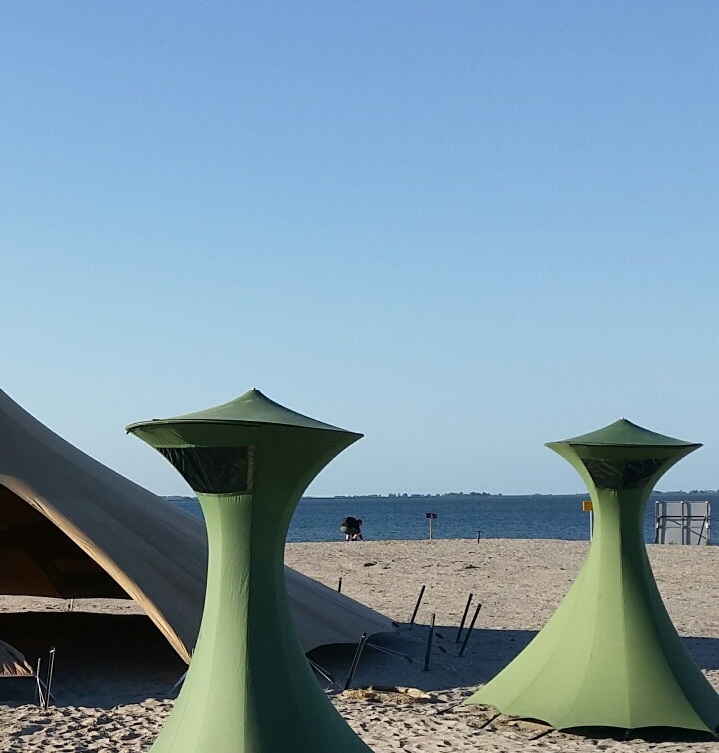}
	\end{subfigure}
	\begin{subfigure}{0.085\textwidth}
		\includegraphics[width=\textwidth]{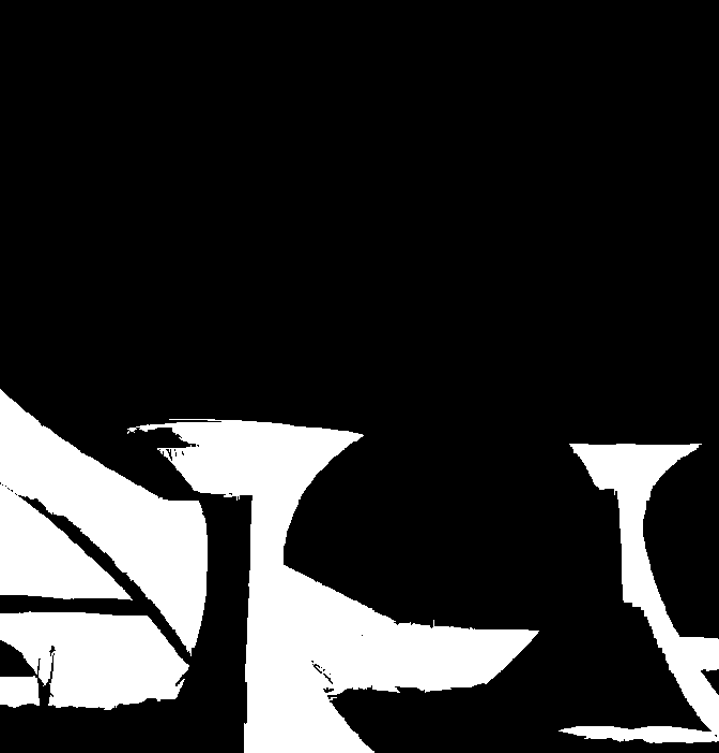}
	\end{subfigure}
	\begin{subfigure}{0.085\textwidth}
		\includegraphics[width=\textwidth]{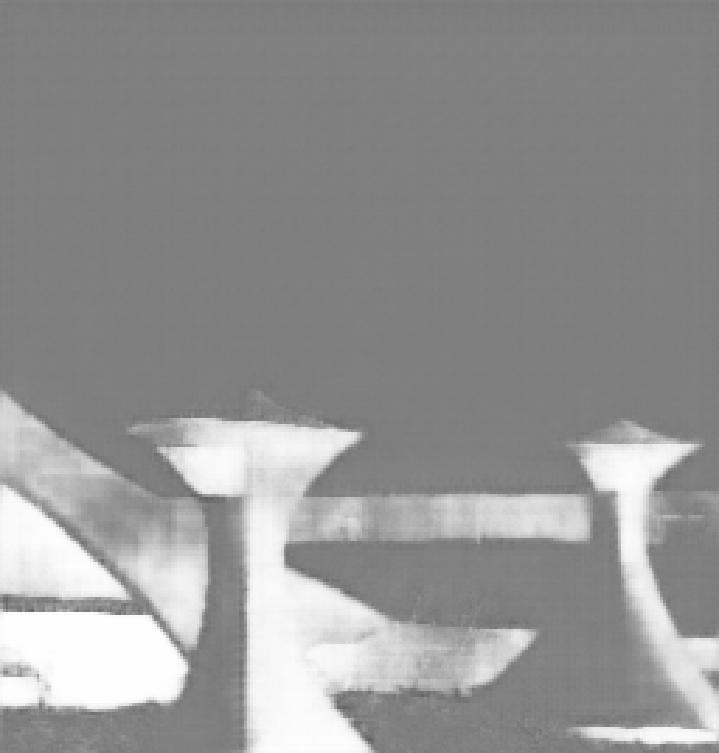}
	\end{subfigure}
	\begin{subfigure}{0.085\textwidth}
		\includegraphics[width=\textwidth]{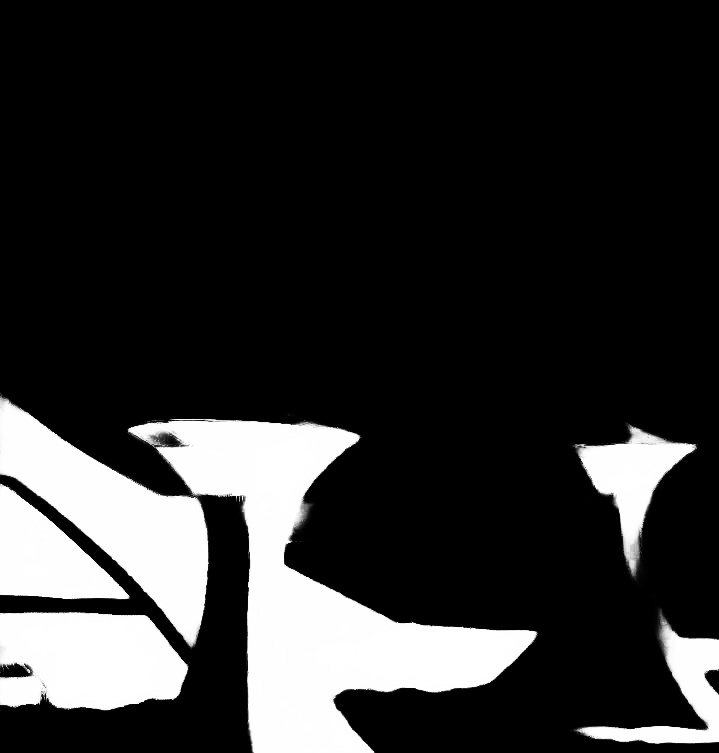}
	\end{subfigure}
	\begin{subfigure}{0.085\textwidth}
		\includegraphics[width=\textwidth]{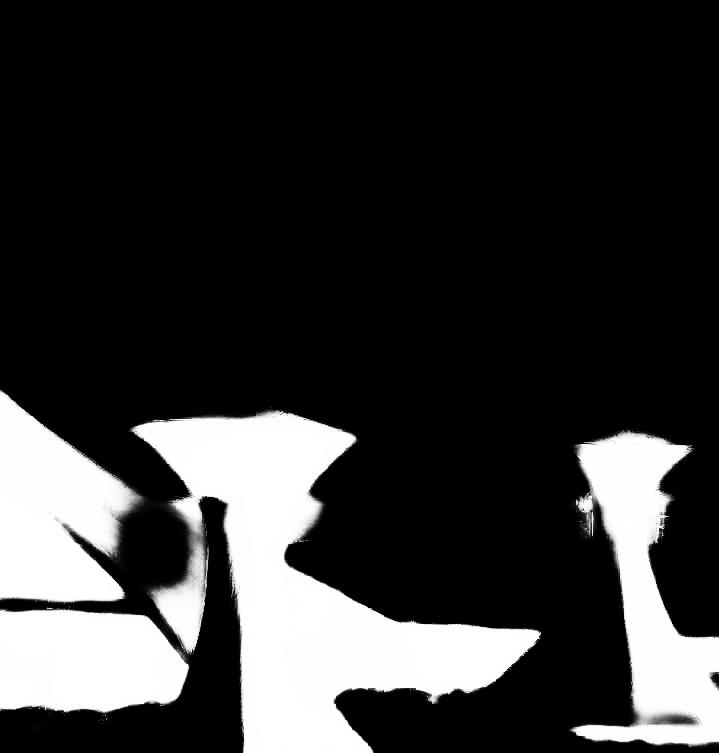}
	\end{subfigure}
	\begin{subfigure}{0.085\textwidth}
		\includegraphics[width=\textwidth]{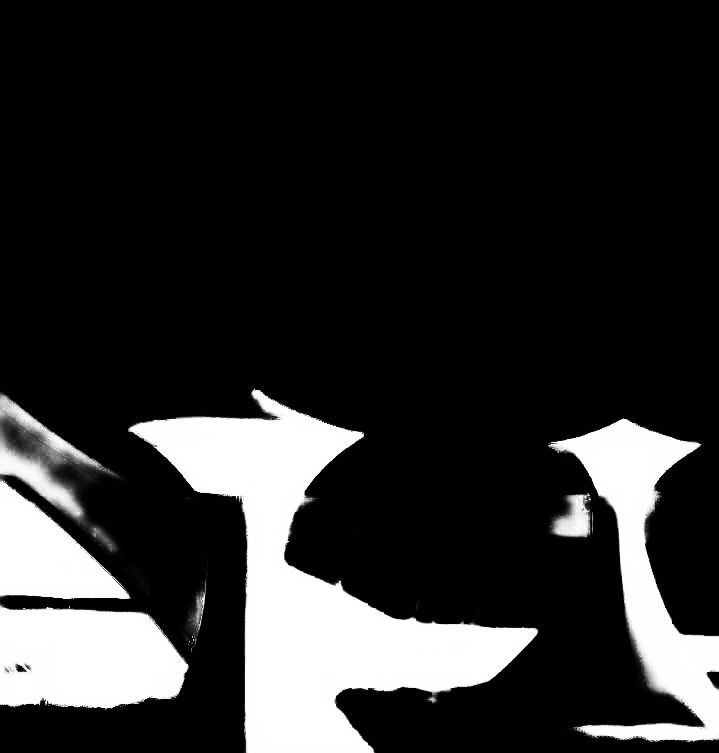}
	\end{subfigure}
	\begin{subfigure}{0.085\textwidth}
		\includegraphics[width=\textwidth]{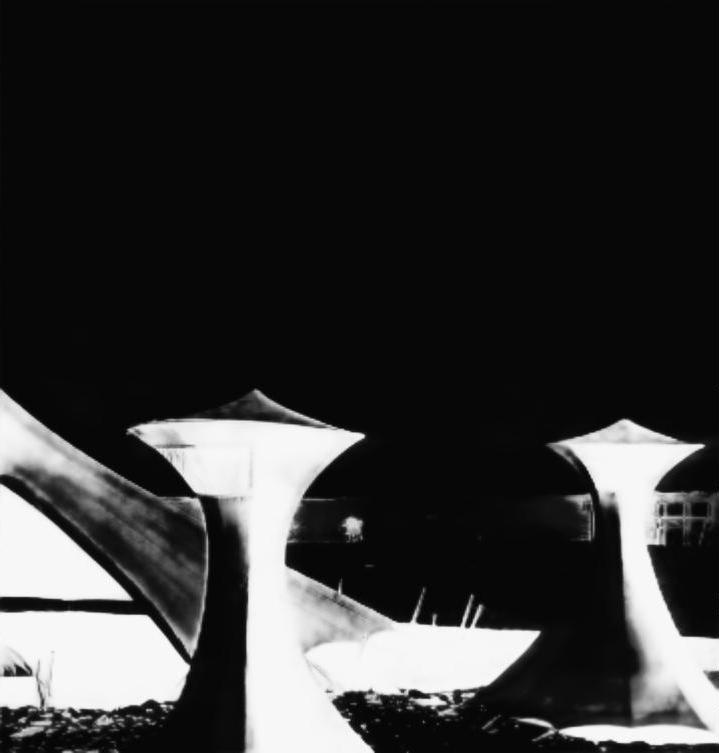}
	\end{subfigure}
	\begin{subfigure}{0.085\textwidth}
		\includegraphics[width=\textwidth]{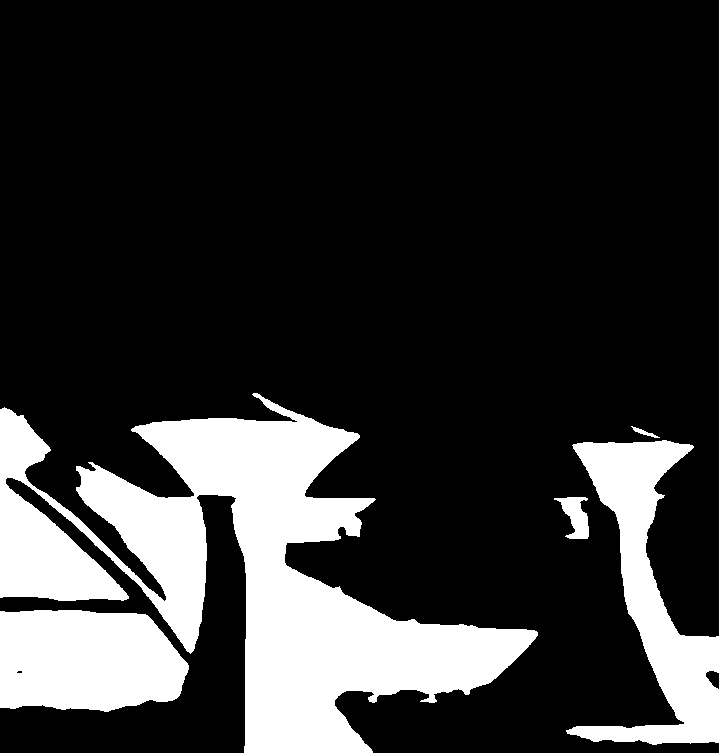}
	\end{subfigure}
	\begin{subfigure}{0.085\textwidth}
		\includegraphics[width=\textwidth]{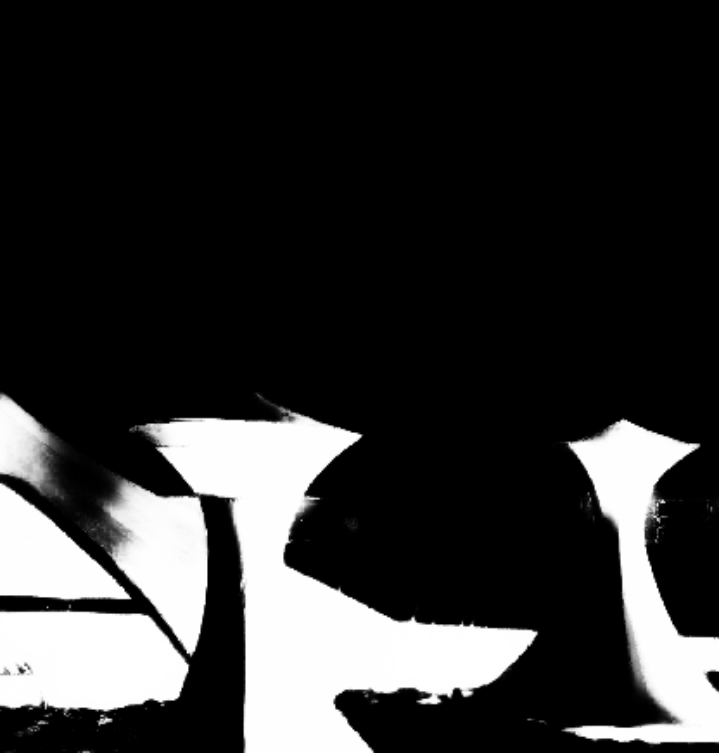}
	\end{subfigure}
	\begin{subfigure}{0.085\textwidth}
		\includegraphics[width=\textwidth]{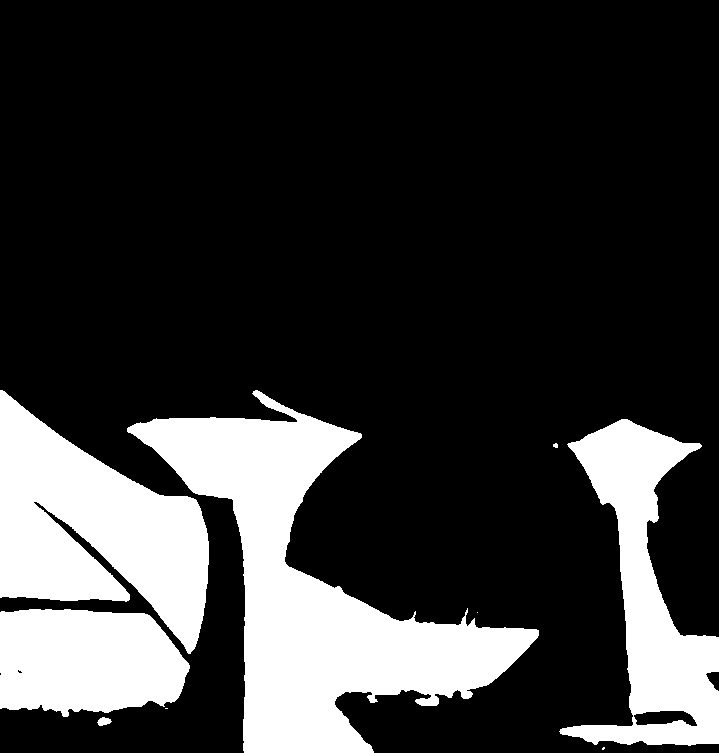}
	\end{subfigure}
	\begin{subfigure}{0.085\textwidth}
		\includegraphics[width=\textwidth]{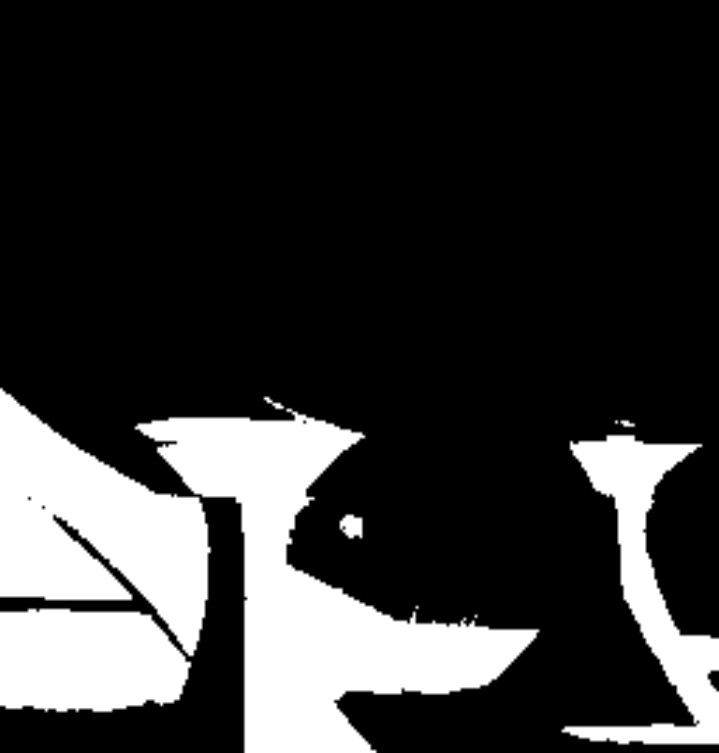}
	\end{subfigure}
	
	\vspace*{1.3mm}
	\begin{subfigure}{0.085\textwidth}
		\includegraphics[width=\textwidth]{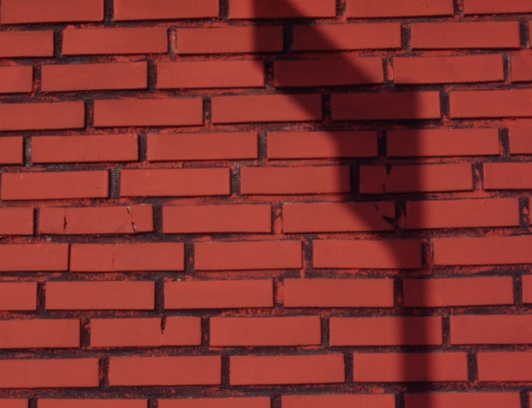}
	\end{subfigure}
	\begin{subfigure}{0.085\textwidth}
		\includegraphics[width=\textwidth]{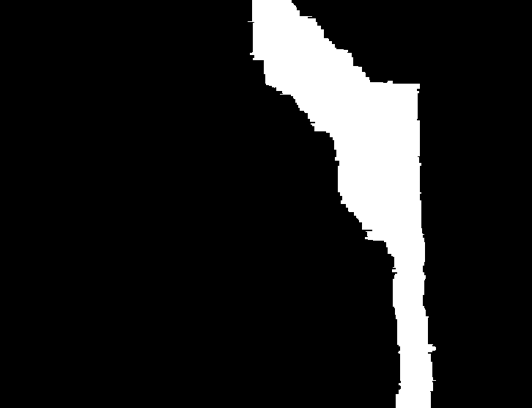}
	\end{subfigure}
	\begin{subfigure}{0.085\textwidth}
		\includegraphics[width=\textwidth]{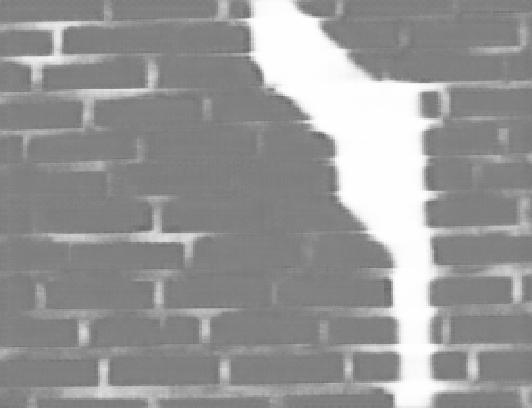}
	\end{subfigure}
	\begin{subfigure}{0.085\textwidth}
		\includegraphics[width=\textwidth]{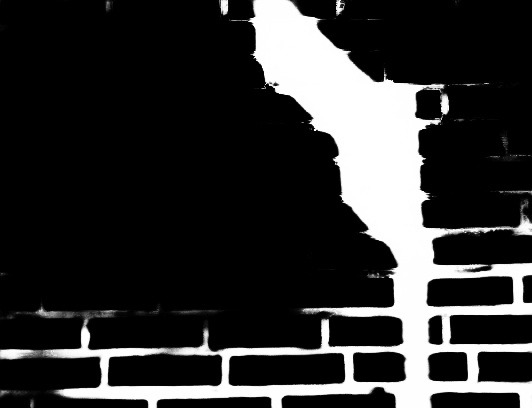}
	\end{subfigure}
	\begin{subfigure}{0.085\textwidth}
		\includegraphics[width=\textwidth]{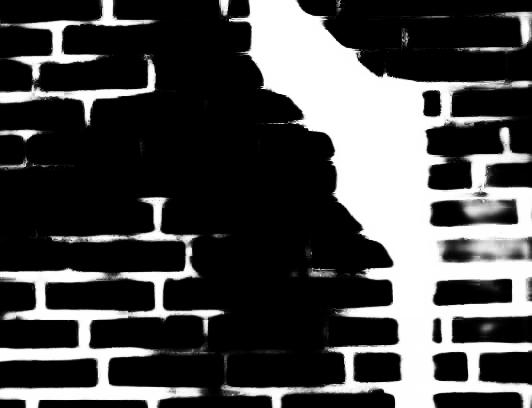}
	\end{subfigure}
	\begin{subfigure}{0.085\textwidth}
		\includegraphics[width=\textwidth]{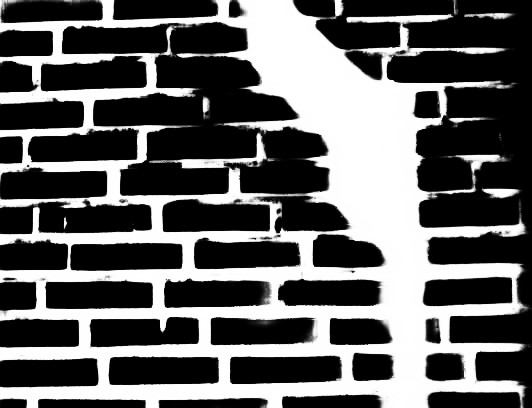}
	\end{subfigure}
	\begin{subfigure}{0.085\textwidth}
		\includegraphics[width=\textwidth]{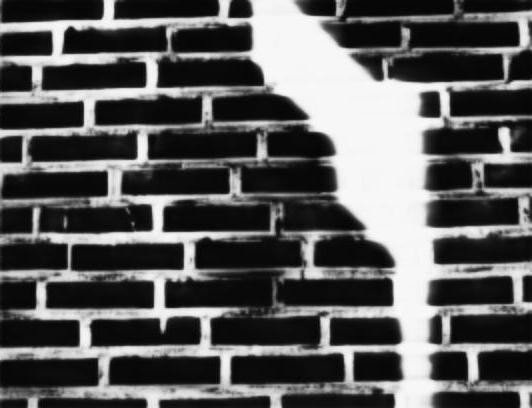}
	\end{subfigure}
	\begin{subfigure}{0.085\textwidth}
		\includegraphics[width=\textwidth]{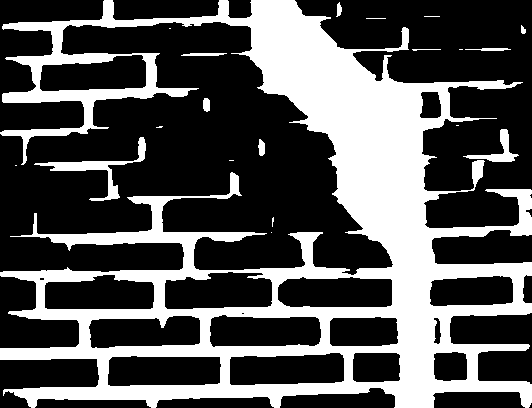}
	\end{subfigure}
	\begin{subfigure}{0.085\textwidth}
		\includegraphics[width=\textwidth]{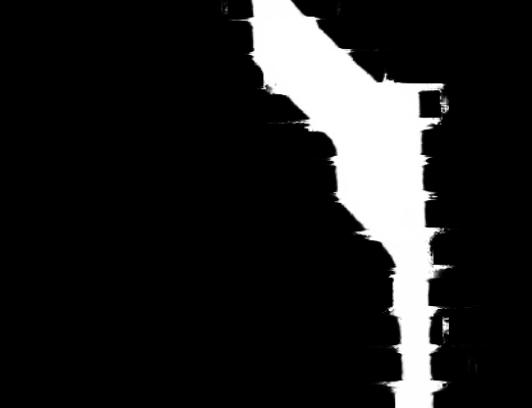}
	\end{subfigure}
	\begin{subfigure}{0.085\textwidth}
		\includegraphics[width=\textwidth]{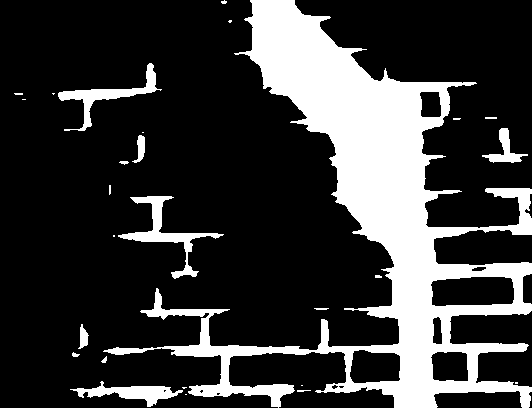}
	\end{subfigure}
	\begin{subfigure}{0.085\textwidth}
		\includegraphics[width=\textwidth]{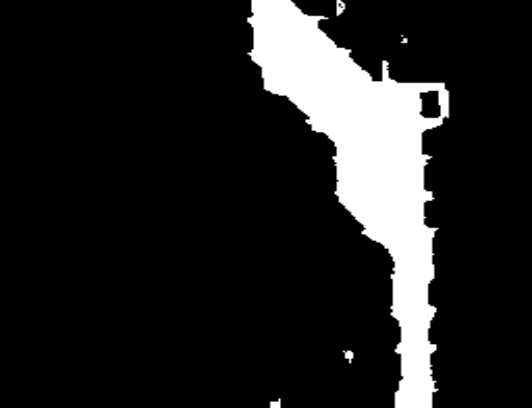}
	\end{subfigure}
	
	\vspace*{1.3mm}
	\begin{subfigure}{0.085\textwidth}
		\includegraphics[width=\textwidth]{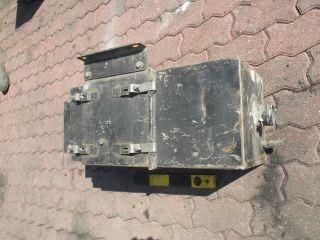}
	\end{subfigure}
	\begin{subfigure}{0.085\textwidth}
		\includegraphics[width=\textwidth]{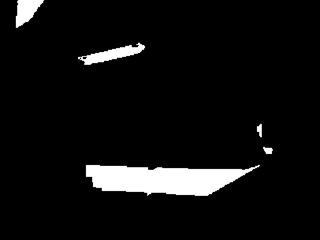}
	\end{subfigure}
	\begin{subfigure}{0.085\textwidth}
		\includegraphics[width=\textwidth]{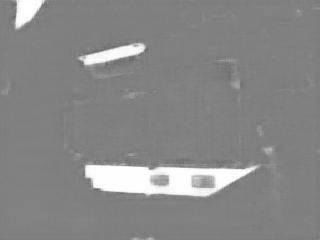}
	\end{subfigure}
	\begin{subfigure}{0.085\textwidth}
		\includegraphics[width=\textwidth]{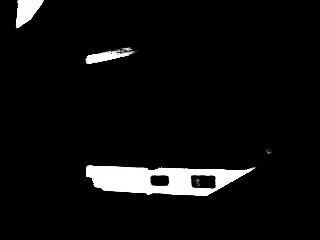}
	\end{subfigure}
	\begin{subfigure}{0.085\textwidth}
		\includegraphics[width=\textwidth]{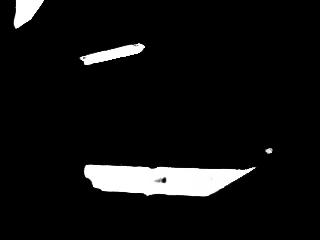}
	\end{subfigure}
	\begin{subfigure}{0.085\textwidth}
		\includegraphics[width=\textwidth]{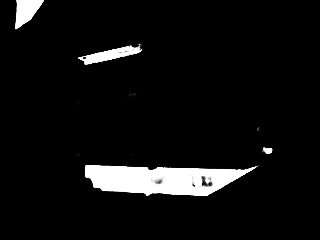}
	\end{subfigure}
	\begin{subfigure}{0.085\textwidth}
		\includegraphics[width=\textwidth]{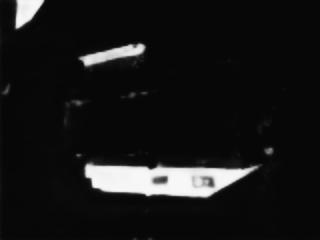}
	\end{subfigure}
	\begin{subfigure}{0.085\textwidth}
		\includegraphics[width=\textwidth]{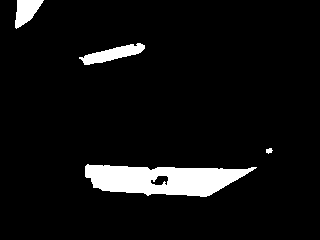}
	\end{subfigure}
	\begin{subfigure}{0.085\textwidth}
		\includegraphics[width=\textwidth]{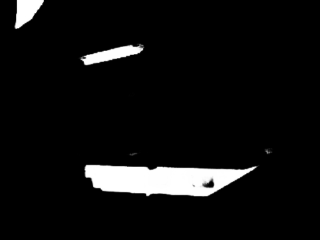}
	\end{subfigure}
	\begin{subfigure}{0.085\textwidth}
		\includegraphics[width=\textwidth]{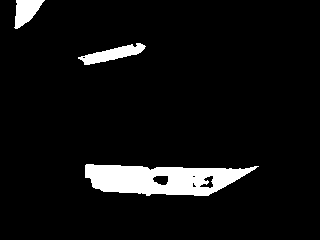}
	\end{subfigure}
	\begin{subfigure}{0.085\textwidth}
		\includegraphics[width=\textwidth]{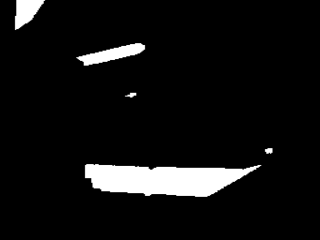}
	\end{subfigure}

	\vspace*{1.3mm}
	\begin{subfigure}{0.085\textwidth}
		\includegraphics[width=\textwidth]{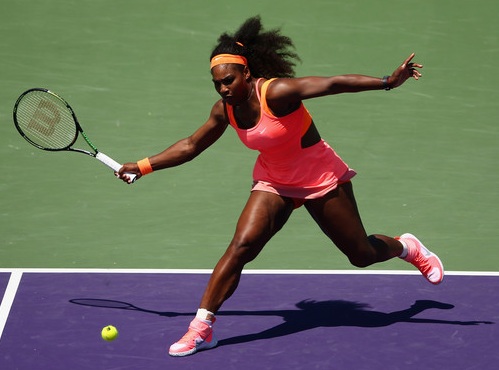}
	\end{subfigure}
	\begin{subfigure}{0.085\textwidth}
		\includegraphics[width=\textwidth]{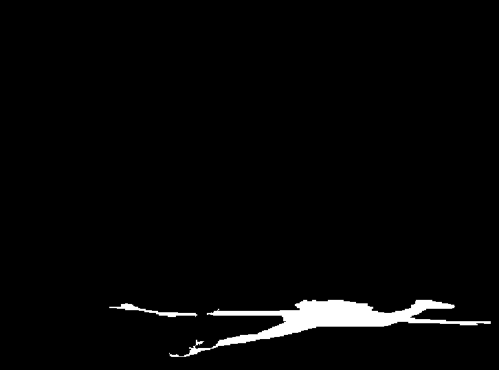}
	\end{subfigure}
	\begin{subfigure}{0.085\textwidth}
		\includegraphics[width=\textwidth]{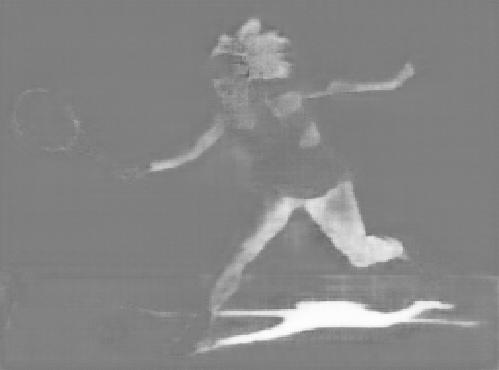}
	\end{subfigure}
	\begin{subfigure}{0.085\textwidth}
		\includegraphics[width=\textwidth]{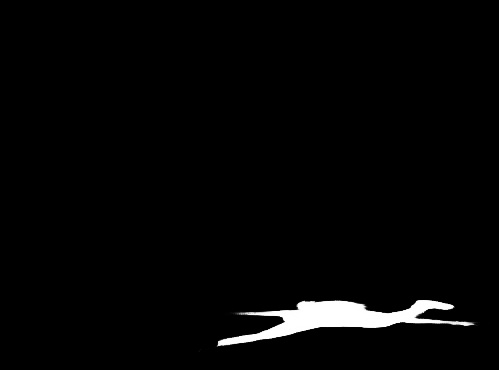}
	\end{subfigure}
	\begin{subfigure}{0.085\textwidth}
		\includegraphics[width=\textwidth]{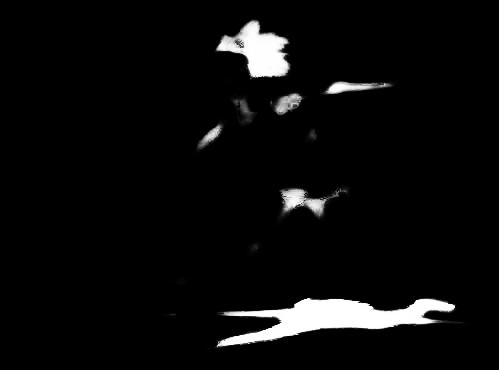}
	\end{subfigure}
	\begin{subfigure}{0.085\textwidth}
		\includegraphics[width=\textwidth]{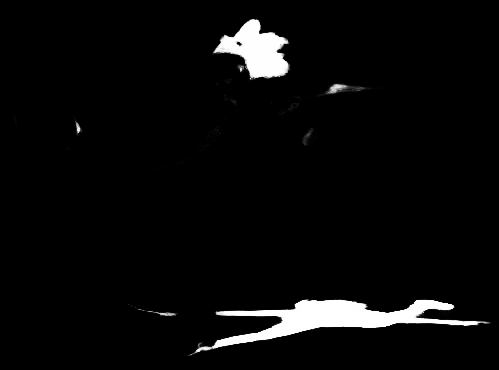}
	\end{subfigure}
	\begin{subfigure}{0.085\textwidth}
		\includegraphics[width=\textwidth]{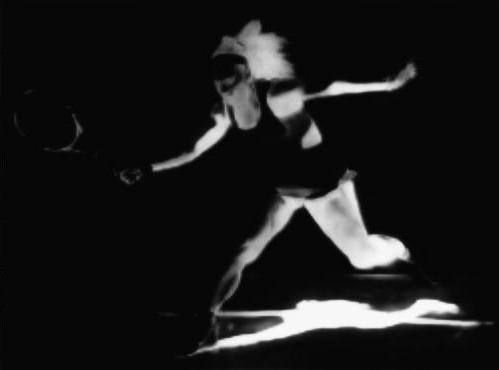}
	\end{subfigure}
	\begin{subfigure}{0.085\textwidth}
		\includegraphics[width=\textwidth]{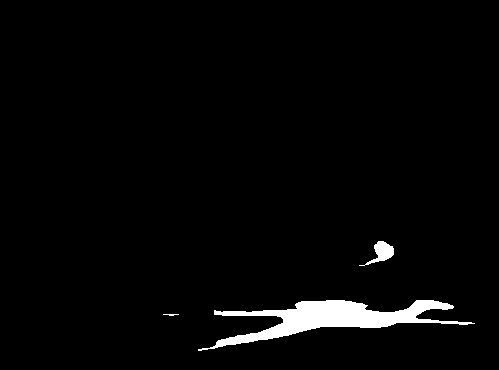}
	\end{subfigure}
	\begin{subfigure}{0.085\textwidth}
		\includegraphics[width=\textwidth]{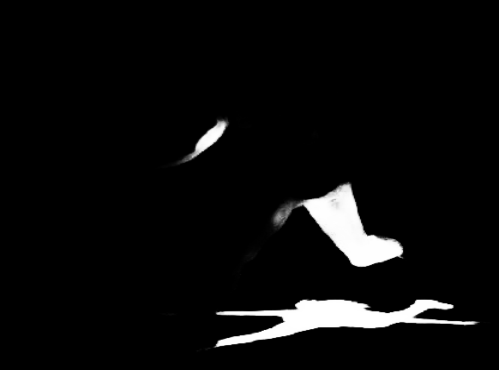}
	\end{subfigure}
	\begin{subfigure}{0.085\textwidth}
		\includegraphics[width=\textwidth]{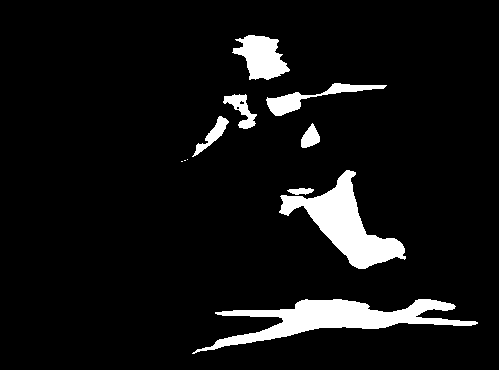}
	\end{subfigure}
	\begin{subfigure}{0.085\textwidth}
		\includegraphics[width=\textwidth]{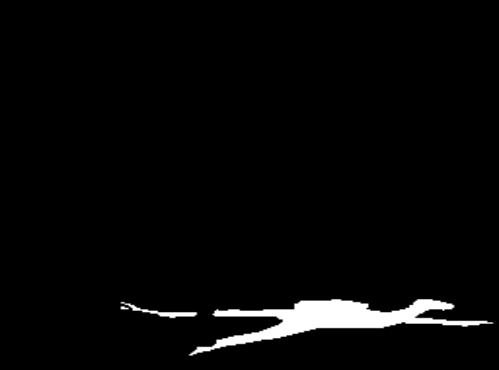}
	\end{subfigure}

	\vspace*{1.3mm}
	\begin{subfigure}{0.085\textwidth}
		\includegraphics[width=\textwidth]{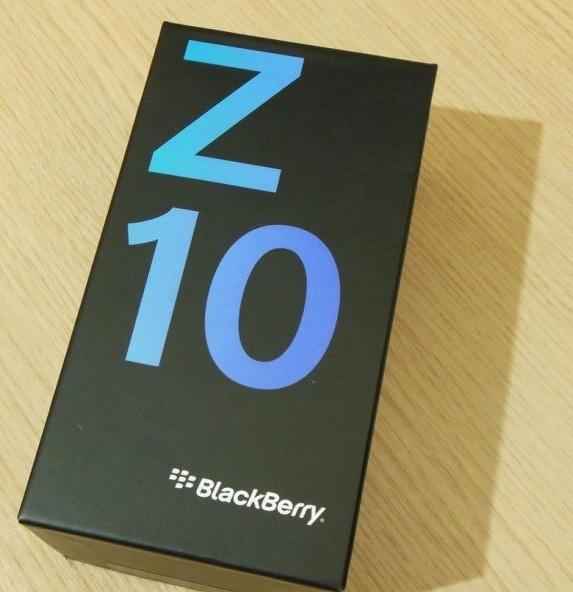}
	\end{subfigure}
	\begin{subfigure}{0.085\textwidth}
		\includegraphics[width=\textwidth]{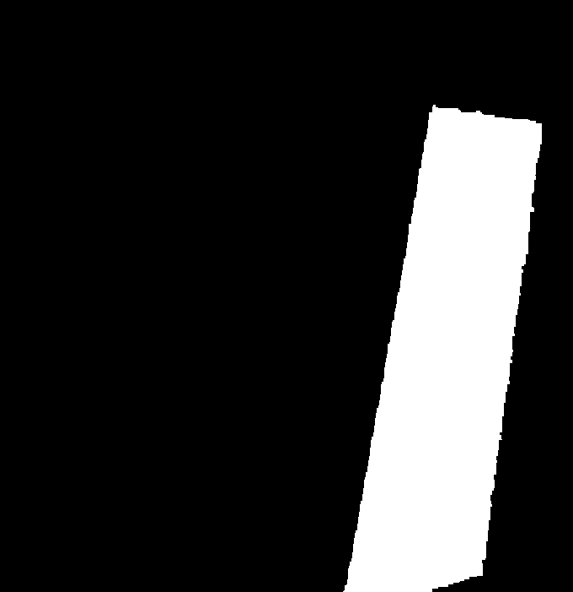}
	\end{subfigure}
	\begin{subfigure}{0.085\textwidth}
		\includegraphics[width=\textwidth]{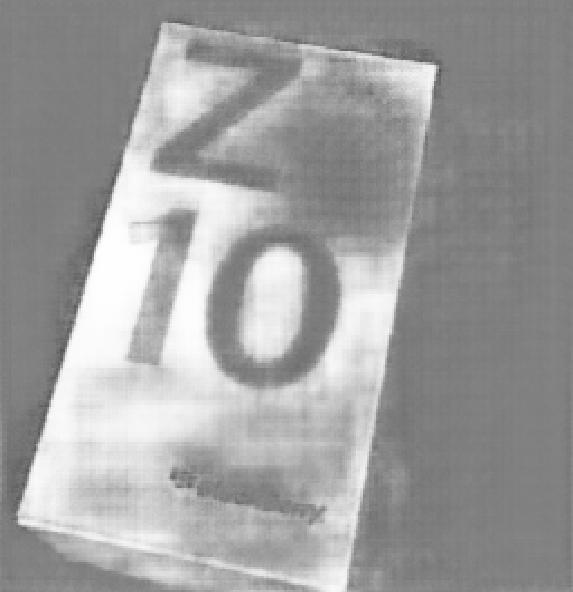}
	\end{subfigure}
	\begin{subfigure}{0.085\textwidth}
		\includegraphics[width=\textwidth]{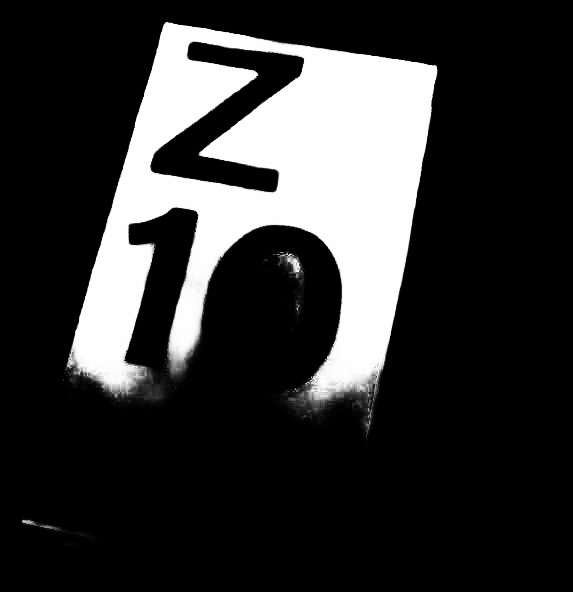}
	\end{subfigure}
	\begin{subfigure}{0.085\textwidth}
		\includegraphics[width=\textwidth]{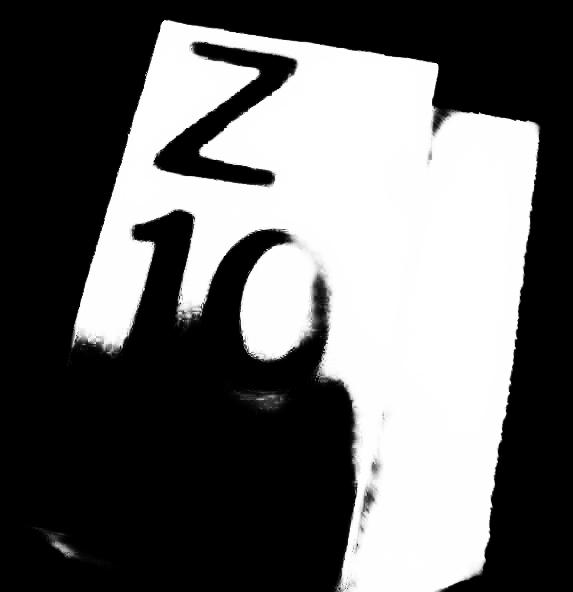}
	\end{subfigure}
	\begin{subfigure}{0.085\textwidth}
		\includegraphics[width=\textwidth]{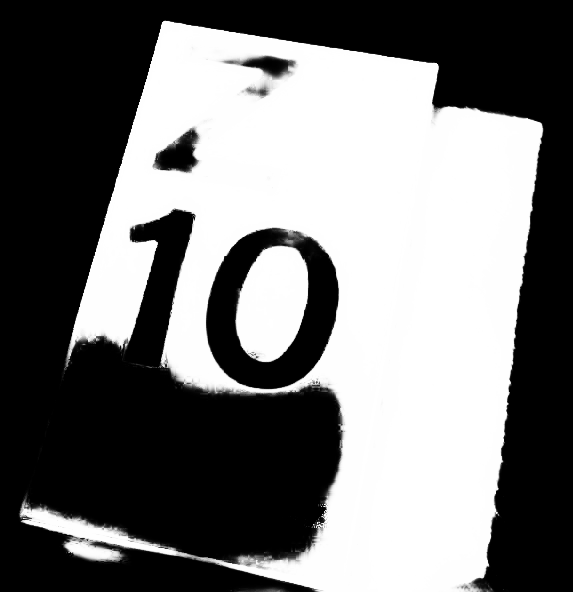}
	\end{subfigure}
	\begin{subfigure}{0.085\textwidth}
		\includegraphics[width=\textwidth]{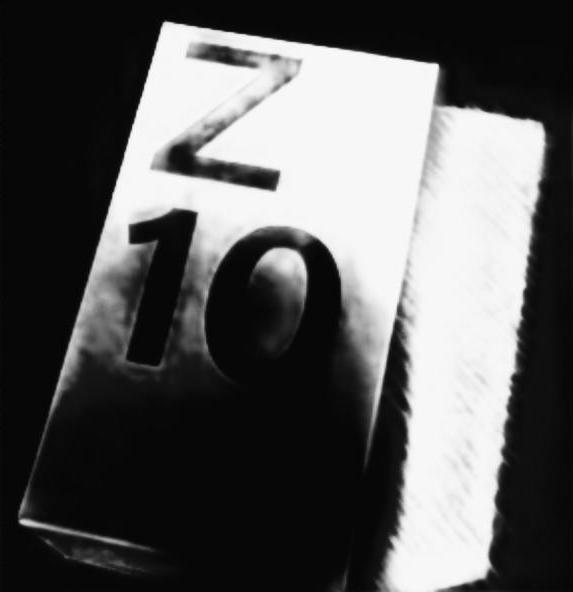}
	\end{subfigure}
	\begin{subfigure}{0.085\textwidth}
		\includegraphics[width=\textwidth]{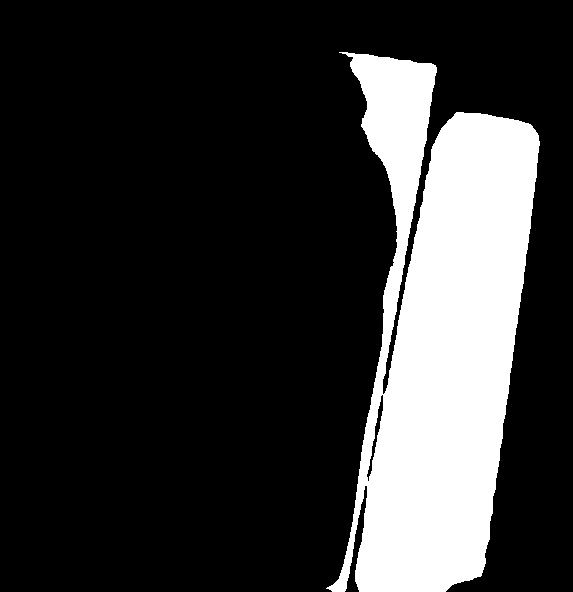}
	\end{subfigure}
	\begin{subfigure}{0.085\textwidth}
		\includegraphics[width=\textwidth]{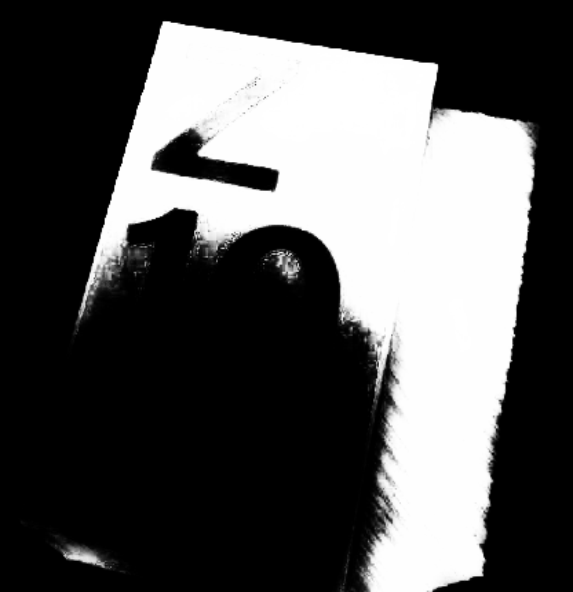}
	\end{subfigure}
	\begin{subfigure}{0.085\textwidth}
		\includegraphics[width=\textwidth]{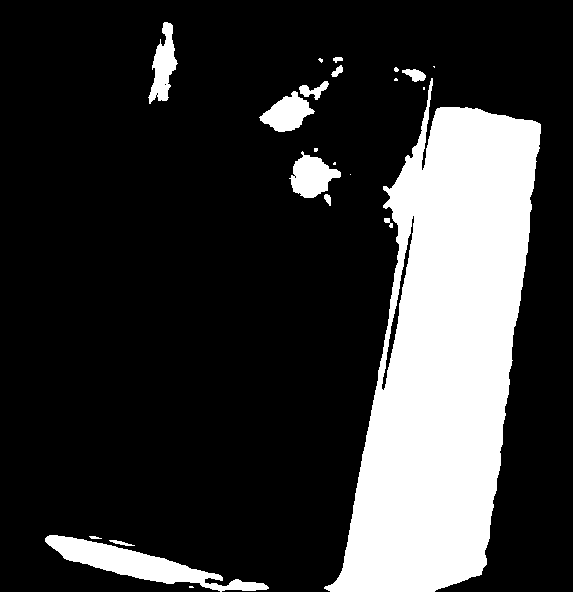}
	\end{subfigure}
	\begin{subfigure}{0.085\textwidth}
		\includegraphics[width=\textwidth]{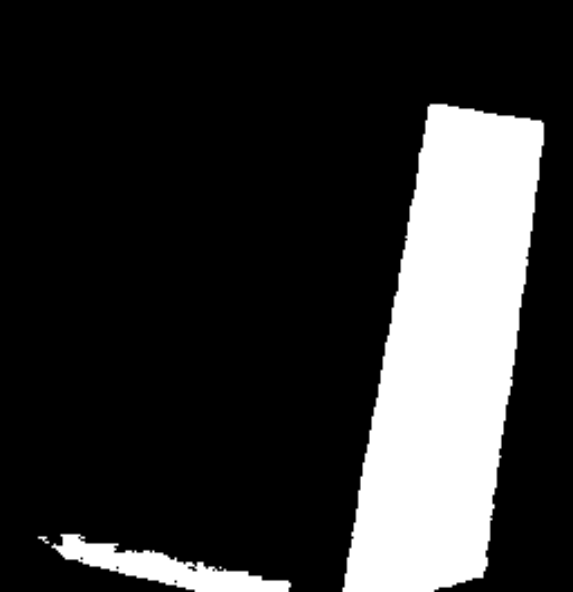}
	\end{subfigure}

	\vspace*{1.3mm}
	\begin{subfigure}{0.085\textwidth}
		\includegraphics[width=\textwidth]{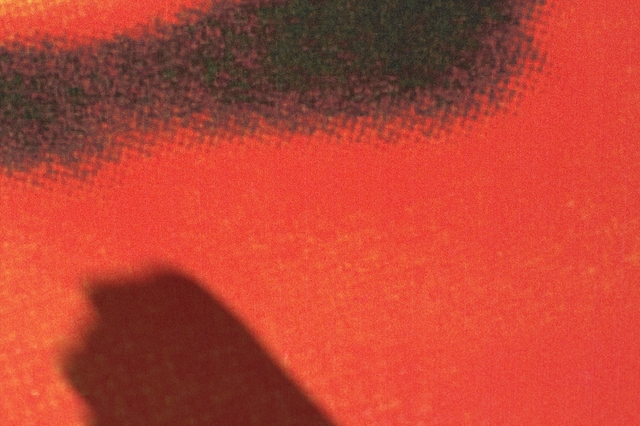}
	\end{subfigure}
	\begin{subfigure}{0.085\textwidth}
		\includegraphics[width=\textwidth]{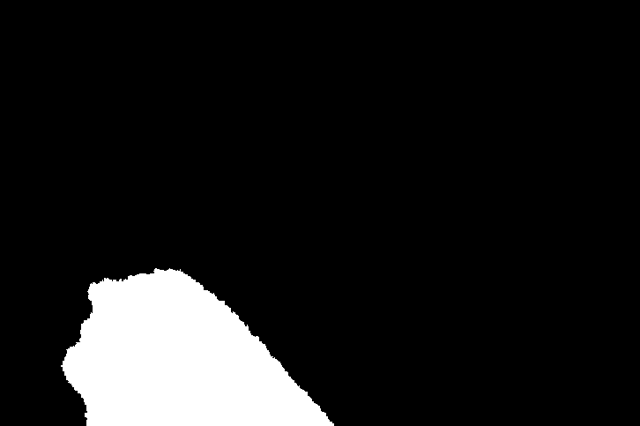}
	\end{subfigure}
	\begin{subfigure}{0.085\textwidth}
		\includegraphics[width=\textwidth]{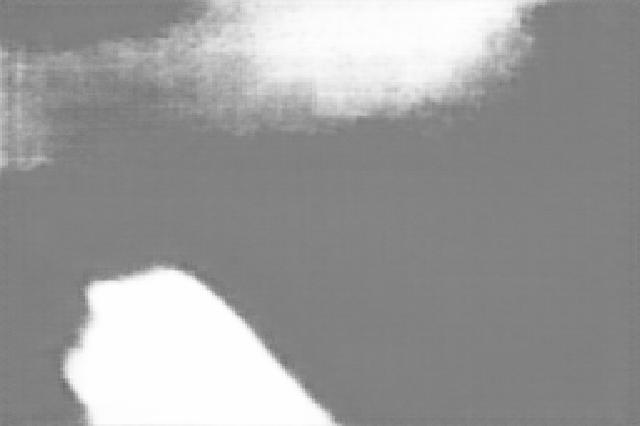}
	\end{subfigure}
	\begin{subfigure}{0.085\textwidth}
		\includegraphics[width=\textwidth]{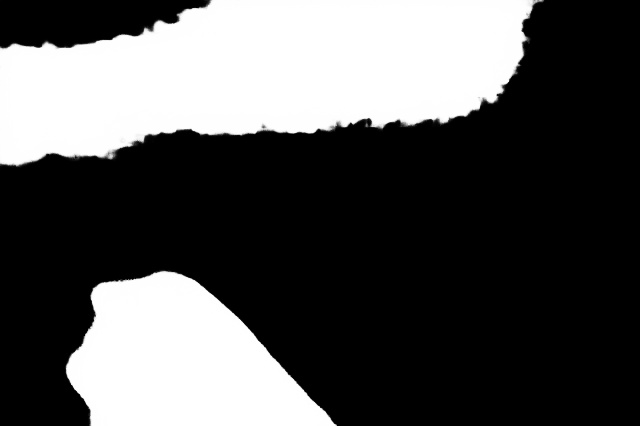}
	\end{subfigure}
	\begin{subfigure}{0.085\textwidth}
		\includegraphics[width=\textwidth]{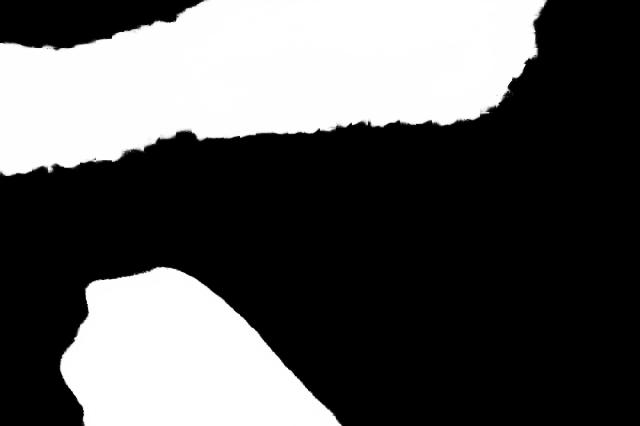}
	\end{subfigure}
	\begin{subfigure}{0.085\textwidth}
		\includegraphics[width=\textwidth]{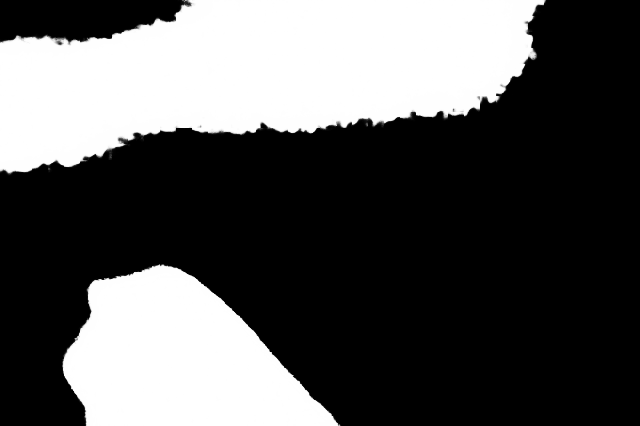}
	\end{subfigure}
	\begin{subfigure}{0.085\textwidth}
		\includegraphics[width=\textwidth]{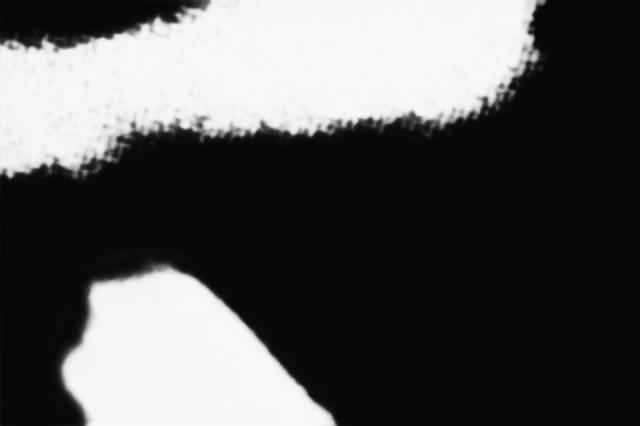}
	\end{subfigure}
	\begin{subfigure}{0.085\textwidth}
		\includegraphics[width=\textwidth]{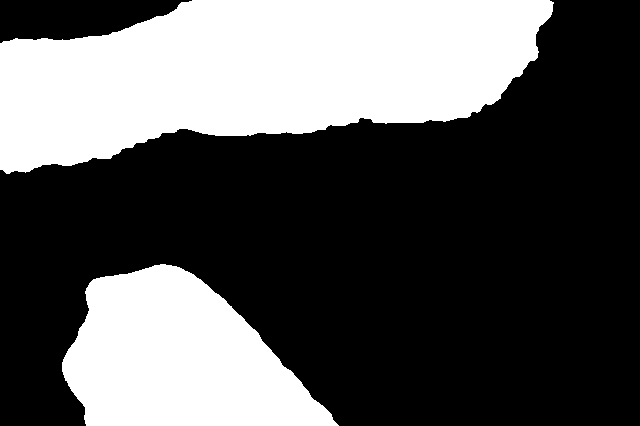}
	\end{subfigure}
	\begin{subfigure}{0.085\textwidth}
		\includegraphics[width=\textwidth]{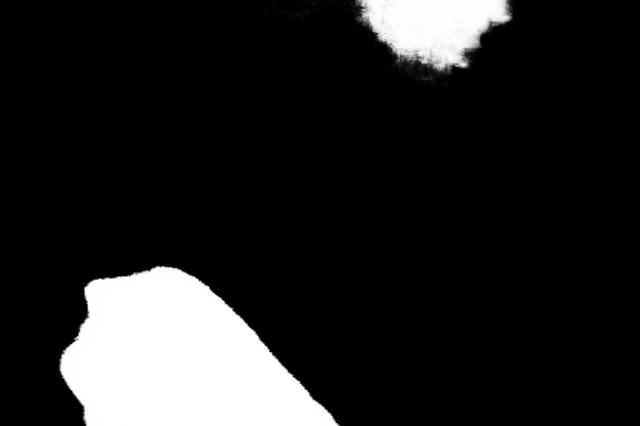}
	\end{subfigure}
	\begin{subfigure}{0.085\textwidth}
		\includegraphics[width=\textwidth]{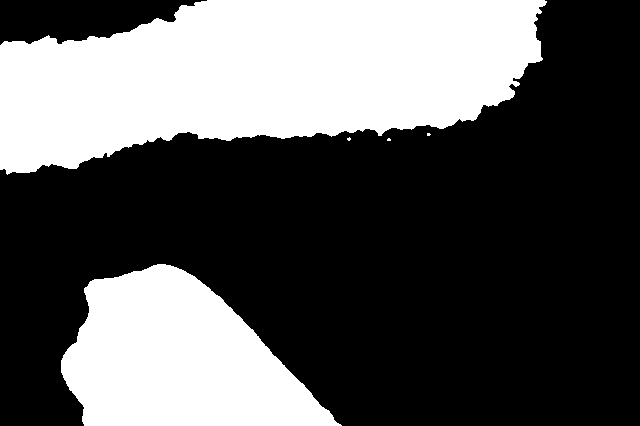}
	\end{subfigure}
	\begin{subfigure}{0.085\textwidth}
		\includegraphics[width=\textwidth]{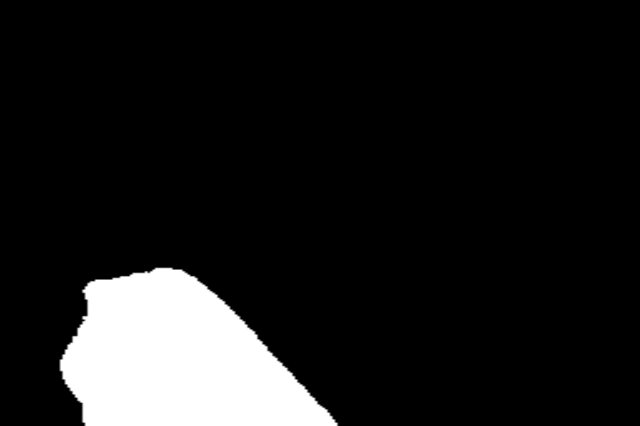}
	\end{subfigure}

	\vspace*{1.3mm}
	\begin{subfigure}{0.085\textwidth}
		\includegraphics[width=\textwidth]{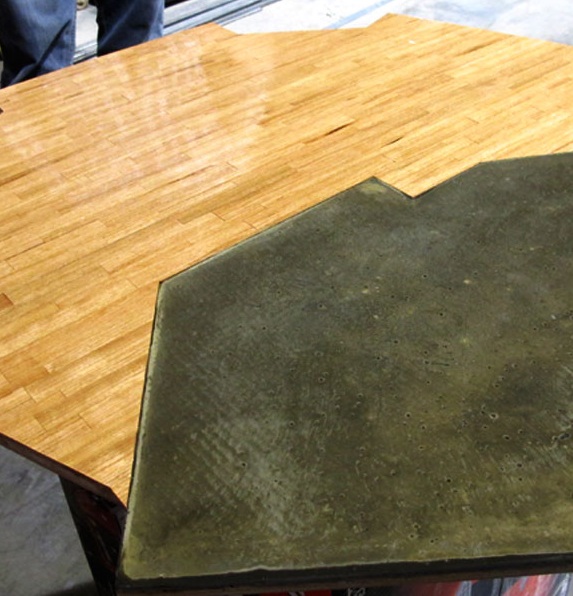}
	\end{subfigure}
	\begin{subfigure}{0.085\textwidth}
		\includegraphics[width=\textwidth]{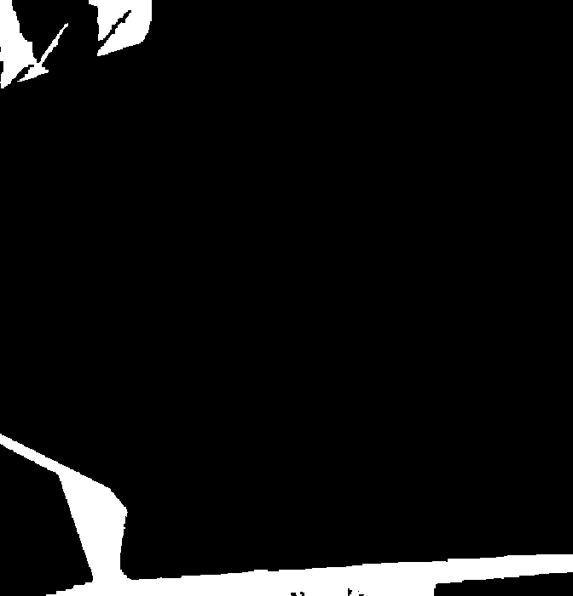}
	\end{subfigure}
	\begin{subfigure}{0.085\textwidth}
		\includegraphics[width=\textwidth]{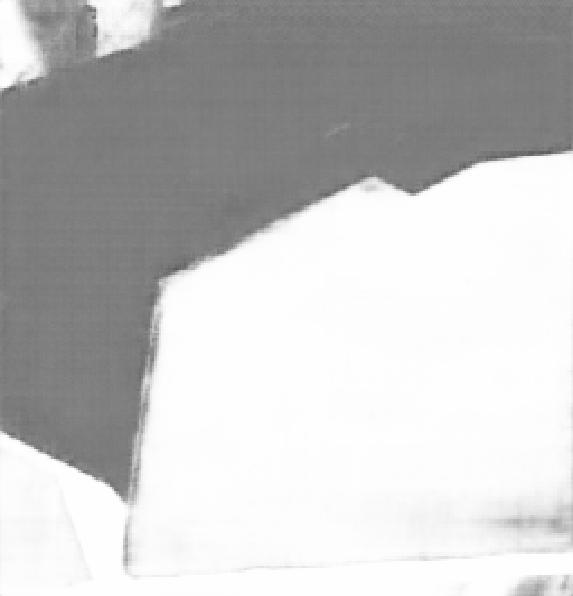}
	\end{subfigure}
	\begin{subfigure}{0.085\textwidth}
		\includegraphics[width=\textwidth]{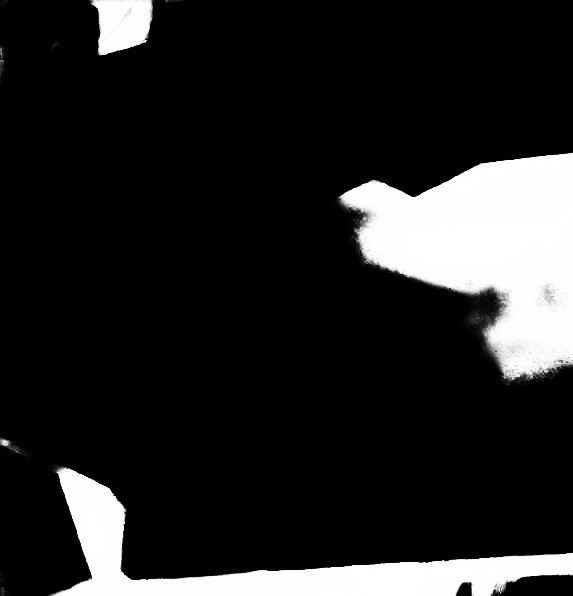}
	\end{subfigure}
	\begin{subfigure}{0.085\textwidth}
		\includegraphics[width=\textwidth]{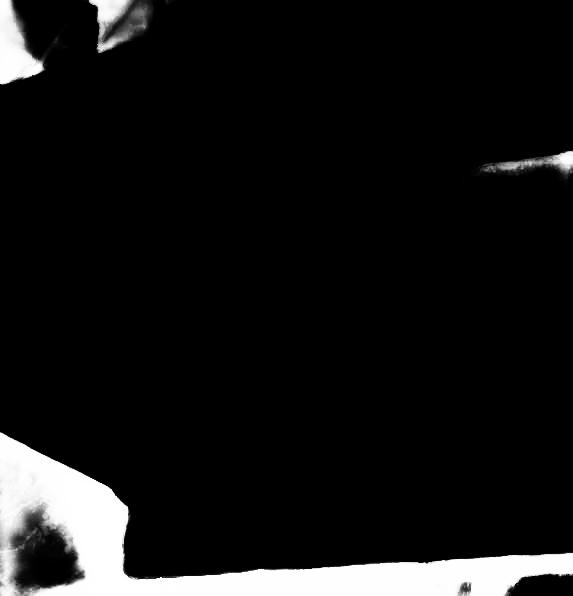}
	\end{subfigure}
	\begin{subfigure}{0.085\textwidth}
		\includegraphics[width=\textwidth]{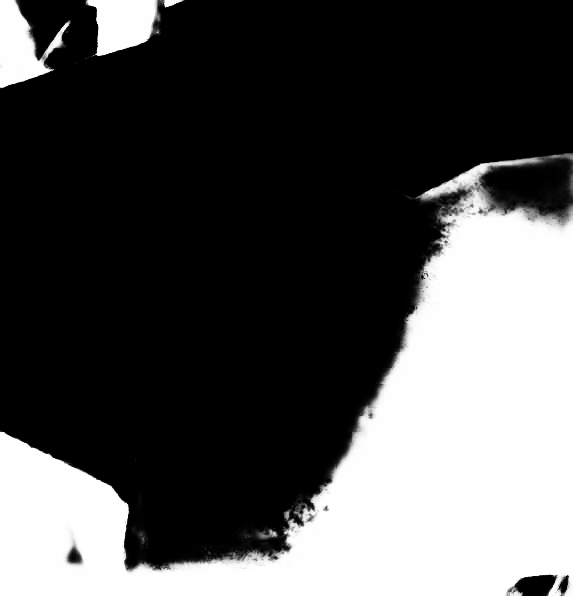}
	\end{subfigure}
	\begin{subfigure}{0.085\textwidth}
		\includegraphics[width=\textwidth]{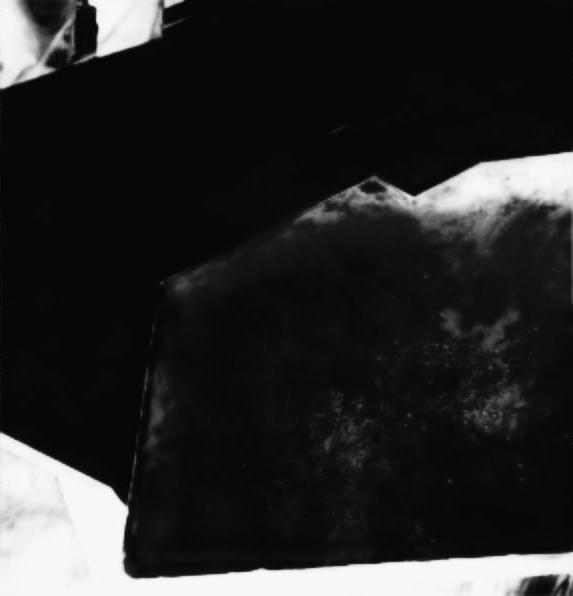}
	\end{subfigure}
	\begin{subfigure}{0.085\textwidth}
		\includegraphics[width=\textwidth]{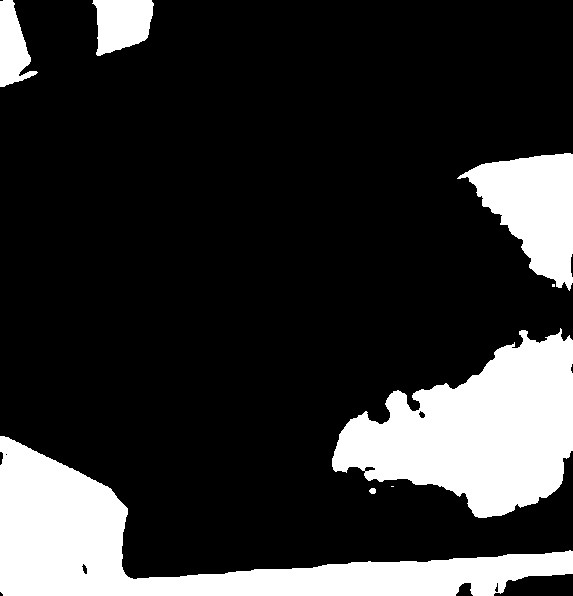}
	\end{subfigure}
	\begin{subfigure}{0.085\textwidth}
		\includegraphics[width=\textwidth]{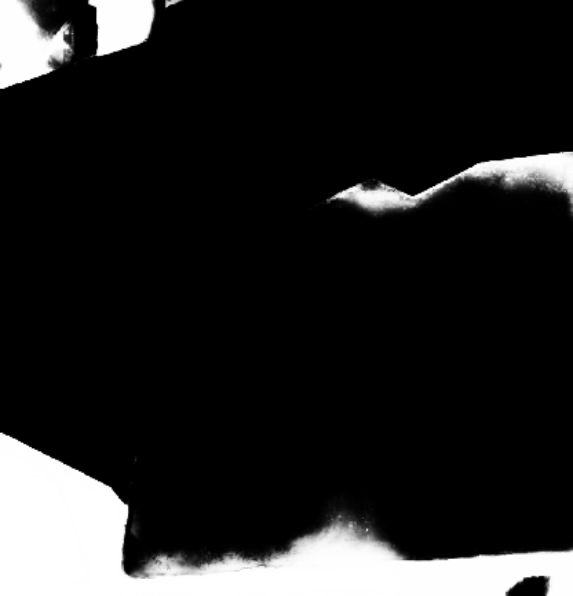}
	\end{subfigure}
	\begin{subfigure}{0.085\textwidth}
		\includegraphics[width=\textwidth]{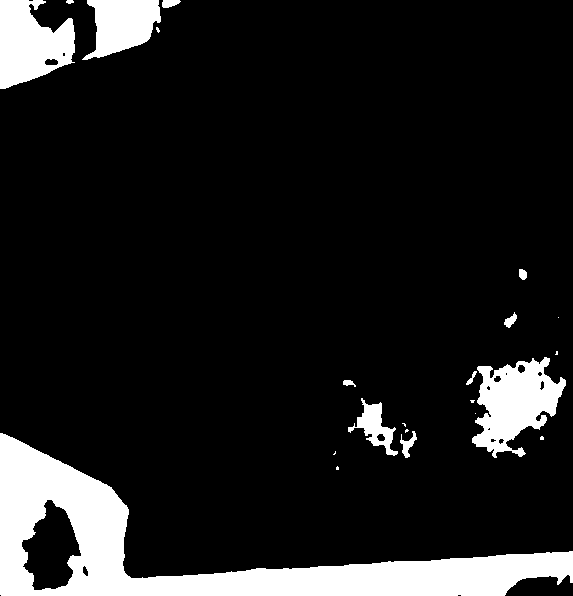}
	\end{subfigure}
	\begin{subfigure}{0.085\textwidth}
		\includegraphics[width=\textwidth]{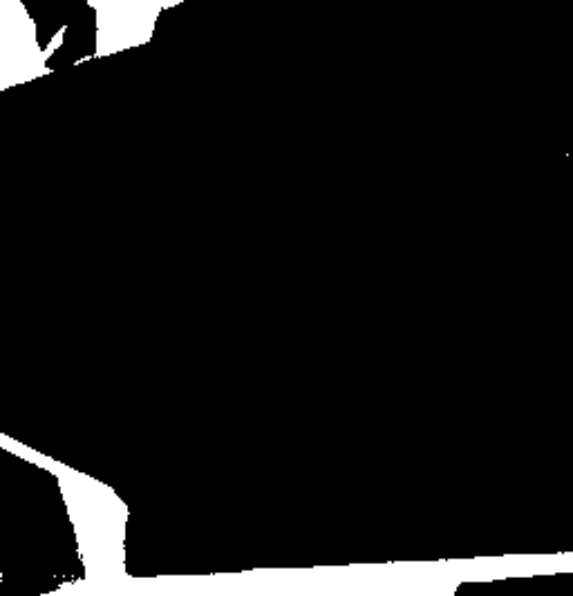}
	\end{subfigure}

	\vspace*{1.3mm}
	\begin{subfigure}{0.085\textwidth}
		\includegraphics[width=\textwidth]{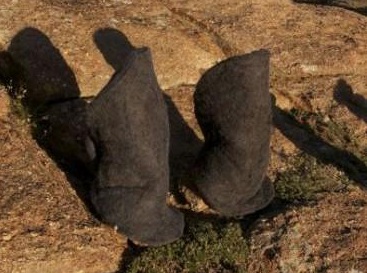}
	\end{subfigure}
	\begin{subfigure}{0.085\textwidth}
		\includegraphics[width=\textwidth]{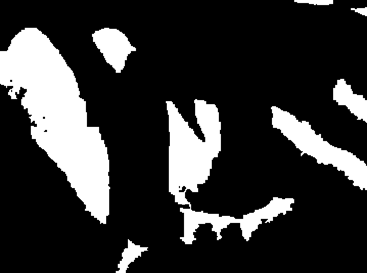}
	\end{subfigure}
	\begin{subfigure}{0.085\textwidth}
		\includegraphics[width=\textwidth]{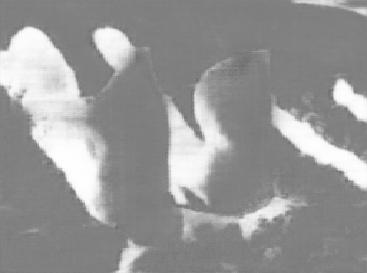}
	\end{subfigure}
	\begin{subfigure}{0.085\textwidth}
		\includegraphics[width=\textwidth]{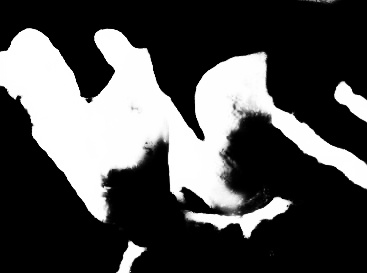}
	\end{subfigure}
	\begin{subfigure}{0.085\textwidth}
		\includegraphics[width=\textwidth]{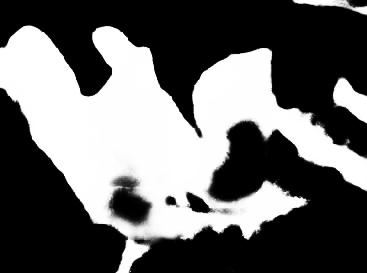}
	\end{subfigure}
	\begin{subfigure}{0.085\textwidth}
		\includegraphics[width=\textwidth]{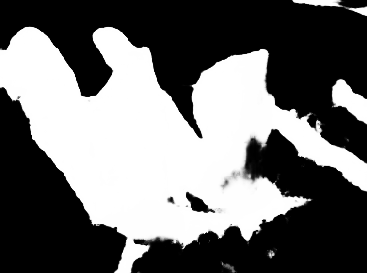}
	\end{subfigure}
	\begin{subfigure}{0.085\textwidth}
		\includegraphics[width=\textwidth]{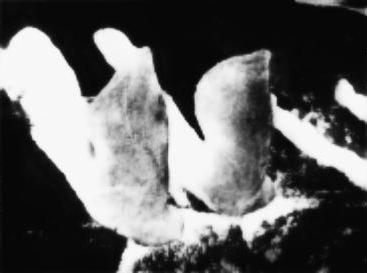}
	\end{subfigure}
	\begin{subfigure}{0.085\textwidth}
		\includegraphics[width=\textwidth]{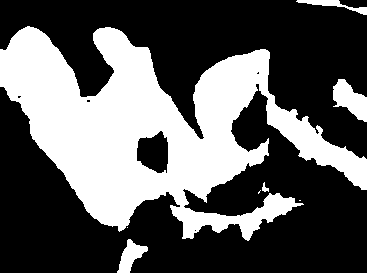}
	\end{subfigure}
	\begin{subfigure}{0.085\textwidth}
		\includegraphics[width=\textwidth]{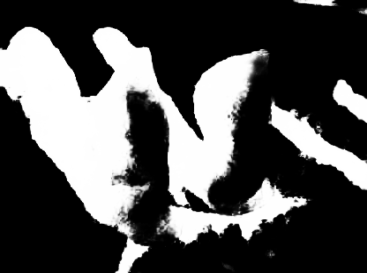}
	\end{subfigure}
	\begin{subfigure}{0.085\textwidth}
		\includegraphics[width=\textwidth]{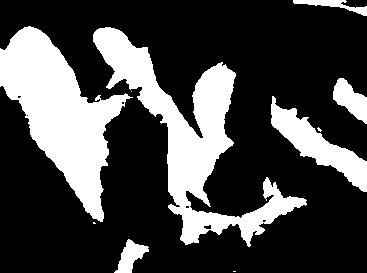}
	\end{subfigure}
	\begin{subfigure}{0.085\textwidth}
		\includegraphics[width=\textwidth]{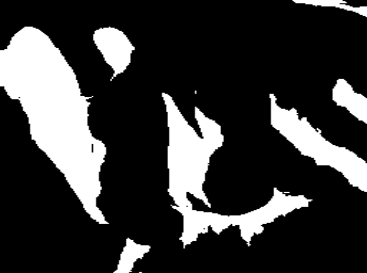}
	\end{subfigure}

	\vspace*{1.3mm}
	\begin{subfigure}{0.085\textwidth}
		\includegraphics[width=\textwidth]{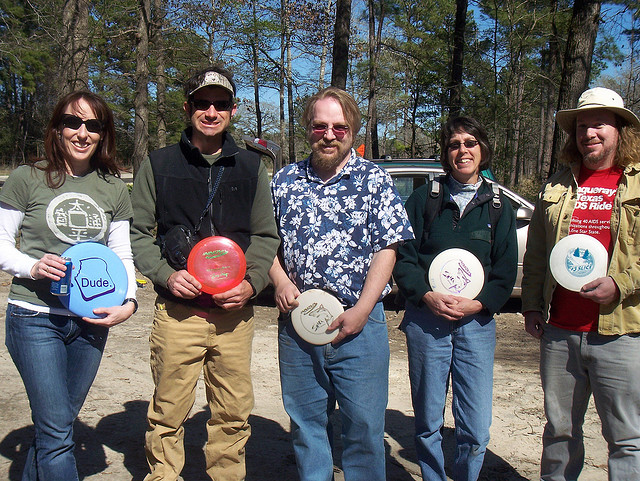}
	\end{subfigure}
	\begin{subfigure}{0.085\textwidth}
		\includegraphics[width=\textwidth]{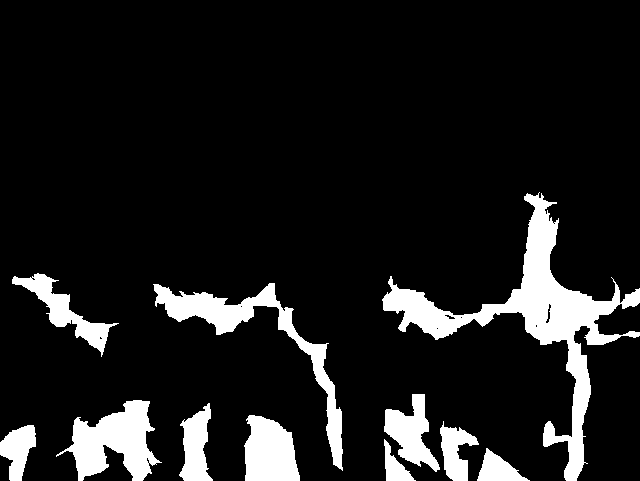}
	\end{subfigure}
	\begin{subfigure}{0.085\textwidth}
		\includegraphics[width=\textwidth]{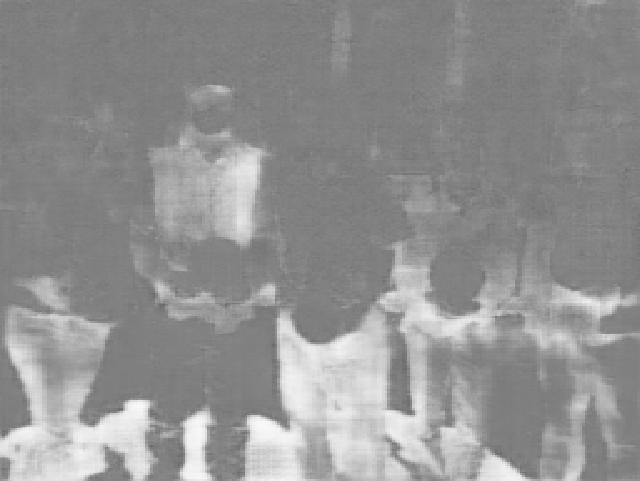}
	\end{subfigure}
	\begin{subfigure}{0.085\textwidth}
		\includegraphics[width=\textwidth]{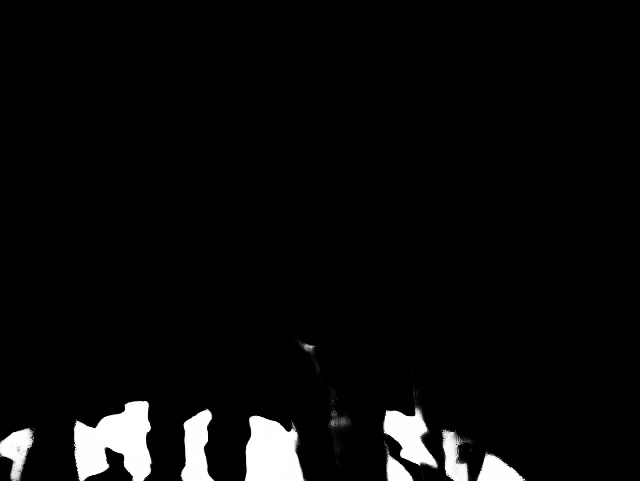}
	\end{subfigure}
	\begin{subfigure}{0.085\textwidth}
		\includegraphics[width=\textwidth]{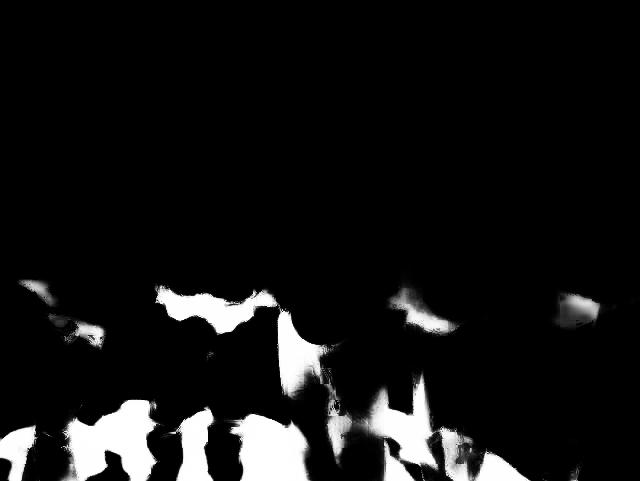}
	\end{subfigure}
	\begin{subfigure}{0.085\textwidth}
		\includegraphics[width=\textwidth]{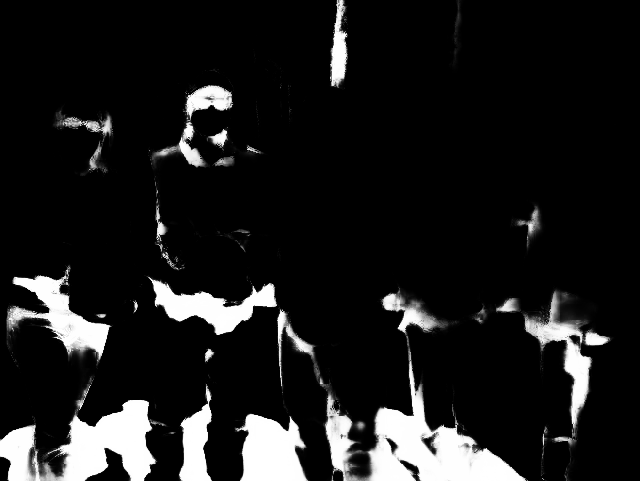}
	\end{subfigure}
	\begin{subfigure}{0.085\textwidth}
		\includegraphics[width=\textwidth]{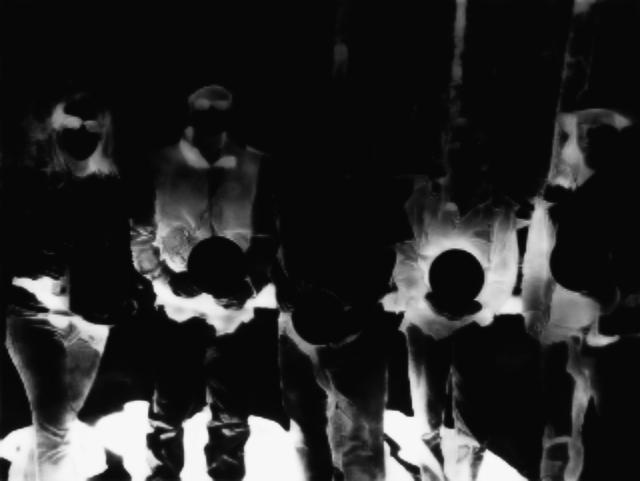}
	\end{subfigure}
	\begin{subfigure}{0.085\textwidth}
		\includegraphics[width=\textwidth]{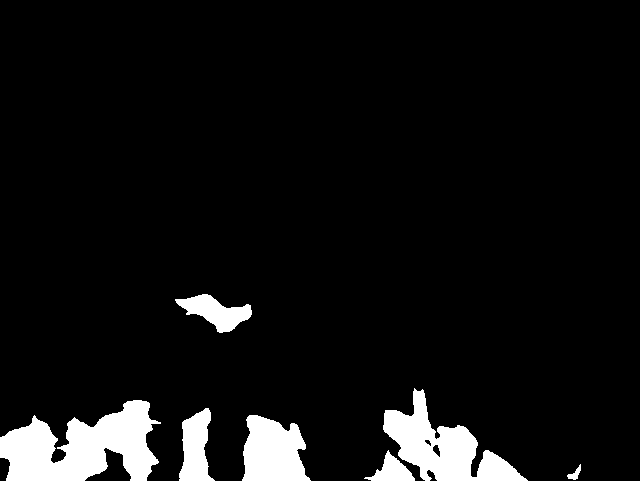}
	\end{subfigure}
	\begin{subfigure}{0.085\textwidth}
		\includegraphics[width=\textwidth]{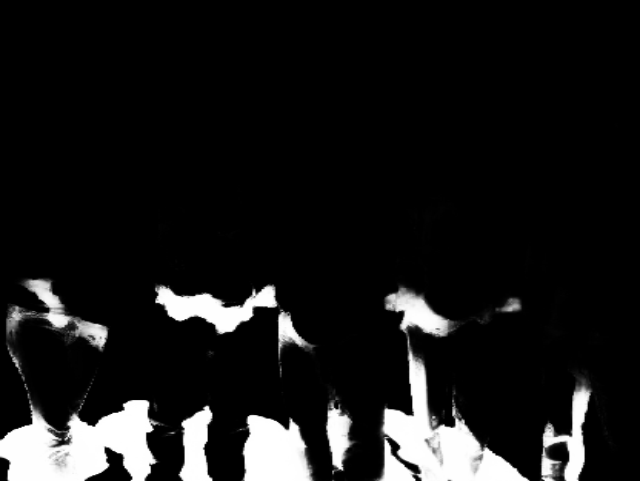}
	\end{subfigure}
	\begin{subfigure}{0.085\textwidth}
		\includegraphics[width=\textwidth]{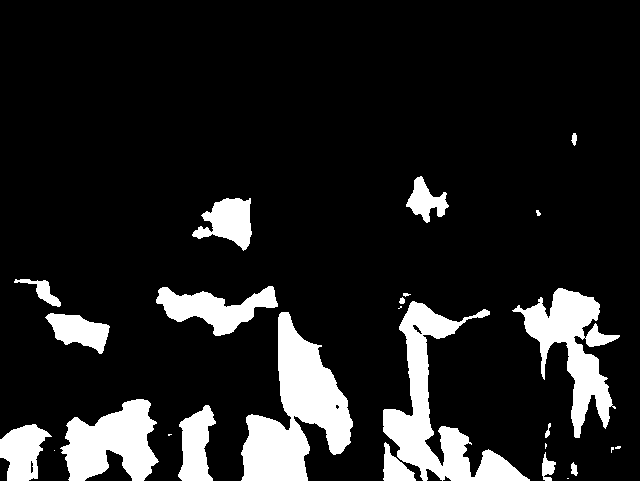}
	\end{subfigure}
	\begin{subfigure}{0.085\textwidth}
		\includegraphics[width=\textwidth]{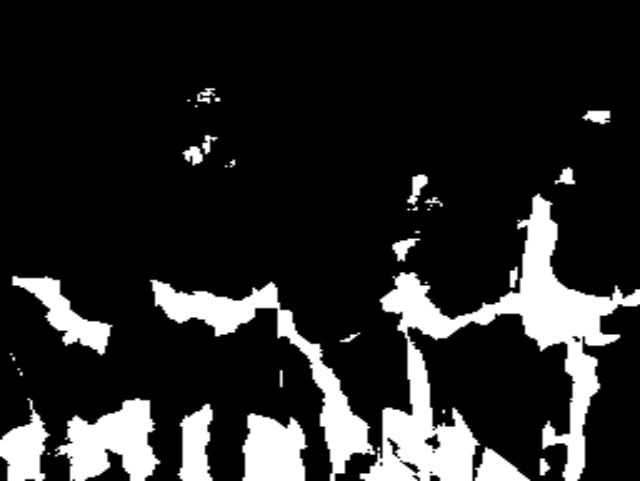}
	\end{subfigure}

	\vspace*{1.3mm}
	\begin{subfigure}{0.085\textwidth}
		\includegraphics[width=\textwidth]{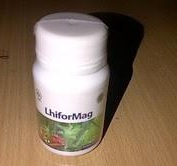}
	\end{subfigure}
	\begin{subfigure}{0.085\textwidth}
		\includegraphics[width=\textwidth]{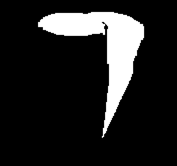}
	\end{subfigure}
	\begin{subfigure}{0.085\textwidth}
		\includegraphics[width=\textwidth]{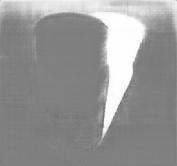}
	\end{subfigure}
	\begin{subfigure}{0.085\textwidth}
		\includegraphics[width=\textwidth]{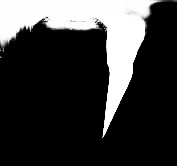}
	\end{subfigure}
	\begin{subfigure}{0.085\textwidth}
		\includegraphics[width=\textwidth]{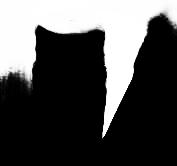}
	\end{subfigure}
	\begin{subfigure}{0.085\textwidth}
		\includegraphics[width=\textwidth]{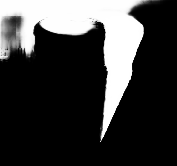}
	\end{subfigure}
	\begin{subfigure}{0.085\textwidth}
		\includegraphics[width=\textwidth]{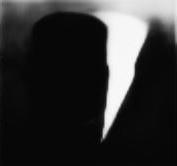}
	\end{subfigure}
	\begin{subfigure}{0.085\textwidth}
		\includegraphics[width=\textwidth]{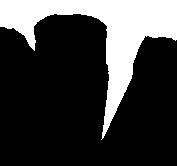}
	\end{subfigure}
	\begin{subfigure}{0.085\textwidth}
		\includegraphics[width=\textwidth]{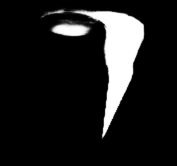}
	\end{subfigure}
	\begin{subfigure}{0.085\textwidth}
		\includegraphics[width=\textwidth]{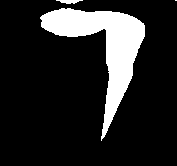}
	\end{subfigure}
	\begin{subfigure}{0.085\textwidth}
		\includegraphics[width=\textwidth]{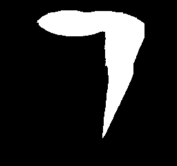}
	\end{subfigure}

	\vspace*{1.3mm}
	\begin{subfigure}{0.085\textwidth}
		\includegraphics[width=\textwidth]{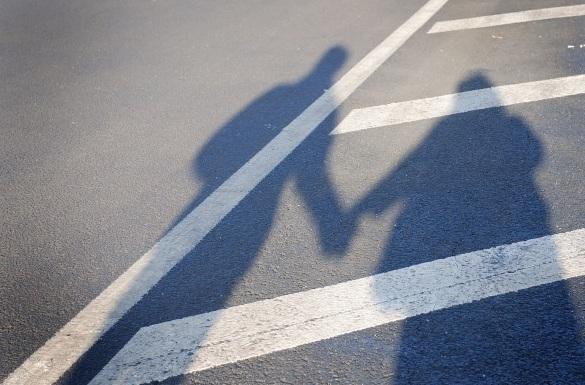}
	\end{subfigure}
	\begin{subfigure}{0.085\textwidth}
		\includegraphics[width=\textwidth]{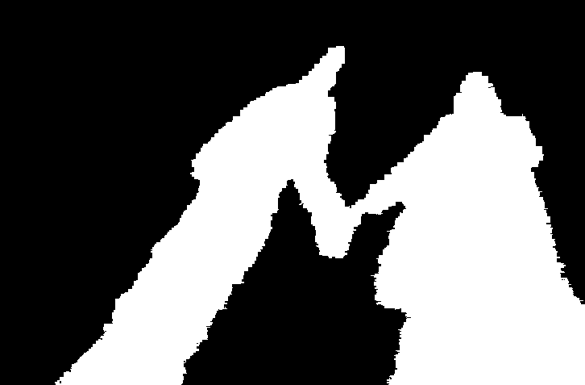}
	\end{subfigure}
	\begin{subfigure}{0.085\textwidth}
		\includegraphics[width=\textwidth]{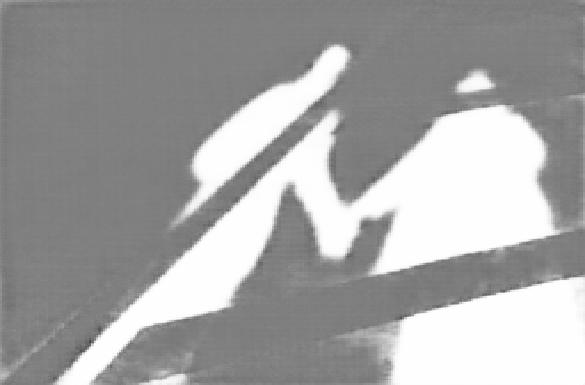}
	\end{subfigure}
	\begin{subfigure}{0.085\textwidth}
		\includegraphics[width=\textwidth]{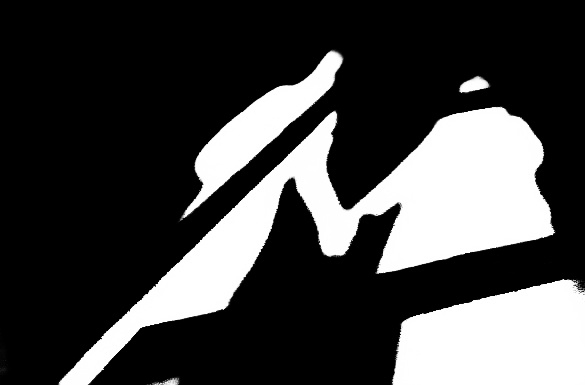}
	\end{subfigure}
	\begin{subfigure}{0.085\textwidth}
		\includegraphics[width=\textwidth]{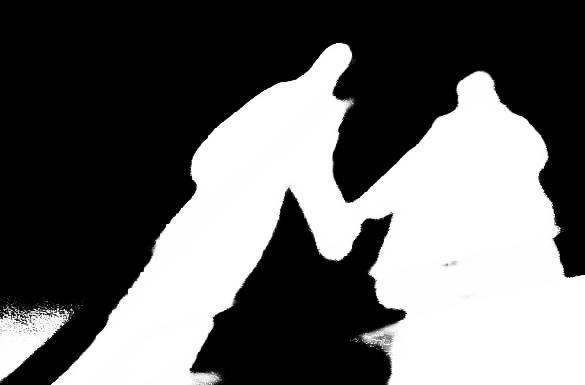}
	\end{subfigure}
	\begin{subfigure}{0.085\textwidth}
		\includegraphics[width=\textwidth]{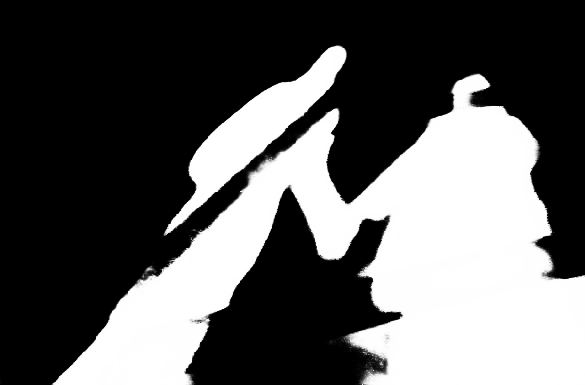}
	\end{subfigure}
	\begin{subfigure}{0.085\textwidth}
		\includegraphics[width=\textwidth]{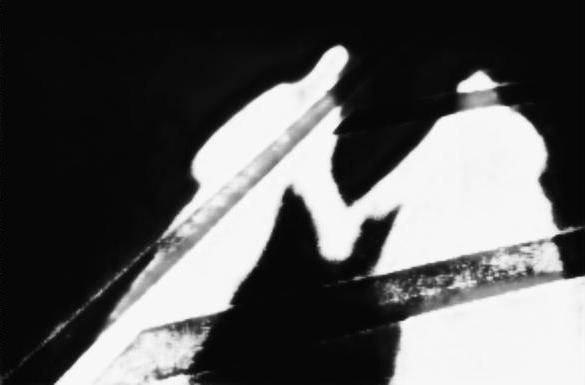}
	\end{subfigure}
	\begin{subfigure}{0.085\textwidth}
		\includegraphics[width=\textwidth]{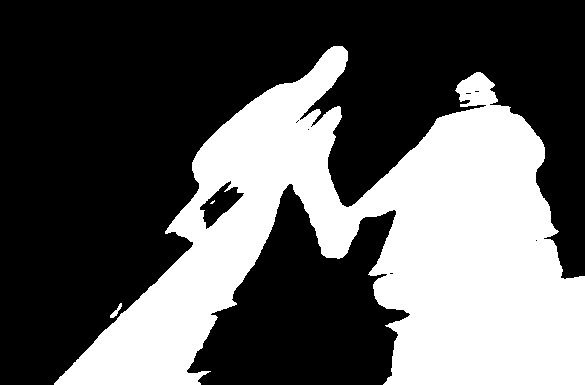}
	\end{subfigure}
	\begin{subfigure}{0.085\textwidth}
		\includegraphics[width=\textwidth]{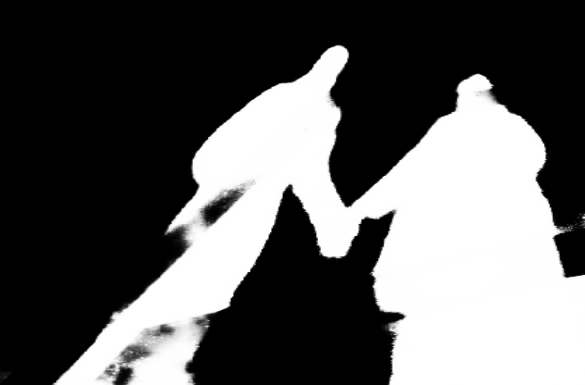}
	\end{subfigure}
	\begin{subfigure}{0.085\textwidth}
		\includegraphics[width=\textwidth]{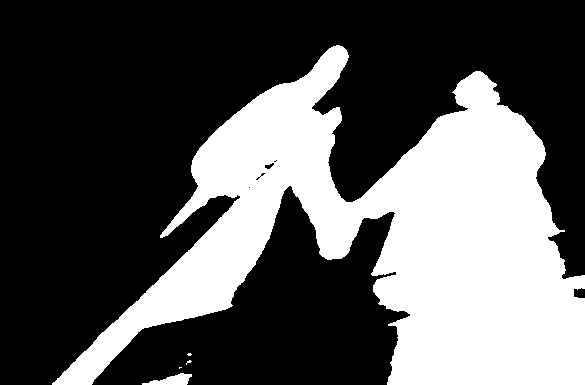}
	\end{subfigure}
	\begin{subfigure}{0.085\textwidth}
		\includegraphics[width=\textwidth]{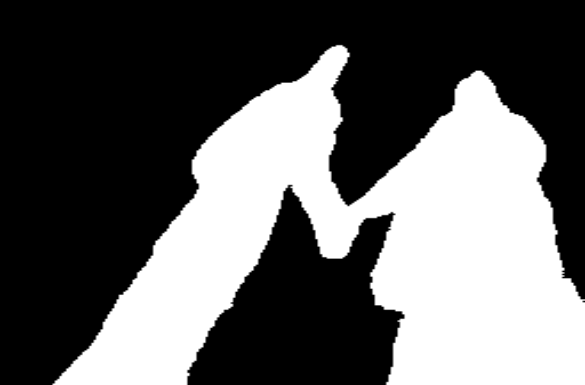}
	\end{subfigure}


	\vspace*{1.3mm}
	\begin{subfigure}{0.085\textwidth}
		\includegraphics[width=\textwidth]{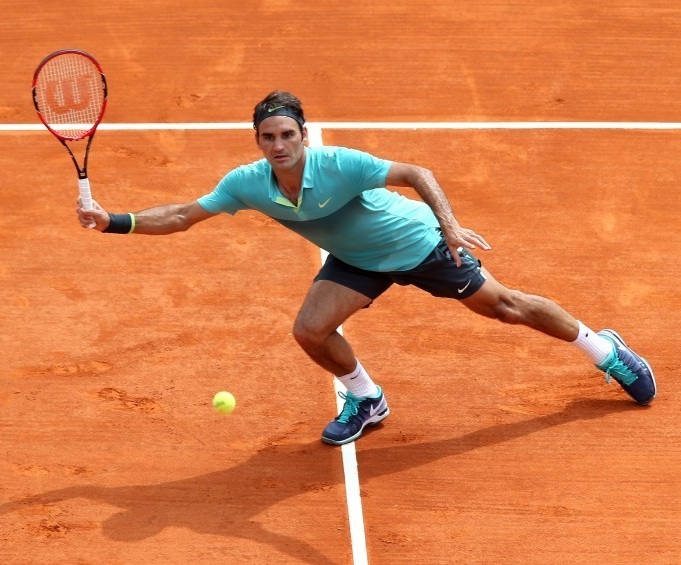}
	\end{subfigure}
	\begin{subfigure}{0.085\textwidth}
		\includegraphics[width=\textwidth]{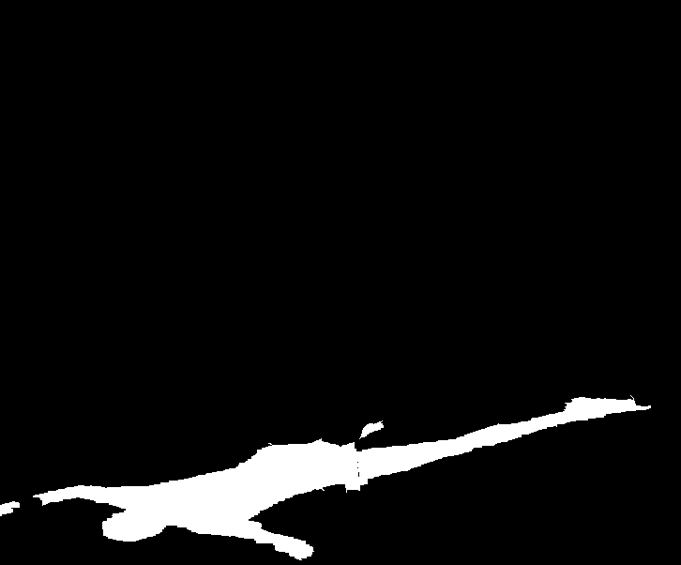}
	\end{subfigure}
	\begin{subfigure}{0.085\textwidth}
		\includegraphics[width=\textwidth]{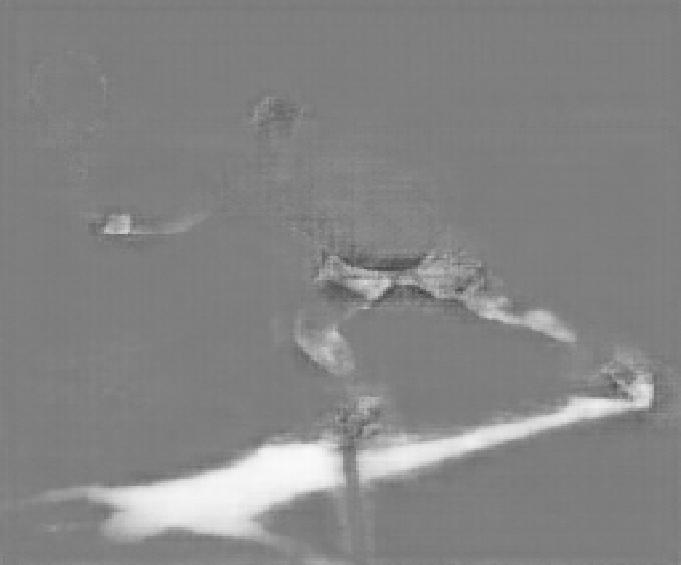}
	\end{subfigure}
	\begin{subfigure}{0.085\textwidth}
		\includegraphics[width=\textwidth]{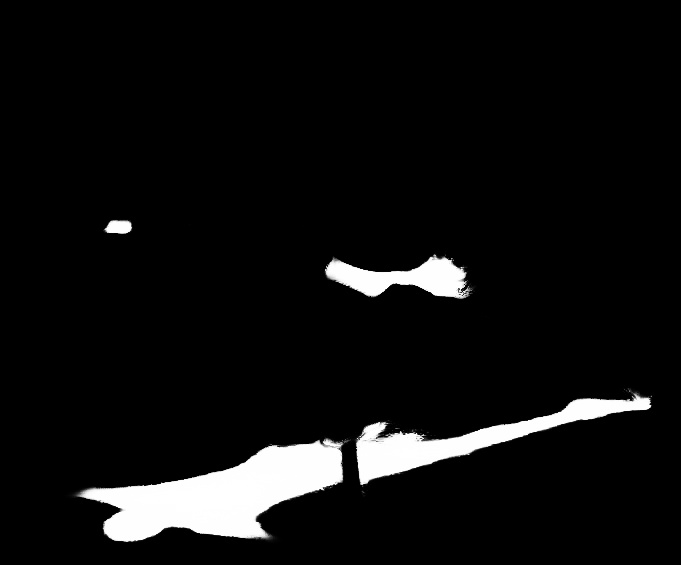}
	\end{subfigure}
	\begin{subfigure}{0.085\textwidth}
		\includegraphics[width=\textwidth]{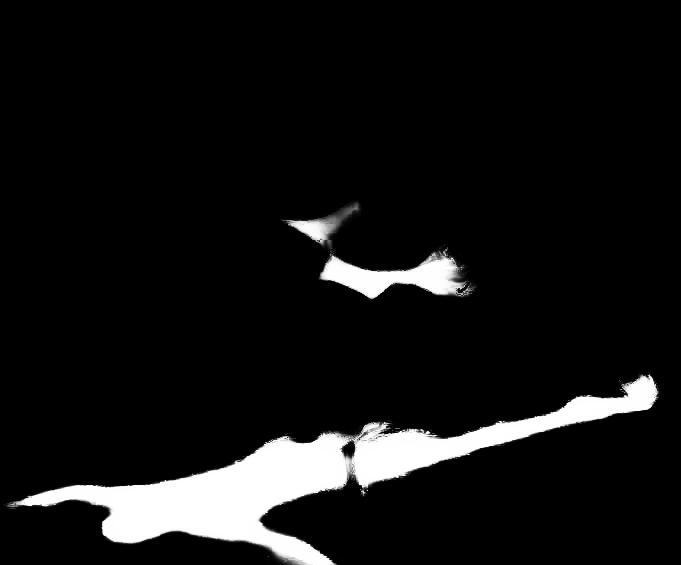}
	\end{subfigure}
	\begin{subfigure}{0.085\textwidth}
		\includegraphics[width=\textwidth]{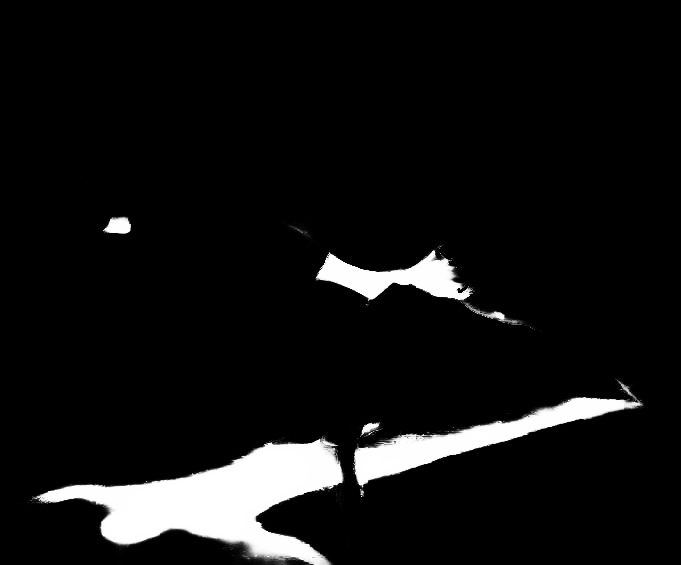}
	\end{subfigure}
	\begin{subfigure}{0.085\textwidth}
		\includegraphics[width=\textwidth]{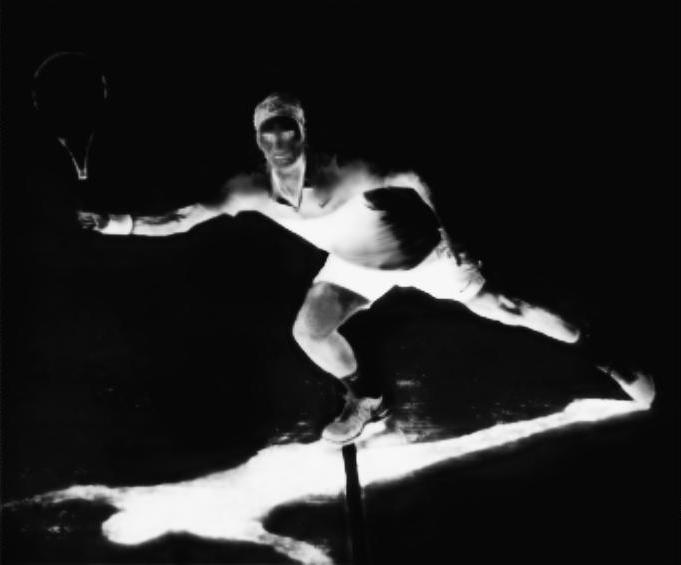}
	\end{subfigure}
	\begin{subfigure}{0.085\textwidth}
		\includegraphics[width=\textwidth]{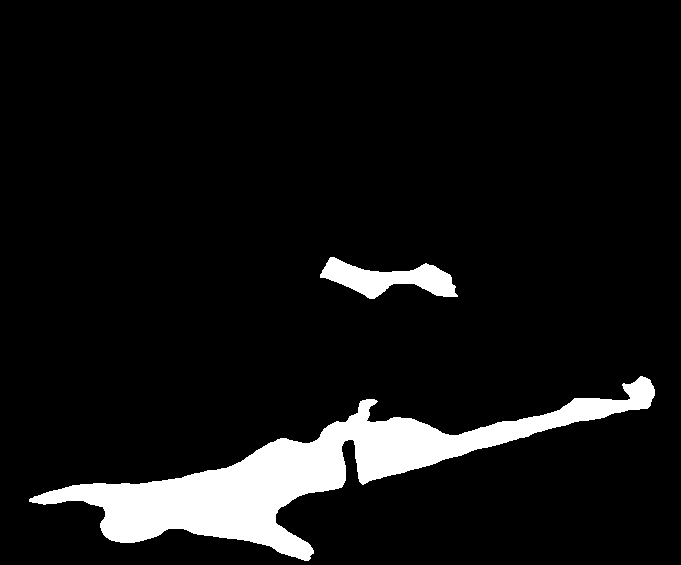}
	\end{subfigure}
	\begin{subfigure}{0.085\textwidth}
		\includegraphics[width=\textwidth]{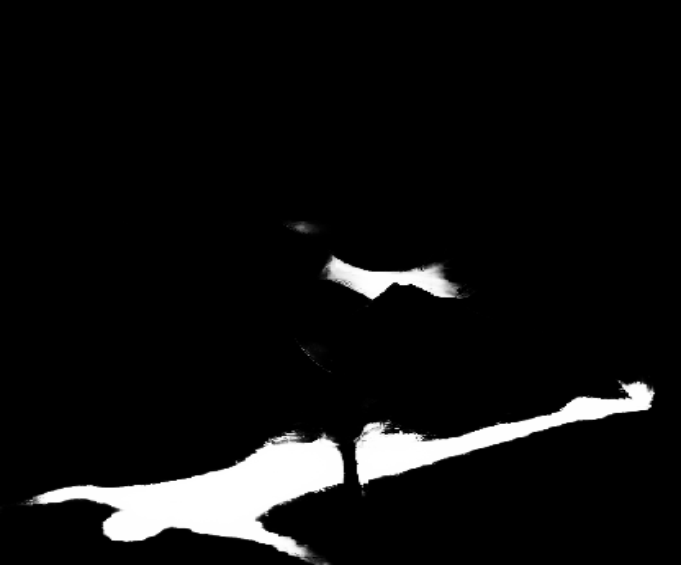}
	\end{subfigure}
	\begin{subfigure}{0.085\textwidth}
		\includegraphics[width=\textwidth]{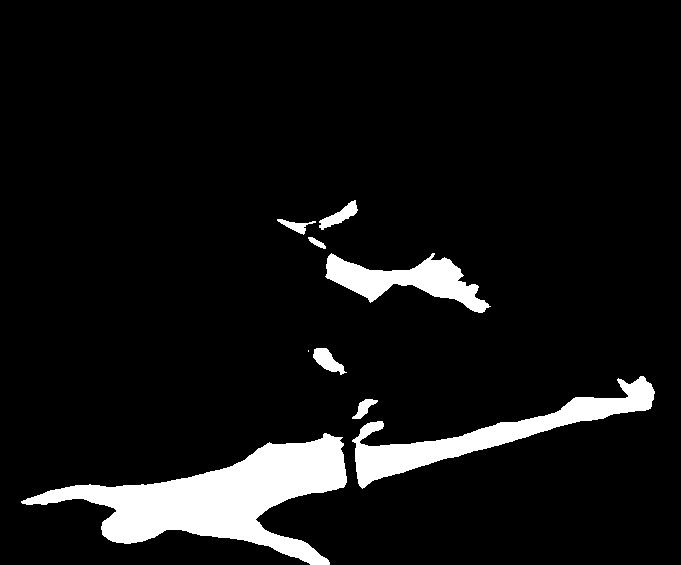}
	\end{subfigure}
	\begin{subfigure}{0.085\textwidth}
		\includegraphics[width=\textwidth]{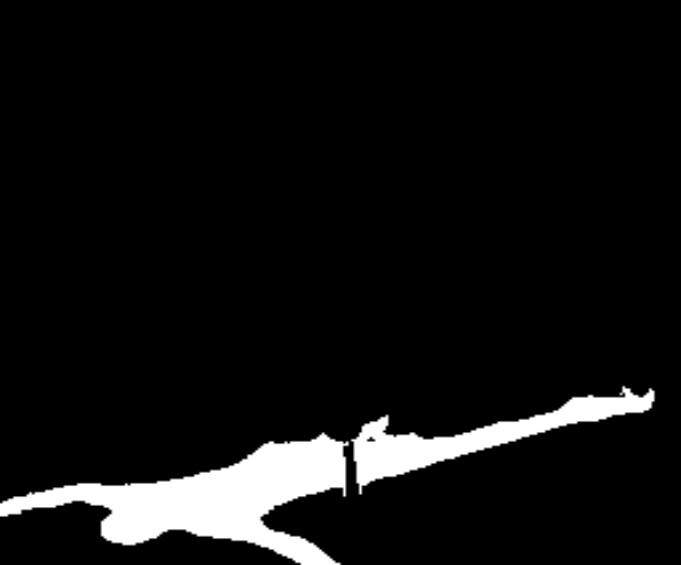}
	\end{subfigure}

	\vspace*{1.3mm}
	\begin{subfigure}{0.085\textwidth}
		\includegraphics[width=\textwidth]{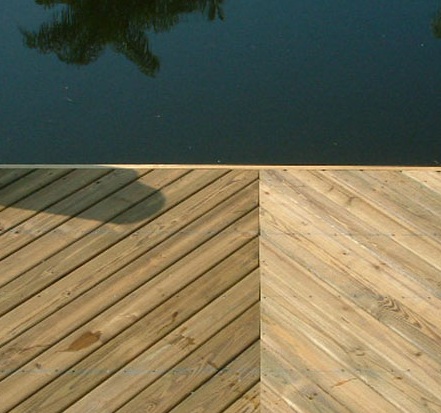}
	\end{subfigure}
	\begin{subfigure}{0.085\textwidth}
		\includegraphics[width=\textwidth]{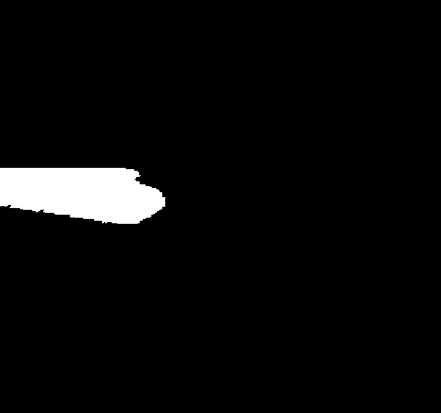}
	\end{subfigure}
	\begin{subfigure}{0.085\textwidth}
		\includegraphics[width=\textwidth]{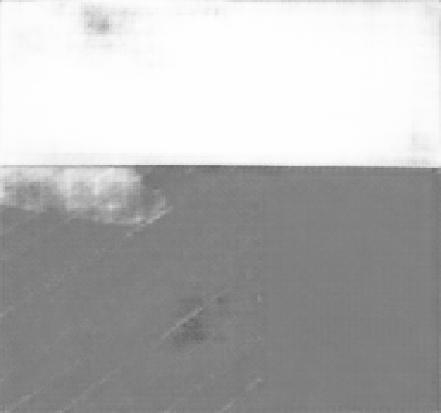}
	\end{subfigure}
	\begin{subfigure}{0.085\textwidth}
		\includegraphics[width=\textwidth]{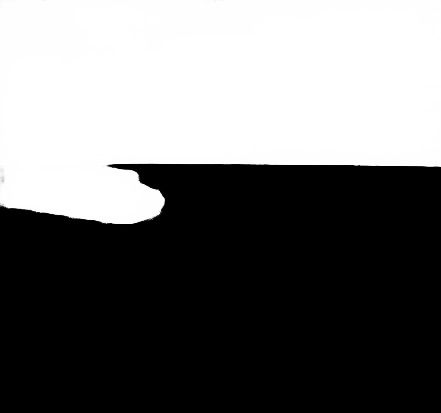}
	\end{subfigure}
	\begin{subfigure}{0.085\textwidth}
		\includegraphics[width=\textwidth]{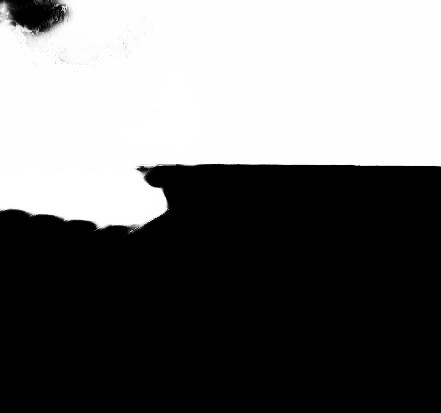}
	\end{subfigure}
	\begin{subfigure}{0.085\textwidth}
		\includegraphics[width=\textwidth]{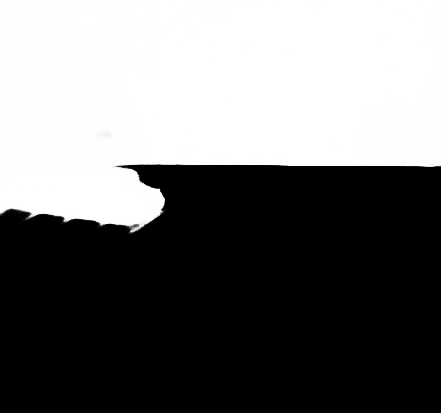}
	\end{subfigure}
	\begin{subfigure}{0.085\textwidth}
		\includegraphics[width=\textwidth]{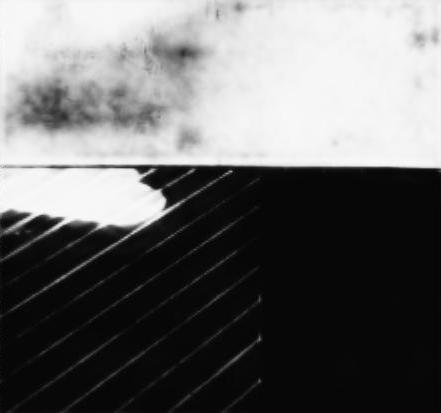}
	\end{subfigure}
	\begin{subfigure}{0.085\textwidth}
		\includegraphics[width=\textwidth]{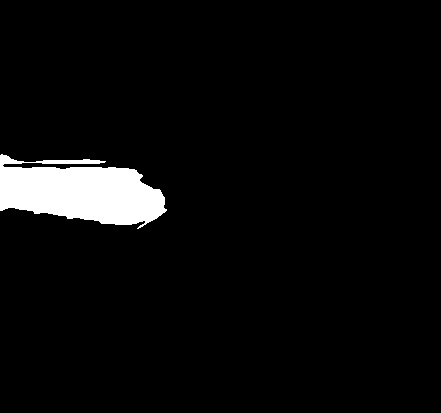}
	\end{subfigure}
	\begin{subfigure}{0.085\textwidth}
		\includegraphics[width=\textwidth]{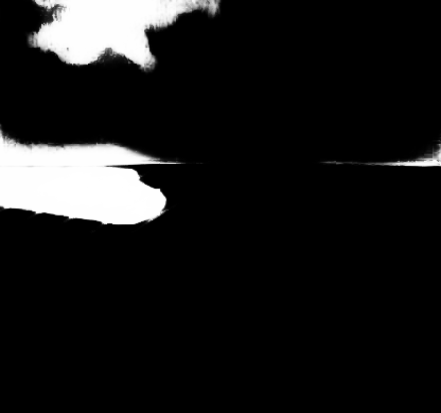}
	\end{subfigure}
	\begin{subfigure}{0.085\textwidth}
		\includegraphics[width=\textwidth]{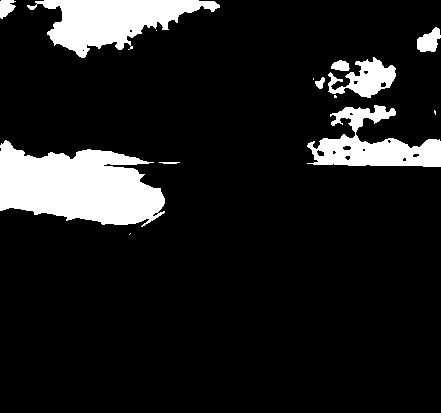}
	\end{subfigure}
	\begin{subfigure}{0.085\textwidth}
		\includegraphics[width=\textwidth]{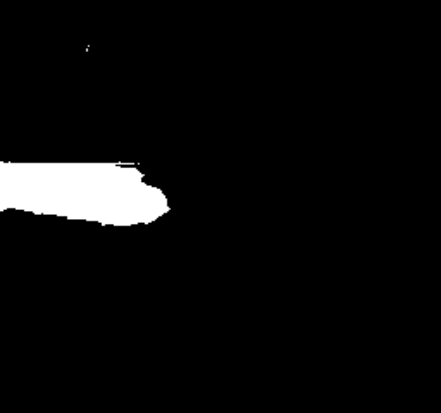}
	\end{subfigure}

	\vspace*{1.3mm}
	\begin{subfigure}{0.085\textwidth}
		\includegraphics[width=\textwidth]{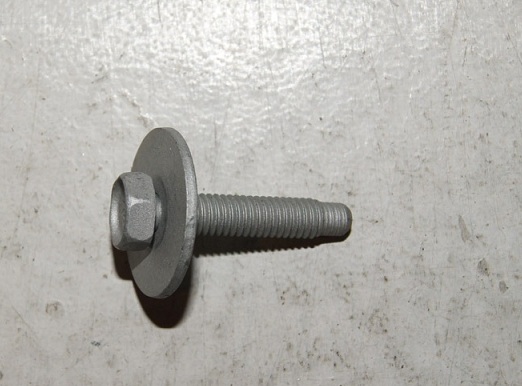}
		\captionsetup{justification=centering}
        \vspace{-5.5mm} \caption{\footnotesize{\\input \\images}}
	\end{subfigure}
	\begin{subfigure}{0.085\textwidth}
		\includegraphics[width=\textwidth]{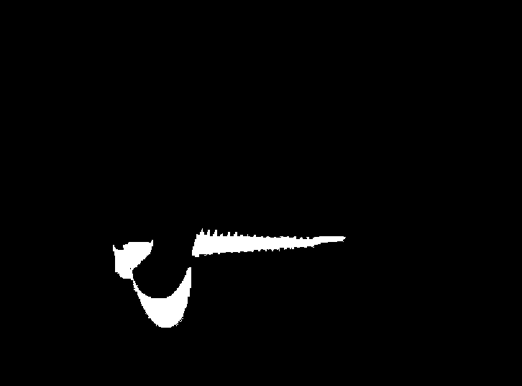}
		\captionsetup{justification=centering}
        \vspace{-5.5mm} \caption{\footnotesize{\\ground \\truths}}
	\end{subfigure}
	\begin{subfigure}{0.085\textwidth}
		\includegraphics[width=\textwidth]{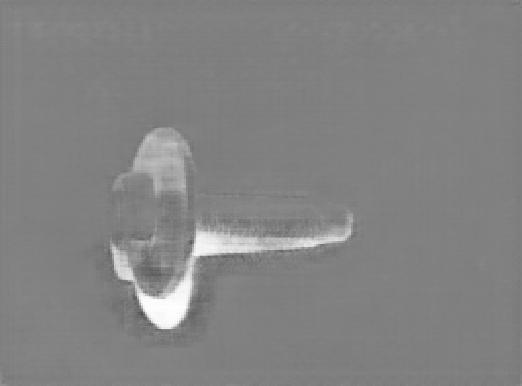}
		\captionsetup{justification=centering}
        \vspace{-5.5mm} \caption{\scriptsize{\\ADNet\\~}\footnotesize{\cite{le18_eccv}}}
	\end{subfigure}
	\begin{subfigure}{0.085\textwidth}
		\includegraphics[width=\textwidth]{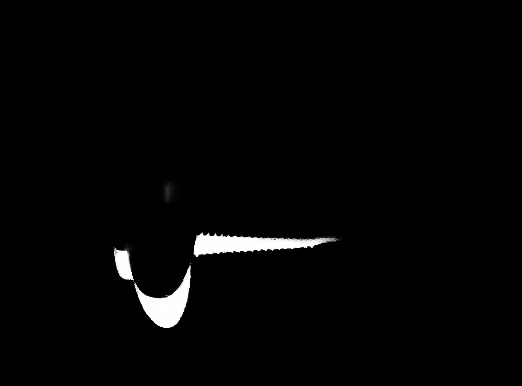}
		\captionsetup{justification=centering}
        \vspace{-5.5mm} \caption{\scriptsize{\\DSC\\~}\footnotesize{\cite{hu18_tpami}}}
	\end{subfigure}
	\begin{subfigure}{0.085\textwidth}
		\includegraphics[width=\textwidth]{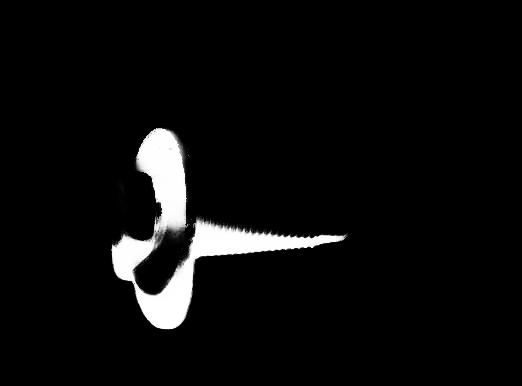}
		\captionsetup{justification=centering}
        \vspace{-5.5mm} \caption{\scriptsize{\\BDRAR\\~}\footnotesize{\cite{zhu18_eccv}}}
	\end{subfigure}
	\begin{subfigure}{0.085\textwidth}
		\includegraphics[width=\textwidth]{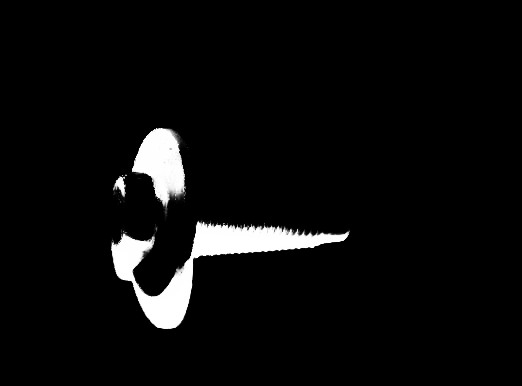}
		\captionsetup{justification=centering}
        \vspace{-5.5mm} \caption{\scriptsize{\\DSDNet\\~}\footnotesize{\cite{zheng19_cvpr}}}
	\end{subfigure}
	\begin{subfigure}{0.085\textwidth}
		\includegraphics[width=\textwidth]{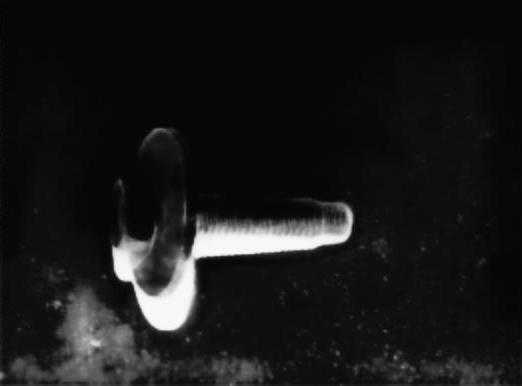}
		\captionsetup{justification=centering}
        \vspace{-5.5mm} \caption{\scriptsize{\\FDRNet\\~}\footnotesize{\cite{Zhu_2021_ICCV}}}
	\end{subfigure}
	\begin{subfigure}{0.085\textwidth}
		\includegraphics[width=\textwidth]{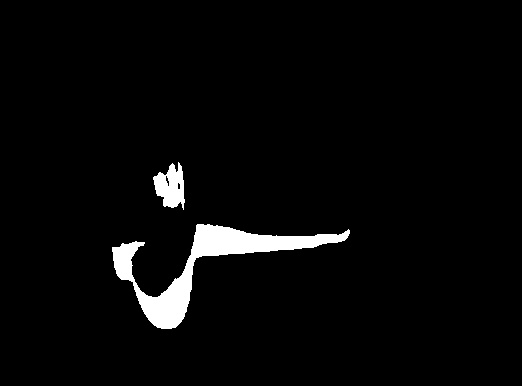}
		\captionsetup{justification=centering}
        \vspace{-5.5mm} \caption{\scriptsize{\\MTMT-Net\\~}\footnotesize{\cite{chen20_cvpr}}}
	\end{subfigure}
	\begin{subfigure}{0.085\textwidth}
		\includegraphics[width=\textwidth]{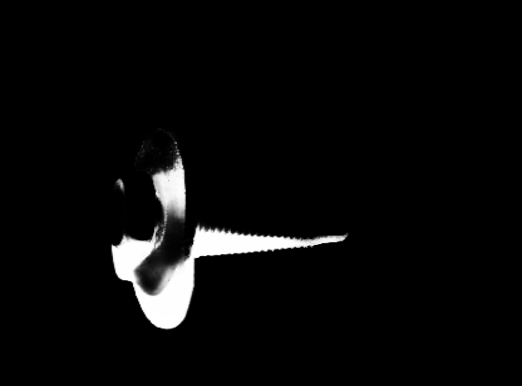}
		\captionsetup{justification=centering}
        \vspace{-5.5mm} \caption{\scriptsize{\\SDCM\\~}\footnotesize{\cite{zhu_mm2022}}}
	\end{subfigure}
	\begin{subfigure}{0.085\textwidth}
		\includegraphics[width=\textwidth]{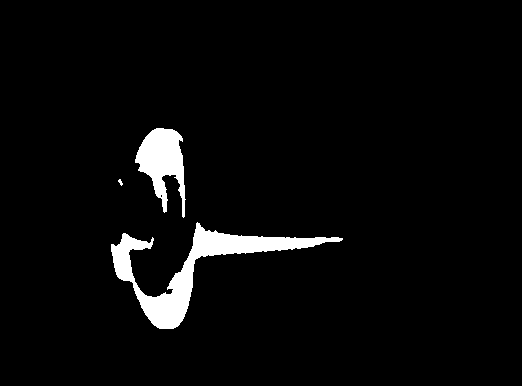}
		\captionsetup{justification=centering}
        \vspace{-5.5mm} \caption{\scriptsize{\\RMLANet\\~}\footnotesize{\cite{jie2022icme, jie2023rmlanet}}}
	\end{subfigure}
	\begin{subfigure}{0.085\textwidth}
		\includegraphics[width=\textwidth]{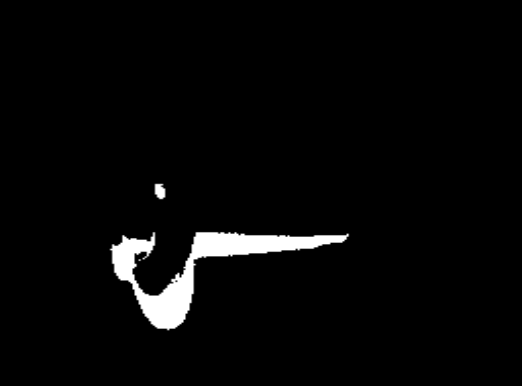}
		\captionsetup{justification=centering}
        \vspace{-5.5mm} \caption{\footnotesize{\\our\\method}}
	\end{subfigure}

	\caption{Qualitative comparison of the predicted shadow maps between our approach and other methods. From left to right: input RGB images, ground truth shadow masks, ADNet~\cite{le18_eccv}, DSC~\cite{hu18_cvpr}, BDRAR~\cite{zhu18_eccv}, DSDNet~\cite{zheng19_cvpr}, FDRNet~\cite{Zhu_2021_ICCV}, MTMT-Net~\cite{chen20_cvpr}, SDCM~\cite{zhu_mm2022}, RMLANet~\cite{jie2022icme, jie2023rmlanet} and our predictions. Best viewed on screen.}
	\label{fig_quantitative_sbu}
	
\end{figure*}
\subsection{Implementation Details}
We implement our method based on PyTorch~\cite{Paszke_PyTorch_An_Imperative_2019}. All training and testing experiments are conducted on 4 NVIDIA Tesla A100 with each has 40GB GPU memory. When generating point prompt, we employ the pretrained EfficientNet-B1~\cite{tan19_icml} as our backbone. Except the added adaptor and mask encoder, all the other parameters in SAM are frozen. The input image size for the image encoder of SAM and the pretrained backbone are $1024\times 1024$. When using top-k as our point sampling strategy, $k$ is set to different values as shown in \tabref{table_number_of_topk_points}. When grid sampling strategy is adopte, we set $k=1$ to avoid generating too many point prompts.
\\ 
\noindent\textbf{Training.} Our model is trained by 40, 200 and 200 epochs for the SBU, ISTD and CUHK dataset, respectively. We use Adam as our optimizer with a learning rate $0.0001$, beta1 0.9 and beta2 0.999, respectively. Only random crop and horizontal flipping is applied for data augmentation.
\\
\noindent\textbf{Testing.} For fair comparison, we do not apply any post-processing operations such as conditional random field (CRF) and any data augmentation. The reason is that the improvements brought about by post-processing operations vary significantly among different methods.
\subsection{Comparison with the State-of-the-art Methods}
We compare our approach with one traditional methods, namely Unary-Pairwise~\cite{guo11_cvpr}, and 18 deep learning based shadow detection methods which are stacked-CNN~\cite{sbu_dataset}, scGAN~\cite{nguyen17_iccv}, patched-CNN~\cite{hs18_iros}, ST-CGAN~\cite{istd_dataset}, DSC~\cite{hu18_cvpr}, ADNet~\cite{le18_eccv}, BDRAR~\cite{zhu18_eccv}, DC-DSPF~\cite{wang18_ijcai}, DSDNet~\cite{zheng19_cvpr}, MTMT-Net~\cite{chen20_cvpr}, RCMPNet~\cite{liao2021mm}, FDRNet~\cite{Zhu_2021_ICCV}, SDCM ~\cite{zhu_mm2022}, TranShadow~\cite{jie2022icassp}, FCSD-Net~\cite{jose_wacv2023}, RMLANet~\cite{jie2022icme, jie2023rmlanet}, SDDNet~\cite{cong2023sddnet} and SARA~\cite{Sun_2023_ICCV}, both qualitatively and quantitatively. For fair comparison, all the predicted shadow masks or $BER$ values of other methods are directly adopted from their paper or obtained using their official code.

\noindent\textbf{Quantitative Comparison.} In \tabref{table_detection_quantitative}, the qualitative results of our method with other methods are presented. Specifically, our method surpasses Unary-Pairwise~\cite{guo11_cvpr}, stacked-CNN~\cite{sbu_dataset}, scGAN~\cite{nguyen17_iccv}, patched-CNN~\cite{hs18_iros}, ST-CGAN~\cite{istd_dataset}, DSC~\cite{hu18_cvpr}, ADNet~\cite{le18_eccv}, BDRAR~\cite{zhu18_eccv}, DC-DSPF~\cite{wang18_ijcai}, DSDNet~\cite{zheng19_cvpr}, MTMT-Net~\cite{chen20_cvpr}, RCMPNet~\cite{liao2021mm}, FDRNet~\cite{Zhu_2021_ICCV}, SDCM ~\cite{zhu_mm2022}, TranShadow~\cite{jie2022icassp}, FCSD-Net~\cite{jose_wacv2023}, RMLANet~\cite{jie2022icme, jie2023rmlanet}, SDDNet~\cite{cong2023sddnet} and SARA~\cite{Sun_2023_ICCV} by $89.01\%$, $75.00\%$, $69.78\%$, $76.21\%$, $66.22\%$, $50.81\%$, $48.79\%$, $24.45\%$, $43.88\%$, $20.29\%$, $12.70\%$, $12.14\%$, $9.54\%$, $8.94\%$, $13.25\%$, $12.70\%$, $7.41\%$, $6.46\%$ and $4.18\%$ respectively on the SBU dataset. More importantly, our method demonstrates the best generalization ability in terms of the performance on the UCF dataset, when directly evaluated the performance on the UCF dataset using the model trained on the SBU dataset. In particular, our method outperforms stacked-CNN~\cite{sbu_dataset}, scGAN~\cite{nguyen17_iccv}, ST-CGAN~\cite{istd_dataset}, DSC~\cite{hu18_cvpr}, ADNet~\cite{le18_eccv}, BDRAR~\cite{zhu18_eccv}, DC-DSPF~\cite{wang18_ijcai}, DSDNet~\cite{zheng19_cvpr}, MTMT-Net~\cite{chen20_cvpr}, RCMPNet~\cite{liao2021mm}, FDRNet~\cite{Zhu_2021_ICCV}, SDCM ~\cite{zhu_mm2022}, TranShadow~\cite{jie2022icassp}, FCSD-Net~\cite{jose_wacv2023}, RMLANet~\cite{jie2022icme, jie2023rmlanet}, SDDNet~\cite{cong2023sddnet} and SARA~\cite{Sun_2023_ICCV} by $51.15\%$, $44.78\%$, $43.46\%$, $39.75\%$, $31.35\%$, $18.69\%$, $19.62\%$, $16.34\%$, $14.99\%$, $5.37\%$, $12.77\%$, $5.08\%$, $8.63\%$, $8.76\%$, $0.94\%$, $3.64\%$ and $9.42\%$ respectively. Moreover, our method also achieves the best performance on the ISTD dataset, by outdistancing stacked-CNN~\cite{sbu_dataset}, scGAN~\cite{nguyen17_iccv}, patched-CNN~\cite{hs18_iros}, DSC~\cite{hu18_cvpr}, ADNet~\cite{le18_eccv}, BDRAR~\cite{zhu18_eccv}, DSDNet~\cite{zheng19_cvpr}, MTMT-Net~\cite{chen20_cvpr}, RCMPNet~\cite{liao2021mm}, FDRNet~\cite{Zhu_2021_ICCV}, SDCM ~\cite{zhu_mm2022}, TranShadow~\cite{jie2022icassp}, FCSD-Net~\cite{jose_wacv2023}, RMLANet~\cite{jie2022icme, jie2023rmlanet}, SDDNet~\cite{cong2023sddnet} and SARA~\cite{Sun_2023_ICCV} by $90.00\%$, $81.70\%$, $77.66\%$, $74.85\%$, $68.03\%$, $60.37\%$, $50.00\%$, $46.58\%$, $44.52\%$, $39.01\%$, $50.29\%$, $49.11\%$, $14.85\%$, $32.28\%$ and $27.12\%$ respectively. 

To further demonstrate the effectiveness of our method, quantitative results on $SBUTestNew$ and CUHK dataset are presented in \tabref{table_detection_sbu_new} and \tabref{table_detection_cuhk}. As shown, our method also achieves the best performance in terms of $BER$ value. Specifically, our method outperforms SILT~\cite{Yang_2023_ICCV} which was elaborately designed to handle annotation errors on the $SBUTestNew$ dataset. On the CUHK dataset, we achieved the best performance with only a quarter of the parameters except FDSNET~\cite{gy_tip2021}. However, our method outperforms FDSNET by $9\%$.

\begin{figure*}
	\centering
	\vspace*{1.0mm}
	\begin{subfigure}{0.1\textwidth}
        \centering
		input images
	\end{subfigure}
	\begin{subfigure}{0.1\textwidth}
		\includegraphics[width=\textwidth]{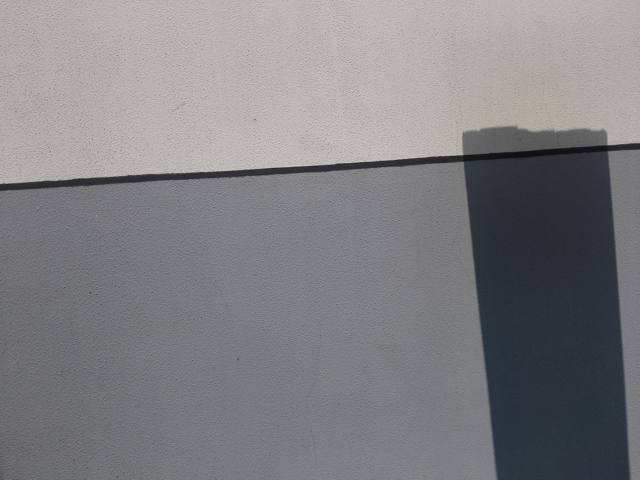}
	\end{subfigure}
	\begin{subfigure}{0.1\textwidth}
		\includegraphics[width=\textwidth]{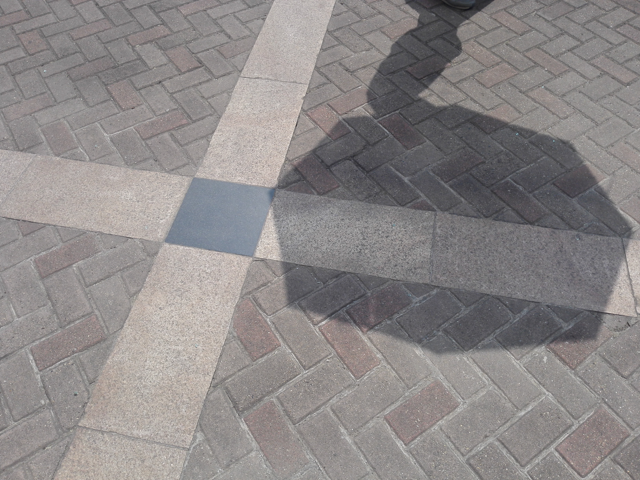}
	\end{subfigure}
	\begin{subfigure}{0.1\textwidth}
		\includegraphics[width=\textwidth]{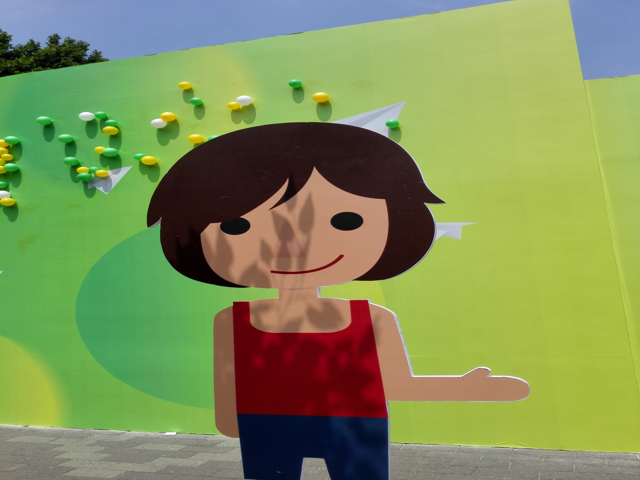}
	\end{subfigure}
	\begin{subfigure}{0.1\textwidth}
		\includegraphics[width=\textwidth]{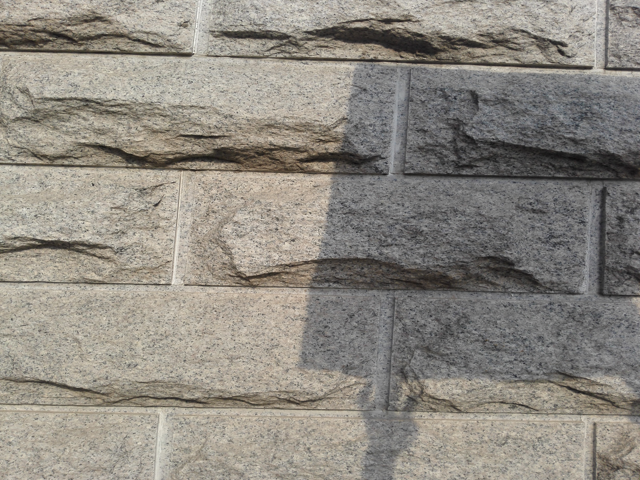}
	\end{subfigure}
	\begin{subfigure}{0.1\textwidth}
		\includegraphics[width=\textwidth]{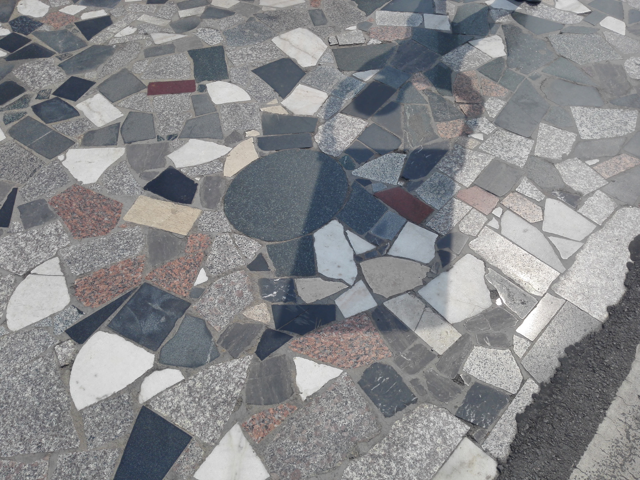}
	\end{subfigure}
	\begin{subfigure}{0.1\textwidth}
		\includegraphics[width=\textwidth]{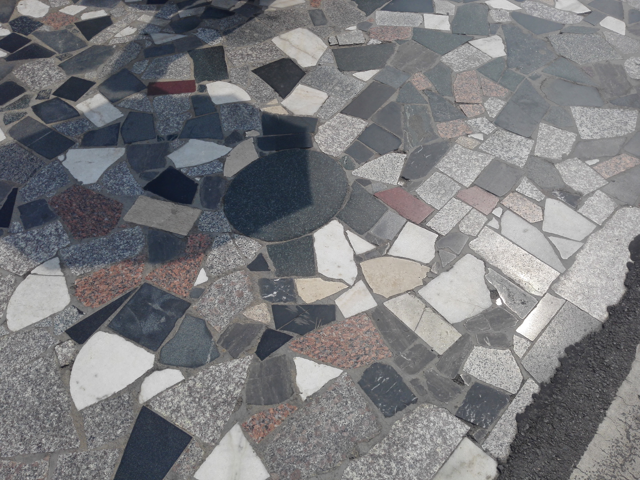}
	\end{subfigure}
	\begin{subfigure}{0.1\textwidth}
		\includegraphics[width=\textwidth]{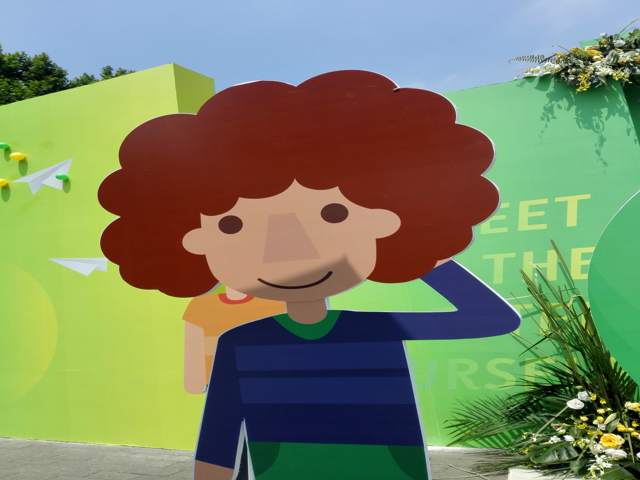}
	\end{subfigure}
	\begin{subfigure}{0.1\textwidth}
		\includegraphics[width=\textwidth]{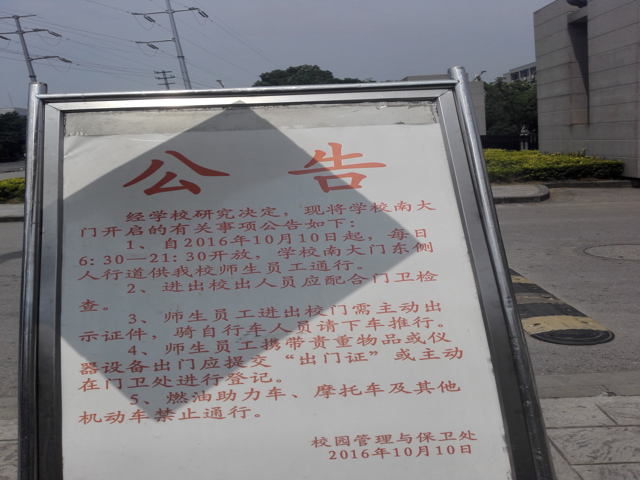}
	\end{subfigure}

	\vspace*{1.0mm}
	\begin{subfigure}{0.1\textwidth}
        \centering
		ground truths
	\end{subfigure}
	\begin{subfigure}{0.1\textwidth}
		\includegraphics[width=\textwidth]{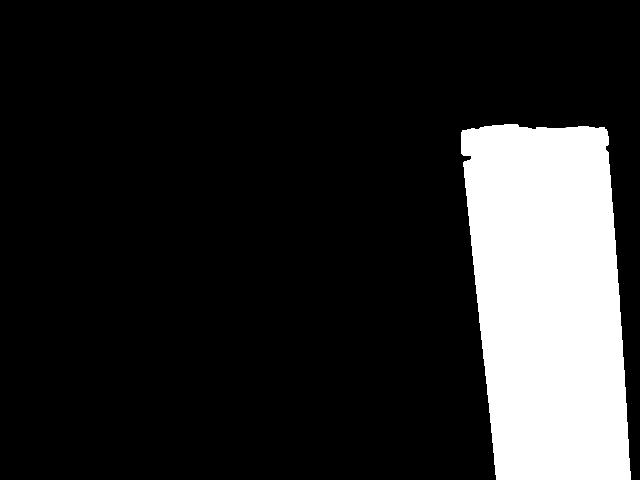}
	\end{subfigure}
	\begin{subfigure}{0.1\textwidth}
		\includegraphics[width=\textwidth]{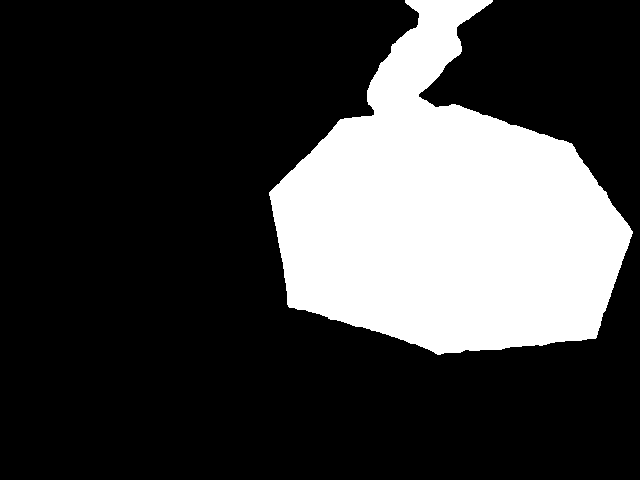}
	\end{subfigure}
	\begin{subfigure}{0.1\textwidth}
		\includegraphics[width=\textwidth]{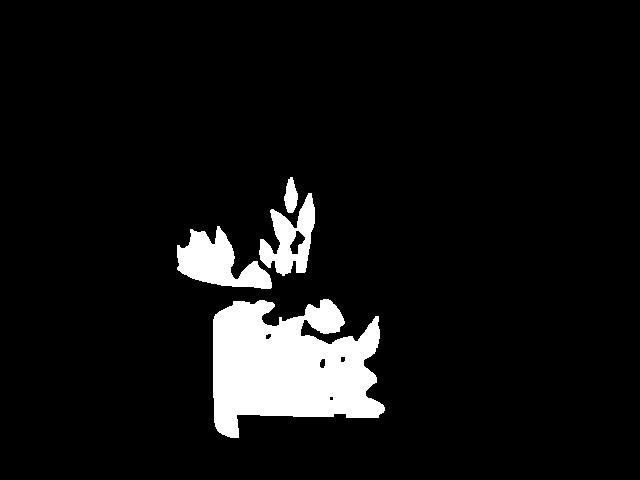}
	\end{subfigure}
	\begin{subfigure}{0.1\textwidth}
		\includegraphics[width=\textwidth]{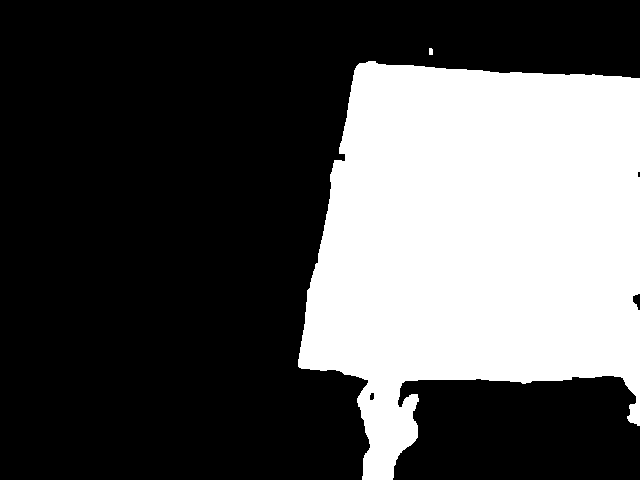}
	\end{subfigure}
	\begin{subfigure}{0.1\textwidth}
		\includegraphics[width=\textwidth]{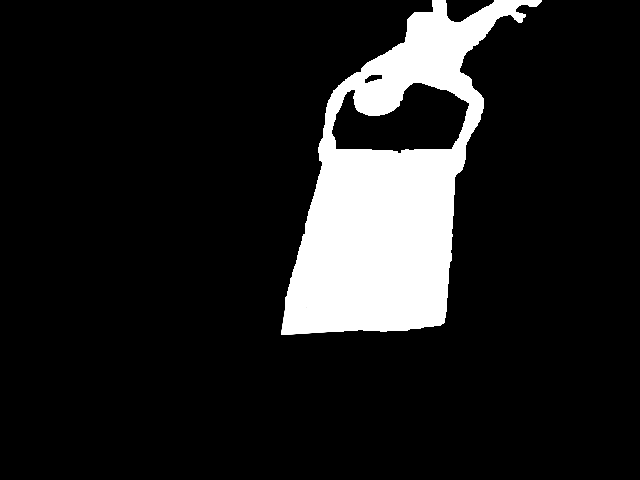}
	\end{subfigure}
	\begin{subfigure}{0.1\textwidth}
		\includegraphics[width=\textwidth]{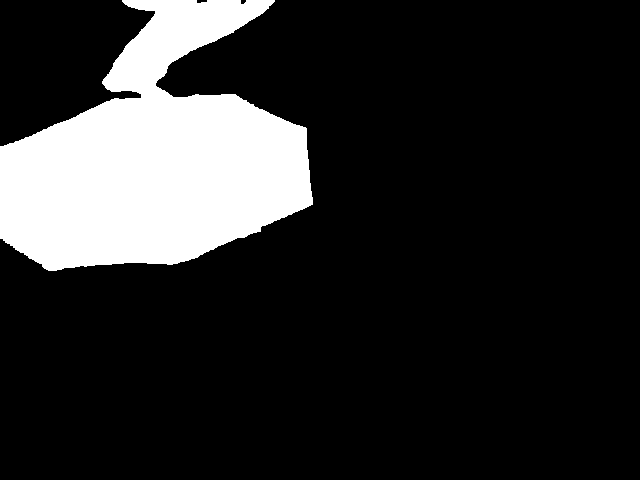}
	\end{subfigure}
	\begin{subfigure}{0.1\textwidth}
		\includegraphics[width=\textwidth]{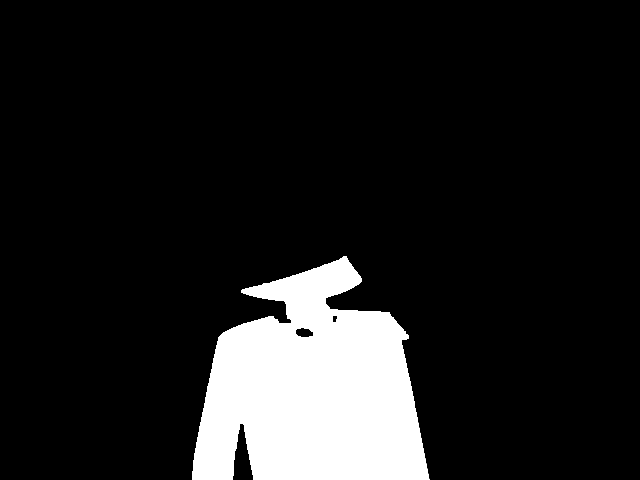}
	\end{subfigure}
	\begin{subfigure}{0.1\textwidth}
		\includegraphics[width=\textwidth]{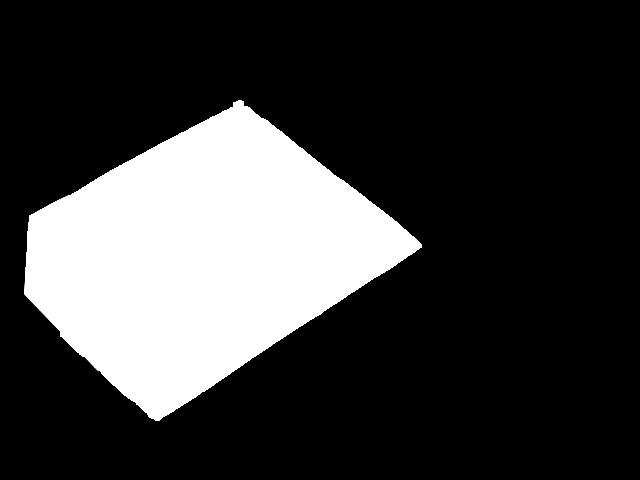}
	\end{subfigure}
    
	\vspace*{1.0mm}
	\begin{subfigure}{0.1\textwidth}
        \centering
		our method
	\end{subfigure}
	\begin{subfigure}{0.1\textwidth}
		\includegraphics[width=\textwidth]{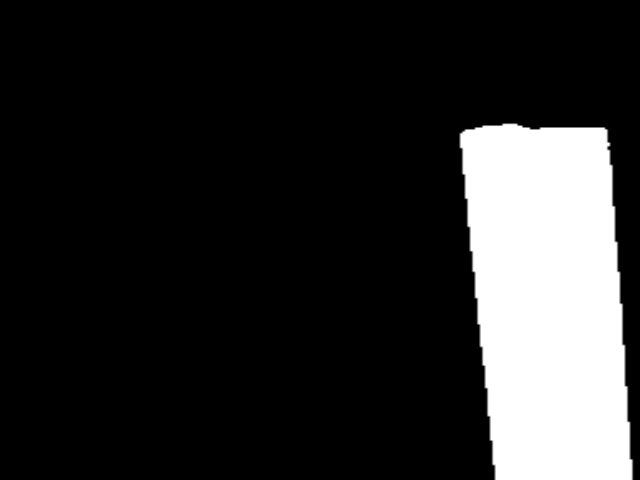}
	\end{subfigure}
	\begin{subfigure}{0.1\textwidth}
		\includegraphics[width=\textwidth]{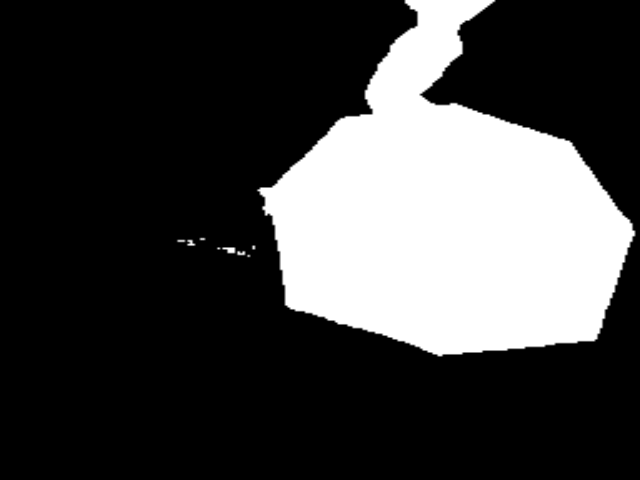}
	\end{subfigure}
	\begin{subfigure}{0.1\textwidth}
		\includegraphics[width=\textwidth]{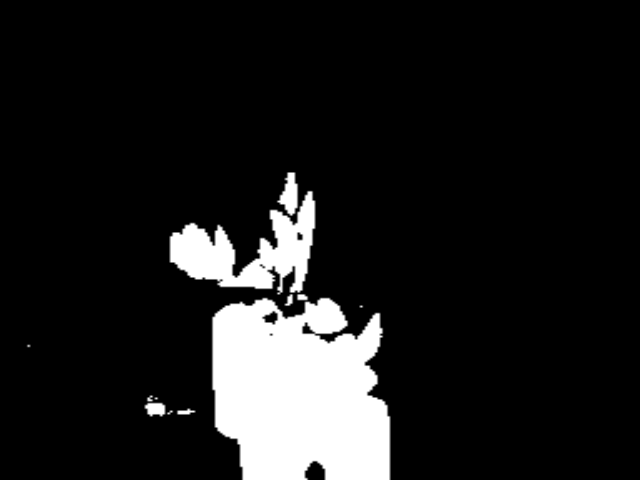}
	\end{subfigure}
	\begin{subfigure}{0.1\textwidth}
		\includegraphics[width=\textwidth]{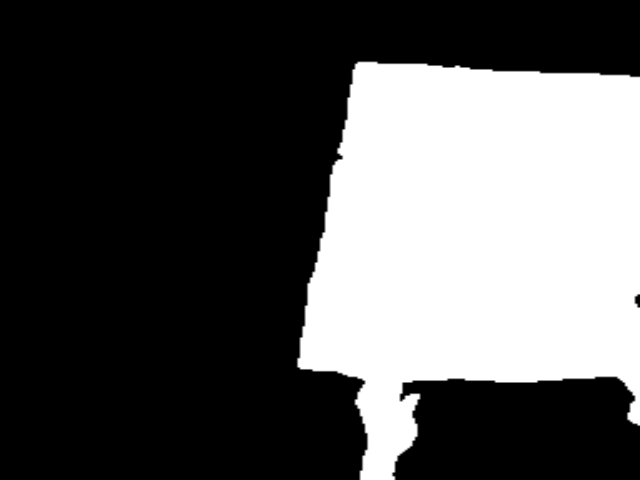}
	\end{subfigure}
	\begin{subfigure}{0.1\textwidth}
		\includegraphics[width=\textwidth]{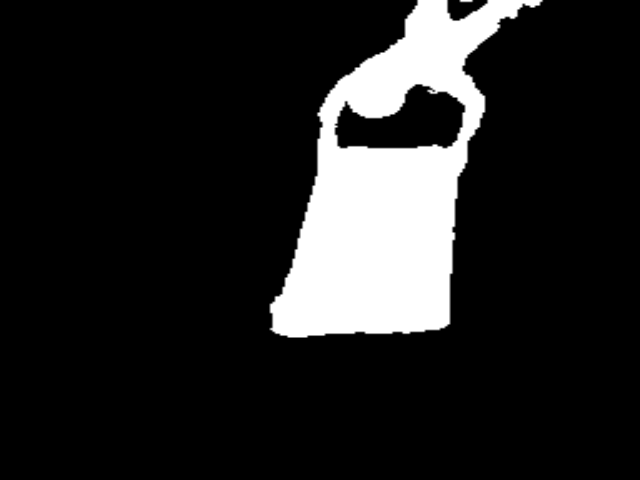}
	\end{subfigure}
	\begin{subfigure}{0.1\textwidth}
		\includegraphics[width=\textwidth]{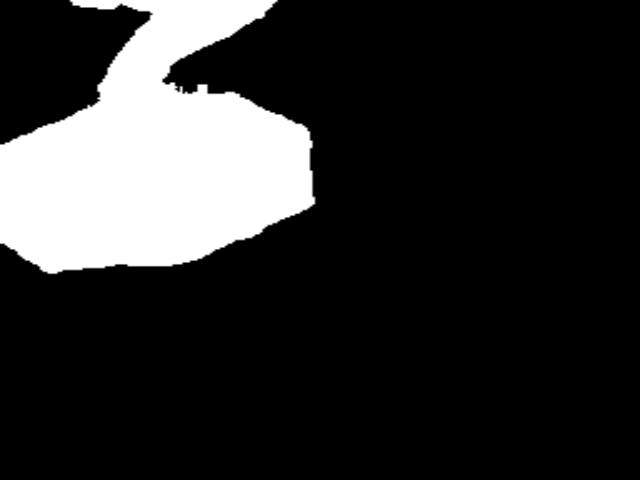}
	\end{subfigure}
	\begin{subfigure}{0.1\textwidth}
		\includegraphics[width=\textwidth]{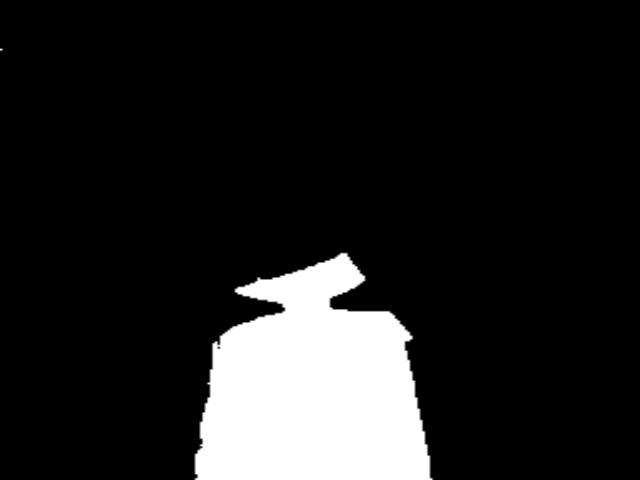}
	\end{subfigure}
	\begin{subfigure}{0.1\textwidth}
		\includegraphics[width=\textwidth]{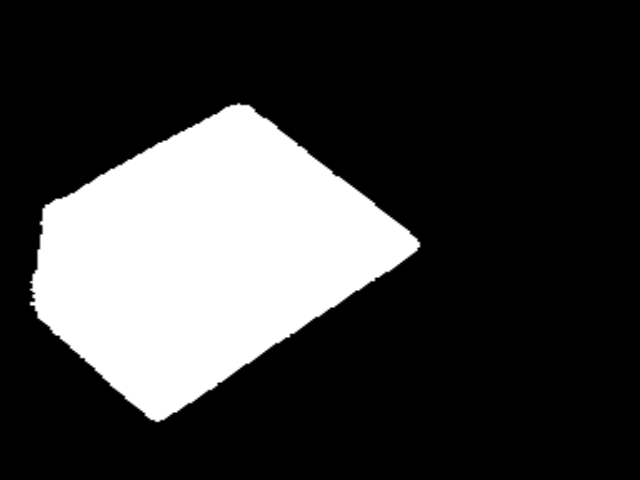}
	\end{subfigure}
    
	\vspace*{1.0mm}
	\begin{subfigure}{0.1\textwidth}
        \centering
		DSDNet\\~\cite{zheng19_cvpr}
	\end{subfigure}
	\begin{subfigure}{0.1\textwidth}
		\includegraphics[width=\textwidth]{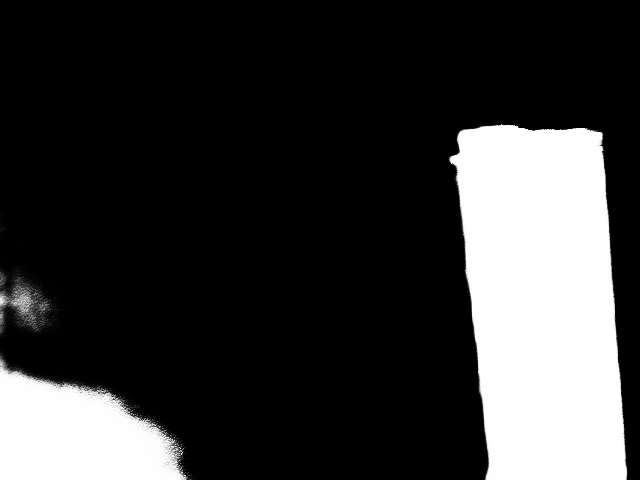}
	\end{subfigure}
	\begin{subfigure}{0.1\textwidth}
		\includegraphics[width=\textwidth]{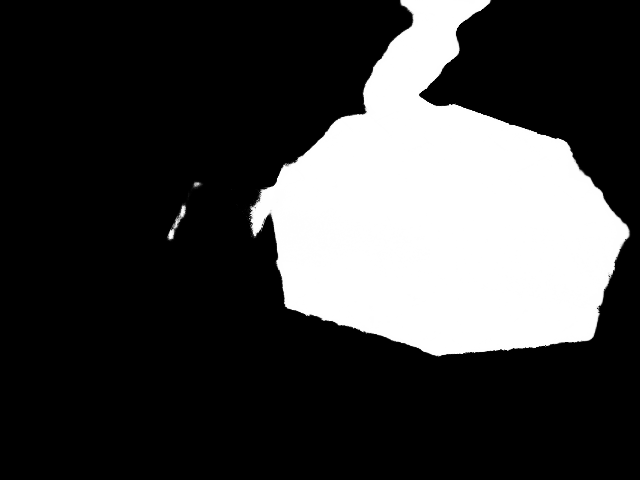}
	\end{subfigure}
	\begin{subfigure}{0.1\textwidth}
		\includegraphics[width=\textwidth]{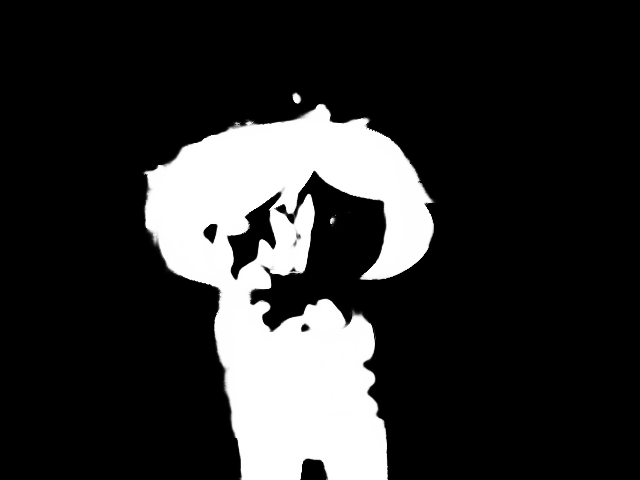}
	\end{subfigure}
	\begin{subfigure}{0.1\textwidth}
		\includegraphics[width=\textwidth]{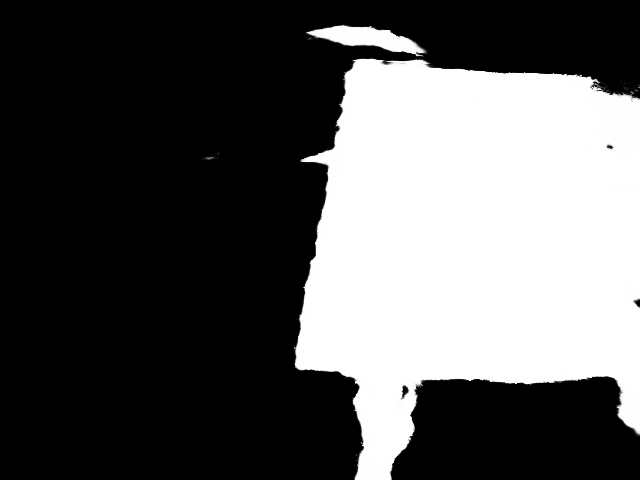}
	\end{subfigure}
	\begin{subfigure}{0.1\textwidth}
		\includegraphics[width=\textwidth]{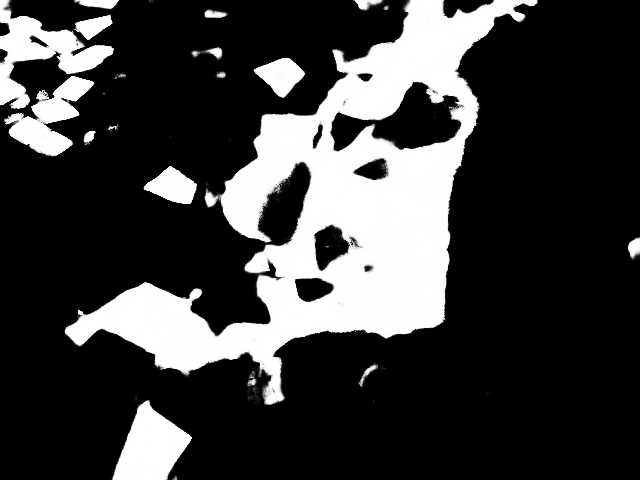}
	\end{subfigure}
	\begin{subfigure}{0.1\textwidth}
		\includegraphics[width=\textwidth]{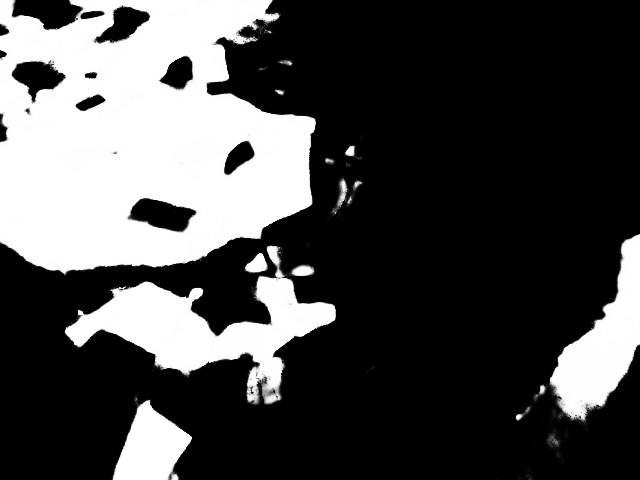}
	\end{subfigure}
	\begin{subfigure}{0.1\textwidth}
		\includegraphics[width=\textwidth]{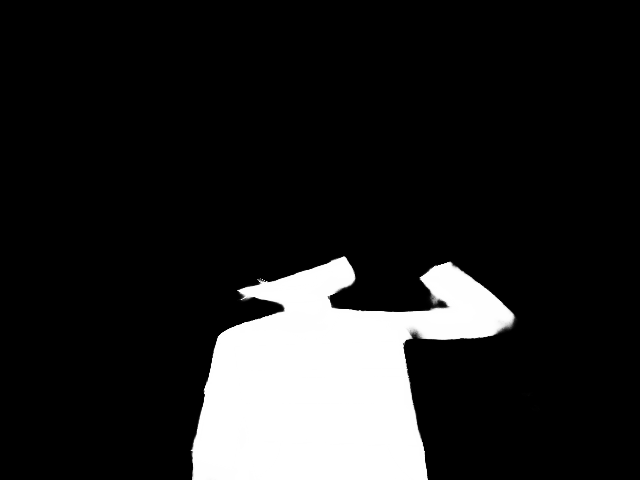}
	\end{subfigure}
	\begin{subfigure}{0.1\textwidth}
		\includegraphics[width=\textwidth]{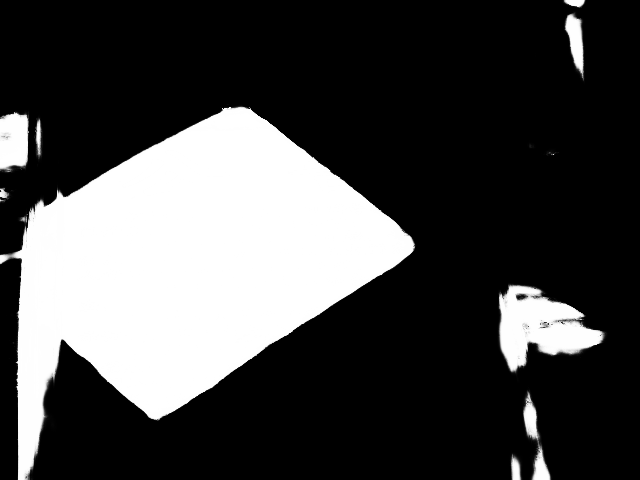}
	\end{subfigure}

	\vspace*{1.0mm}
	\begin{subfigure}{0.1\textwidth}
        \centering
		FDRNet\\~\cite{Zhu_2021_ICCV}
	\end{subfigure}
	\vspace*{1.0mm}
	\begin{subfigure}{0.1\textwidth}
		\includegraphics[width=\textwidth]{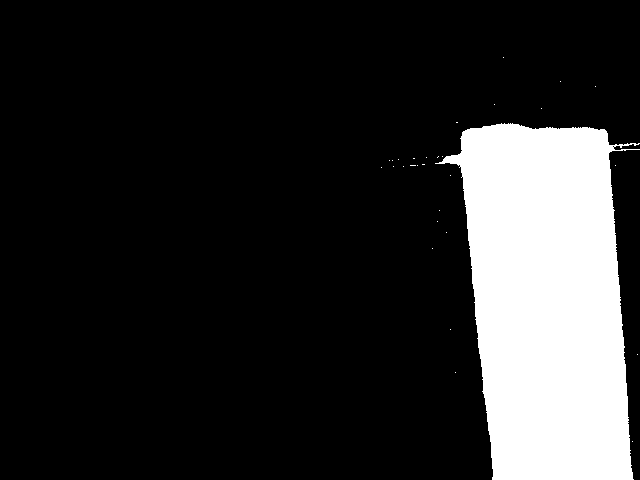}
	\end{subfigure}
	\begin{subfigure}{0.1\textwidth}
		\includegraphics[width=\textwidth]{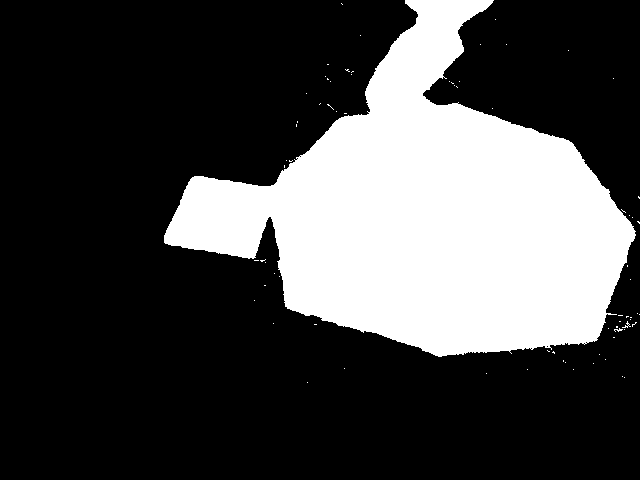}
	\end{subfigure}
	\begin{subfigure}{0.1\textwidth}
		\includegraphics[width=\textwidth]{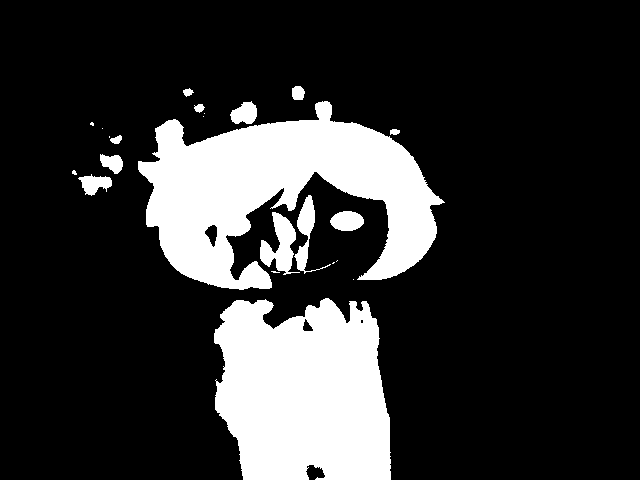}
	\end{subfigure}
	\begin{subfigure}{0.1\textwidth}
		\includegraphics[width=\textwidth]{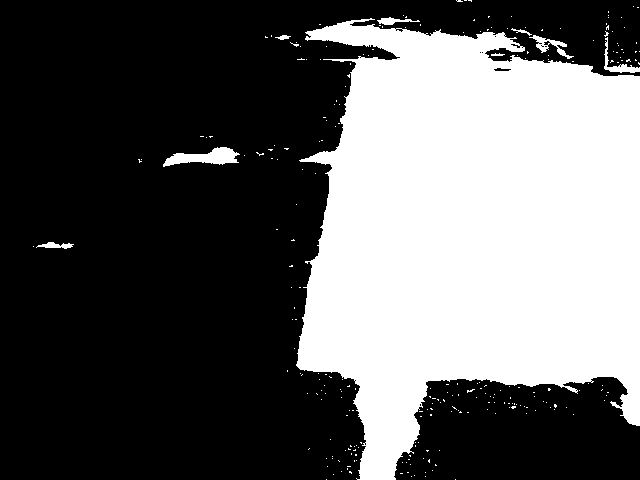}
	\end{subfigure}
	\begin{subfigure}{0.1\textwidth}
		\includegraphics[width=\textwidth]{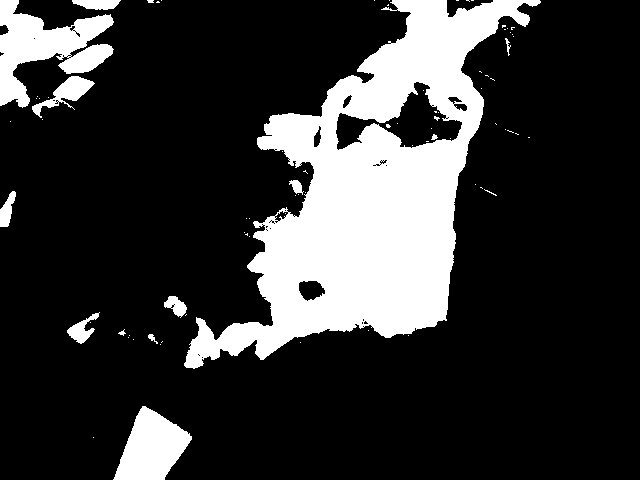}
	\end{subfigure}
	\begin{subfigure}{0.1\textwidth}
		\includegraphics[width=\textwidth]{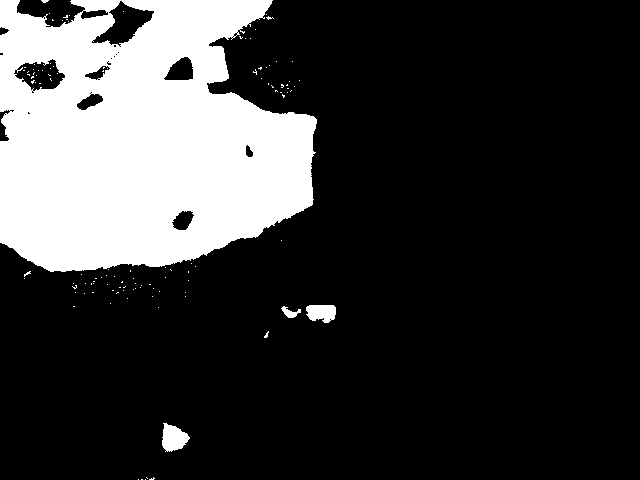}
	\end{subfigure}
	\begin{subfigure}{0.1\textwidth}
		\includegraphics[width=\textwidth]{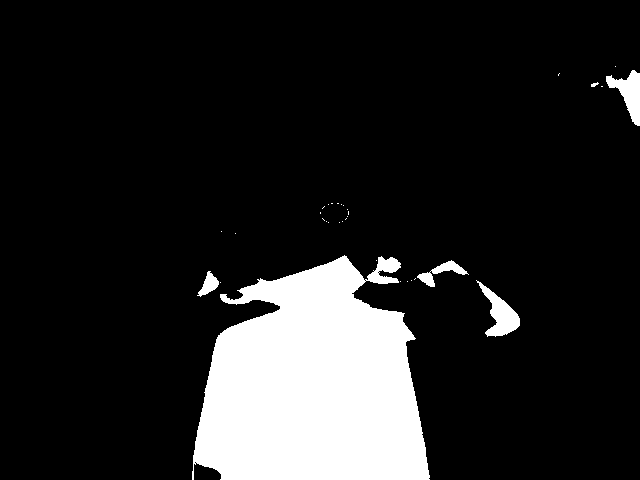}
	\end{subfigure}
	\begin{subfigure}{0.1\textwidth}
		\includegraphics[width=\textwidth]{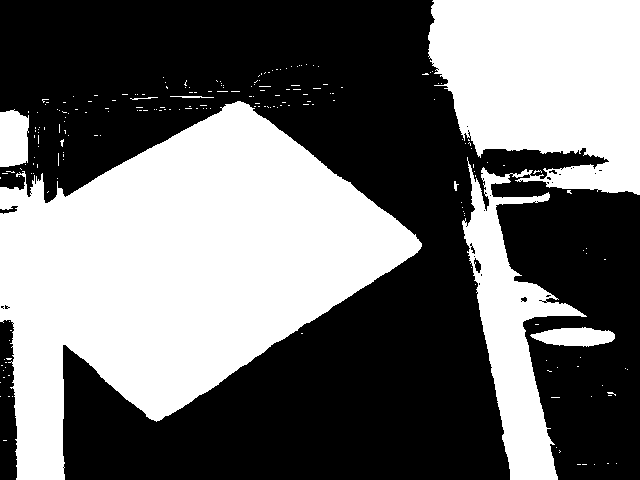}
	\end{subfigure}
    
	\vspace*{1.0mm}
	\begin{subfigure}{0.1\textwidth}
        \centering
		MTMT-Net\\~\cite{chen20_cvpr}
	\end{subfigure}
	\begin{subfigure}{0.1\textwidth}
		\includegraphics[width=\textwidth]{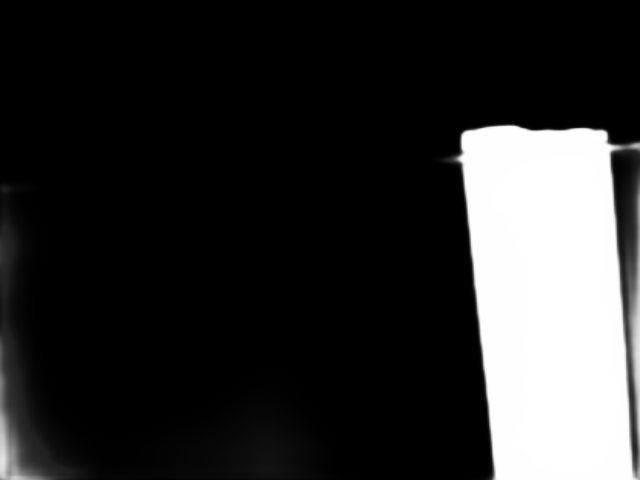}
	\end{subfigure}
	\begin{subfigure}{0.1\textwidth}
		\includegraphics[width=\textwidth]{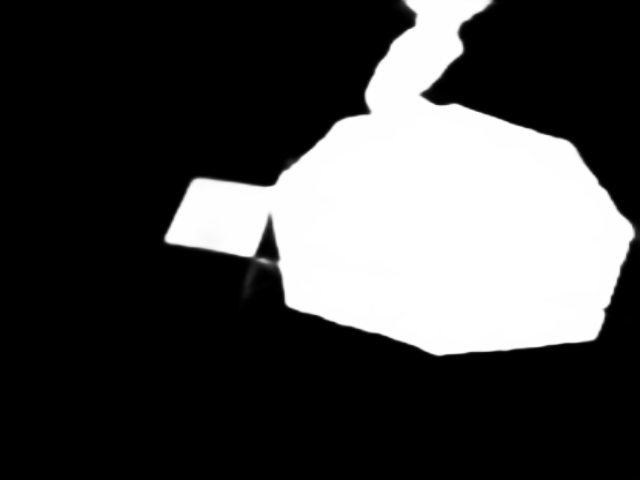}
	\end{subfigure}
	\begin{subfigure}{0.1\textwidth}
		\includegraphics[width=\textwidth]{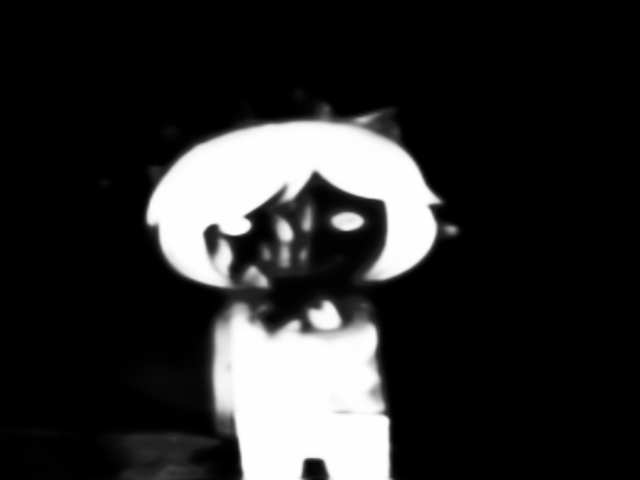}
	\end{subfigure}
	\begin{subfigure}{0.1\textwidth}
		\includegraphics[width=\textwidth]{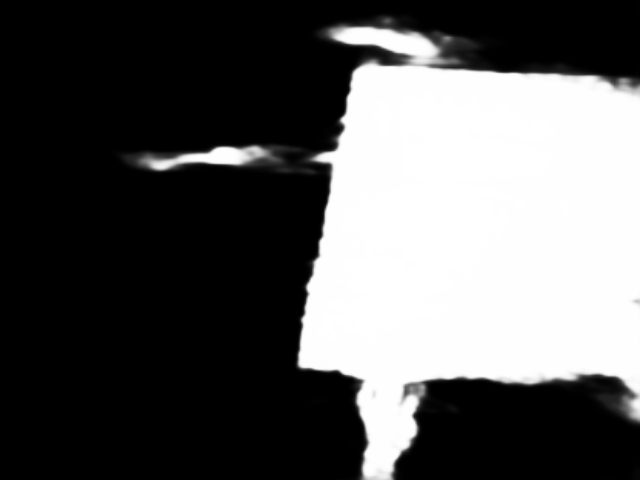}
	\end{subfigure}
	\begin{subfigure}{0.1\textwidth}
		\includegraphics[width=\textwidth]{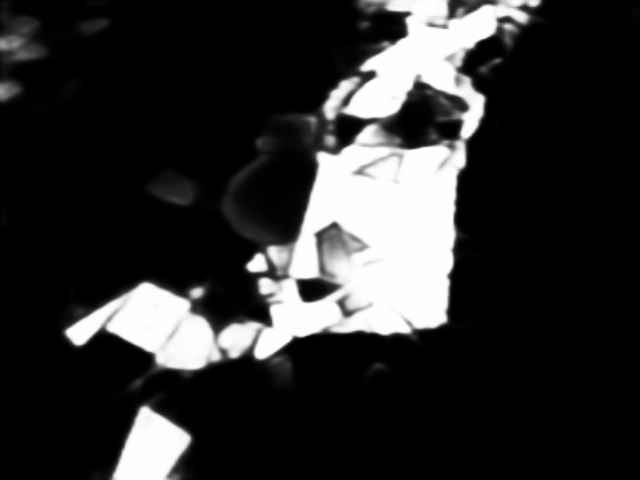}
	\end{subfigure}
	\begin{subfigure}{0.1\textwidth}
		\includegraphics[width=\textwidth]{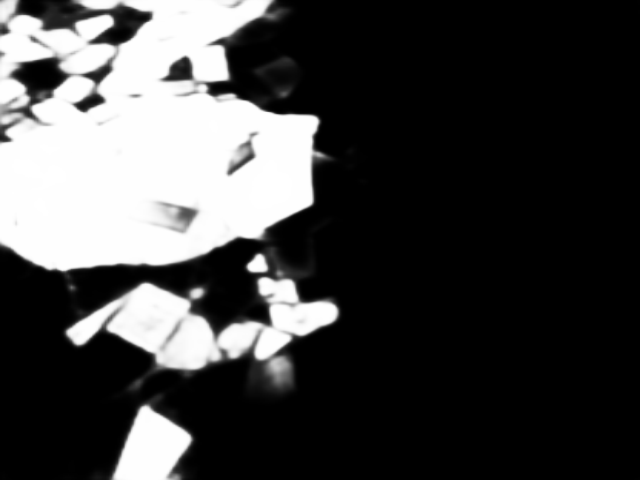}
	\end{subfigure}
	\begin{subfigure}{0.1\textwidth}
		\includegraphics[width=\textwidth]{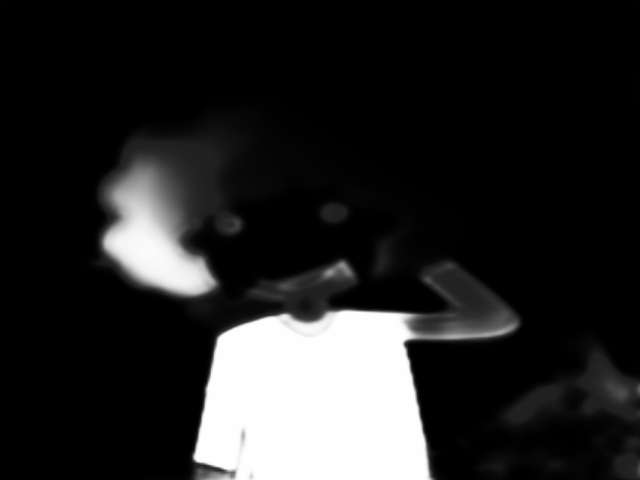}
	\end{subfigure}
	\begin{subfigure}{0.1\textwidth}
		\includegraphics[width=\textwidth]{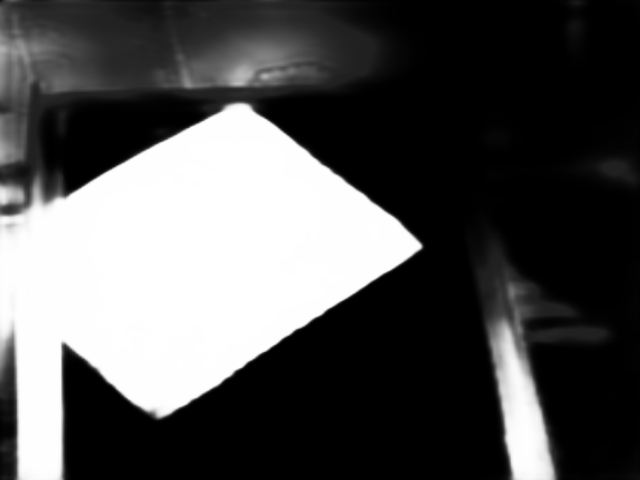}
	\end{subfigure}
    
	\vspace*{1.0mm}
	\begin{subfigure}{0.1\textwidth}
        \centering
		SDCM\\~\cite{zhu_mm2022}
	\end{subfigure}
	\begin{subfigure}{0.1\textwidth}
		\includegraphics[width=\textwidth]{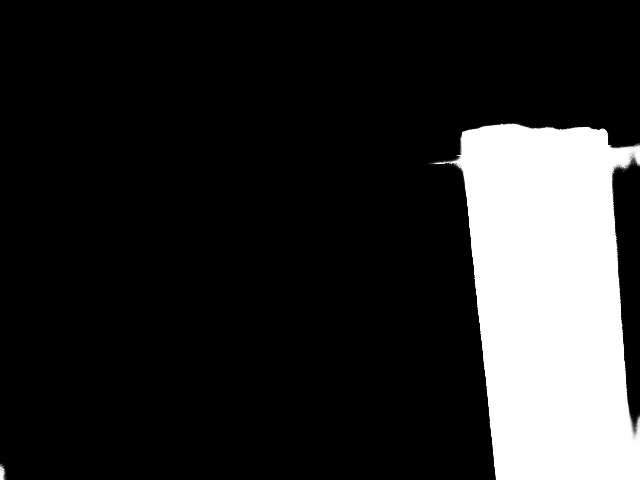}
	\end{subfigure}
	\begin{subfigure}{0.1\textwidth}
		\includegraphics[width=\textwidth]{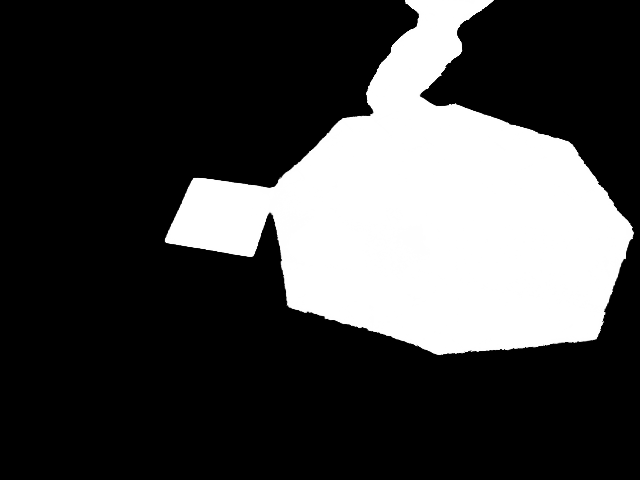}
	\end{subfigure}
	\begin{subfigure}{0.1\textwidth}
		\includegraphics[width=\textwidth]{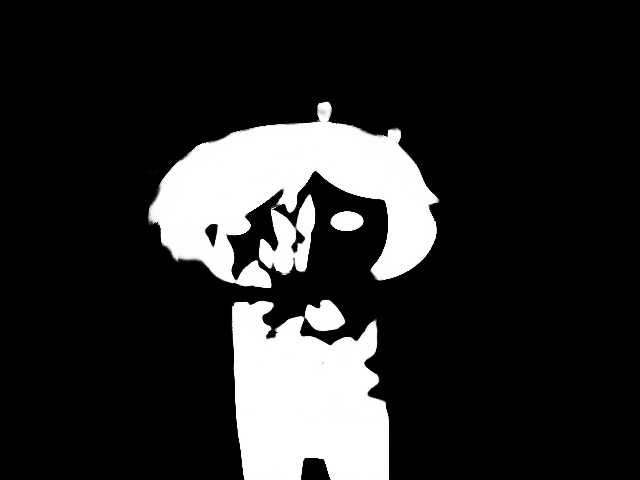}
	\end{subfigure}
	\begin{subfigure}{0.1\textwidth}
		\includegraphics[width=\textwidth]{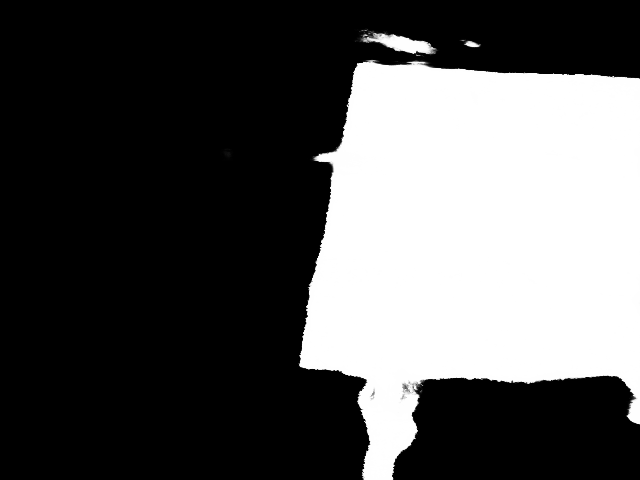}
	\end{subfigure}
	\begin{subfigure}{0.1\textwidth}
		\includegraphics[width=\textwidth]{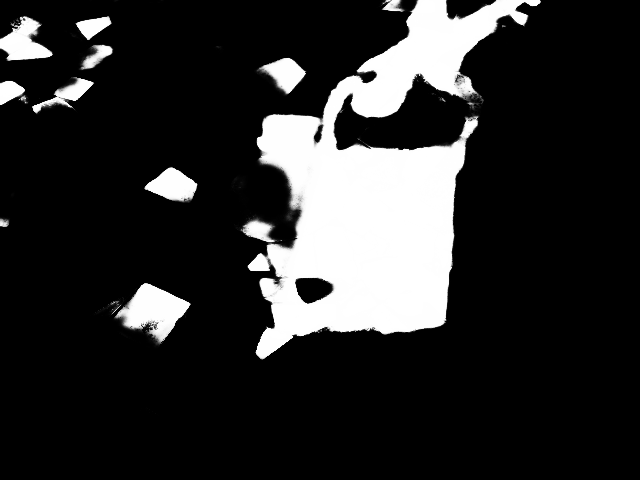}
	\end{subfigure}
	\begin{subfigure}{0.1\textwidth}
		\includegraphics[width=\textwidth]{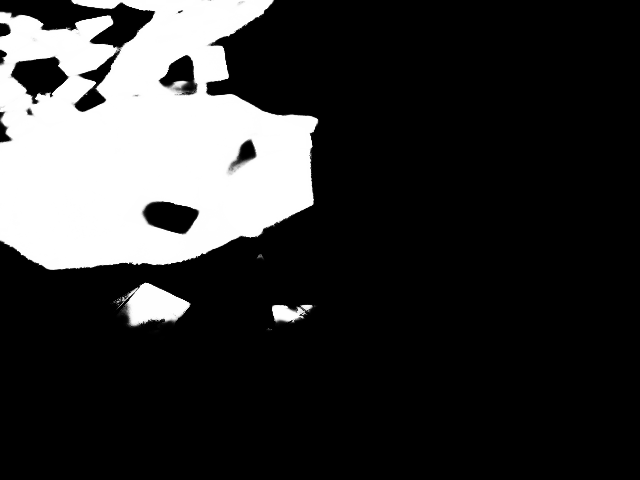}
	\end{subfigure}
	\begin{subfigure}{0.1\textwidth}
		\includegraphics[width=\textwidth]{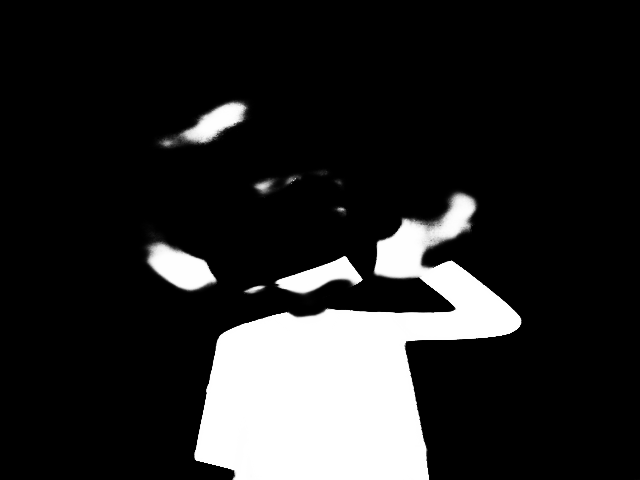}
	\end{subfigure}
	\begin{subfigure}{0.1\textwidth}
		\includegraphics[width=\textwidth]{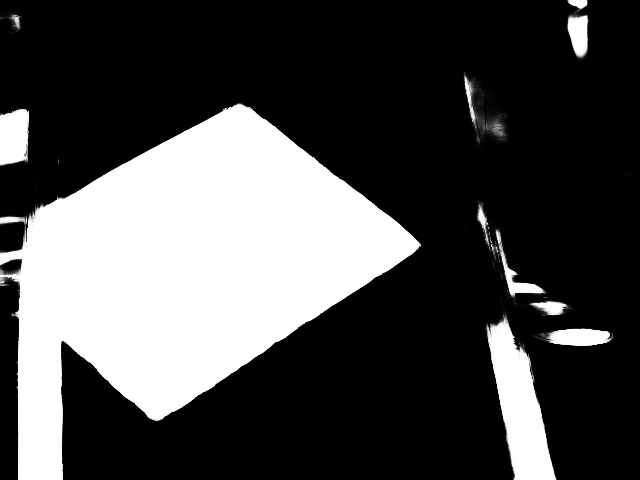}
	\end{subfigure}
    
	\vspace*{1.0mm}
	\begin{subfigure}{0.1\textwidth}
        \centering
		RMLANet\\~\cite{jie2022icme, jie2023rmlanet}
	\end{subfigure}
	\begin{subfigure}{0.1\textwidth}
		\includegraphics[width=\textwidth]{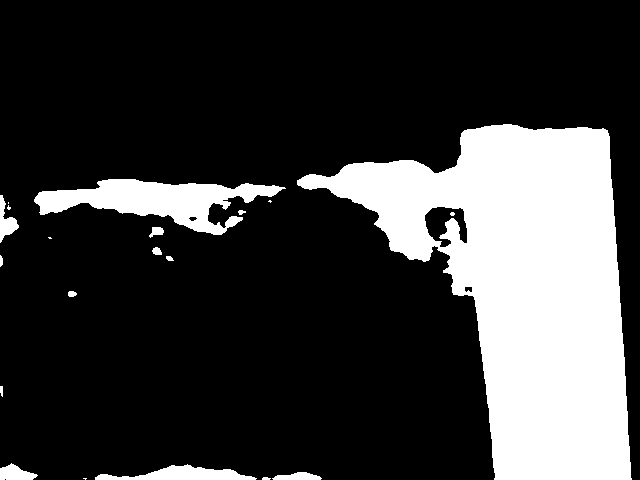}
	\end{subfigure}
	\begin{subfigure}{0.1\textwidth}
		\includegraphics[width=\textwidth]{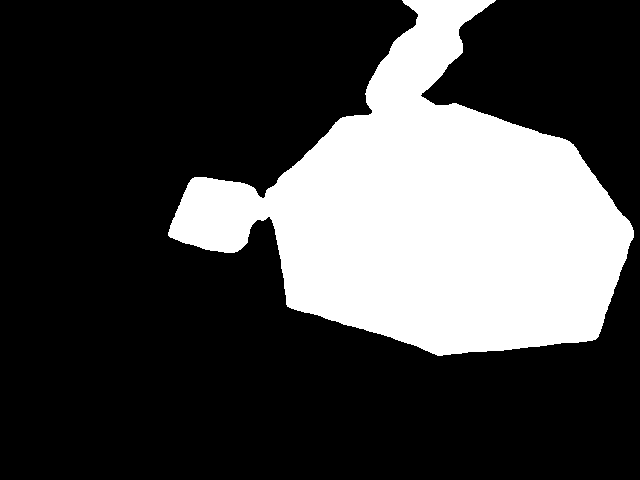}
	\end{subfigure}
	\begin{subfigure}{0.1\textwidth}
		\includegraphics[width=\textwidth]{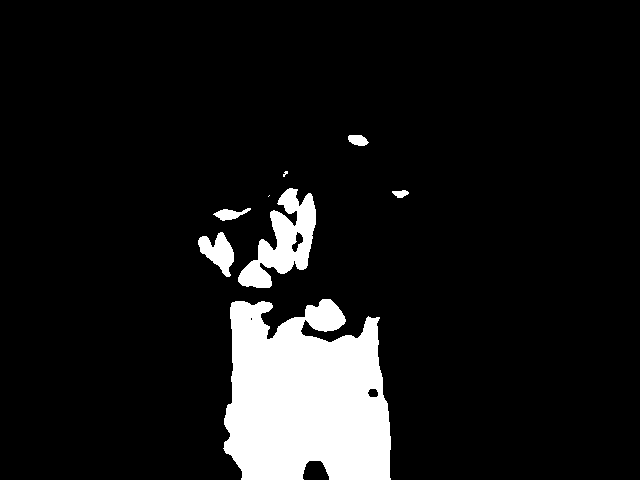}
	\end{subfigure}
	\begin{subfigure}{0.1\textwidth}
		\includegraphics[width=\textwidth]{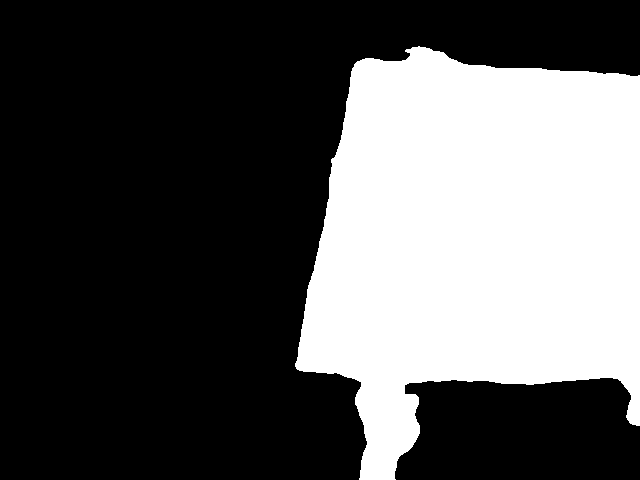}
	\end{subfigure}
	\begin{subfigure}{0.1\textwidth}
		\includegraphics[width=\textwidth]{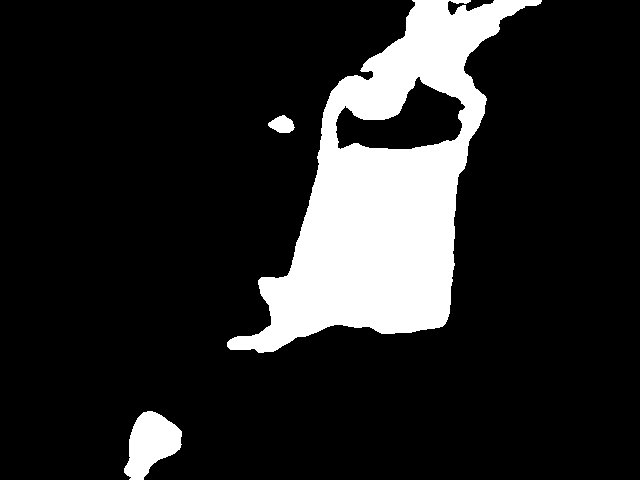}
	\end{subfigure}
	\begin{subfigure}{0.1\textwidth}
		\includegraphics[width=\textwidth]{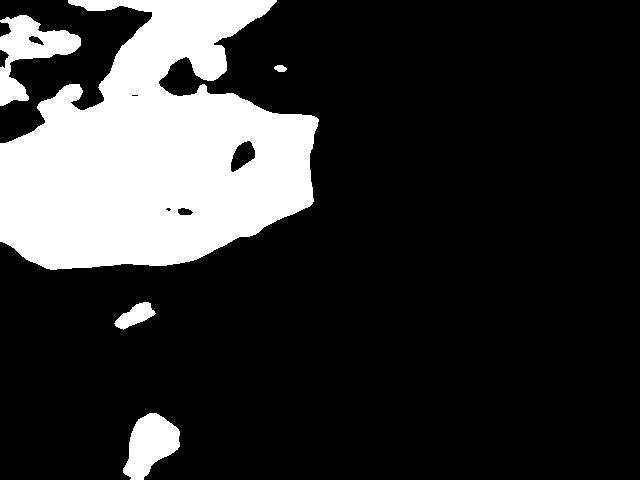}
	\end{subfigure}
	\begin{subfigure}{0.1\textwidth}
		\includegraphics[width=\textwidth]{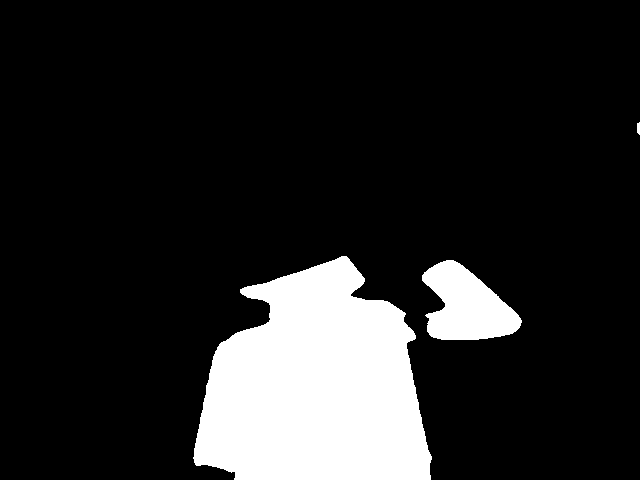}
	\end{subfigure}
	\begin{subfigure}{0.1\textwidth}
		\includegraphics[width=\textwidth]{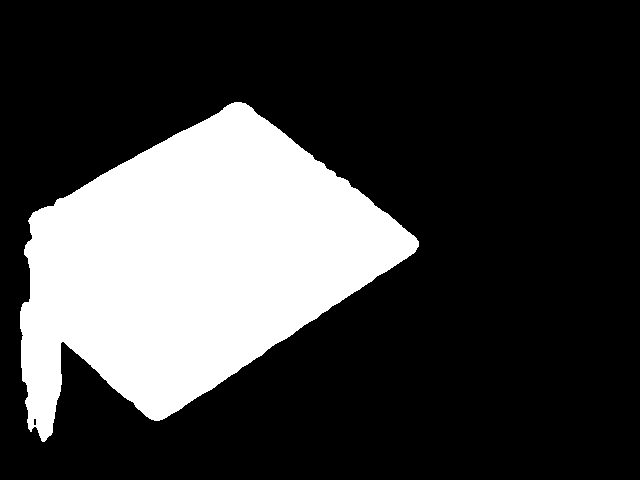}
	\end{subfigure}

	\caption{Qualitative comparison of the predicted shadow maps between our approach (the third row) and five other methods (DSDNet~\cite{zheng19_cvpr}, FDRNet~\cite{Zhu_2021_ICCV}, MTMT-Net~\cite{chen20_cvpr}, SDCM~\cite{zhu_mm2022} and RMLANet~\cite{jie2022icme, jie2023rmlanet}) against the ground truth in the second row on ISTD~\cite{istd_dataset} benchmark dataset. Best viewed on screen.}
	\label{fig_quantitative_istd}
	
\end{figure*}
\noindent\textbf{Quantitative Comparison.} As illustrated in \figref{fig_quantitative_sbu} and \figref{fig_quantitative_istd}, we compare our predicted masks with other approaches quantitatively. \figref{fig_quantitative_sbu} shows the visual comparisons on the SBU dataset. It can be observed that existing approaches struggle to distinguish shadows from complex backgrounds or shadow-like objects, while our method demonstrates better performance on these difficult cases, \eg, in the 6th to the 10th row. 
In addition, visual comparisons on the ISTD dataset are presented in \figref{fig_quantitative_istd}. It is worth noting that the boundaries between shadow and non-shadow areas are smoother in our method. In the 5th and 6th column, our method demonstrates better performance when facing shadows cast on complex backgrounds.

\begin{table}[htbp]
    \begin{center}
        \caption{Quantitative comparison with the state-of-the-art methods for shadow detection on the $SBUTestNew$ in~\cite{Yang_2023_ICCV}. The best and the second best results are highlighted in bold and underlined respectively.}
        \begin{tabular}{|c|c|c|c|c|}
        \hline 
        Method & \makecell{Trainable\\Parameters} & BER $\downarrow$ & BER$_{NS}$ $\downarrow$ & BER$_{S}$ $\downarrow$\\
        \hline 
        \hline 
        BDRAR~\cite{zhu18_eccv} & 42.5M & 6.49 & 3.29 & 9.68\\
        DSC~\cite{hu18_cvpr} & 79M & 8.08 & 3.91 & 12.25 \\
        DSDNet~\cite{zheng19_cvpr} & 58.2M & 5.60 & 3.34 & \underline{7.86} \\
        MTMT-Net~\cite{chen20_cvpr} & 44.1M & 7.41 & \textbf{0.97} & 13.68 \\
        FSDNet~\cite{gy_tip2021} & 4.4M & 10.87 & \underline{1.34} & 20.39 \\
        FDRNet~\cite{Zhu_2021_ICCV} & 10.8M & 5.93 & 1.71 & 10.93 \\
        SDCM ~\cite{zhu_mm2022} & 11.0M & 5.71 & 2.36 & 9.07 \\
        SILT~\cite{Yang_2023_ICCV} & 12.18M & \underline{5.23} & 4.23 & \textbf{6.22} \\
        \hline
        \hline 
        Ours & 11.5M & \textbf{5.14} & 2.09 & 8.18\\
        \hline
        \end{tabular}
        \label{table_detection_sbu_new}
        \end{center}
\end{table}
\begin{table}[htbp]
    \begin{center}
        \caption{Quantitative comparison with the state-of-the-art methods for shadow detection on the CUHK~\cite{gy_tip2021} benchmark dataset. The best and the second best results are highlighted in bold and underlined respectively.}
        \begin{tabular}{|c|c|c|}
        \hline 
        Method & Trainable Parameters & BER $\downarrow$\\
        \hline 
        \hline 
        RCMPNet~\cite{liao2021mm} & - & 21.23 \\
        A+D Net~\cite{le18_eccv} & 54.4M & 12.43 \\
        BDRAR~\cite{zhu18_eccv} & 42.5M & 9.18 \\
        DSC~\cite{hu18_cvpr} & 79M & 8.65 \\
        FSDNet~\cite{gy_tip2021} & 4.4M & 8.65 \\
        DSDNet~\cite{zheng19_cvpr} & 58.2M & \underline{8.27} \\
        RMLANet & 54.9M & 7.64\\
        \hline
        \hline 
        Ours & 11.5M & \textbf{7.51}\\
        \hline
        \end{tabular}
        \label{table_detection_cuhk}
        \end{center}
\end{table}

\subsection{Ablation Studies}
In this section, extensive ablation studies are conducted to verify the impact of different settings and the effectiveness of our proposed network.
\subsubsection{Effectiveness of Our Network}
\begin{table}[htbp]
    \begin{center}
    \caption{Component analysis. All experiments are conducted on the SBU dataset. $mha$ and $ffn$ means inserting adapter in multi-header attention and feedforward network of the transformer block, respectively. $Freeze$ indicates whether freezing the backbone in the point prompt generator. $point$, $box$, $mask$ represents the set of three different spare prompts supported.}
    \begin{tabular}{|c|cc|c|ccc|c|}
    \hline
    \multirow{2}{*}{Baseline} & \multicolumn{2}{c|}{Adapter} & \multirow{2}{*}{Freeze} & \multicolumn{3}{c|}{Prompt} & \multirow{2}{*}{BER $\downarrow$} \\
    \cline{2-3} \cline{5-7}
     & mha & ffn &  & point & box & mask &\\
    \hline
    \hline
    $\checkmark$ & $\times$ & $\times$ & $\times$ & $\times$ & $\times$ & $\times$ & 3.71 \\
    $\checkmark$ & $\times$ & $\times$ & $\times$ & $\checkmark$ & $\times$ & $\times$ & 3.45 \\
    $\checkmark$ & $\checkmark$ & $\times$ & $\times$ & $\checkmark$ & $\times$ & $\times$ & 3.01 \\
    $\checkmark$ & $\checkmark$ & $\checkmark$ & $\times$ & $\checkmark$ & $\times$ & $\times$ & 2.81 \\
    $\checkmark$ & $\checkmark$ & $\checkmark$ & $\checkmark$ & $\checkmark$ & $\times$ & $\times$ & \textbf{2.75} \\
    \hline
    \end{tabular}
    \label{table_components_analysis}
    \end{center}
  \end{table}
In our network, there are three key components: (1) the inserted adapter, (2) the point prompt generator, (3) the point sampling strategies. We set the SAM as our baseline, with only the mask decoder trainable. As shown in \tabref{table_components_analysis}, both adapters in multi-header attention and feedforward network of the transformer block are essential in our method. Specifically, adapters in the feedforward network are more important than in multi-header attention block. Moreover, point prompts outperform the bounding box and mask prompts. The reason is that both bounding box and mask prompts are too noisy and are coarse-grained prompts. We also find that it is beneficial to freeze the backbone (here pre-trained Efficient-B1) in the point prompt generator. The possible reason is that the scale of existing shadow detection dataset is relatively small and prone to overfitting with larger networks. 

\subsubsection{Number of Point Prompts}
\begin{table}[htb!]
    \begin{center}
      \caption{Ablation study on different number of top-k point prompts.}
    \begin{tabular}{|c|c|c||c|c|c|}
    \hline 
    \multicolumn{2}{|c|}{Number of Points} & \multirow{2}{*}{BER $\downarrow$} & \multicolumn{2}{c|}{Number of Points} & \multirow{2}{*}{BER $\downarrow$}\\
    \cline{1-2} 
    \cline{4-5} 
    positive & negative & & positive & negative &  \\
    \hline 
    \hline 
      1 & 1 & 2.91 & 7 & 7 & 2.92 \\
      2 & 2 & 2.89 & 8 & 8 & \textbf{2.82} \\
      3 & 3 & 2.97 & 9 & 9 & 2.93 \\  
      4 & 4 & 2.91 & 10 & 10 & 2.98 \\ 
      5 & 0 & 2.93 & 10 & 0 & 2.90 \\
      5 & 5 & 2.89 & 15 & 15 & 2.85 \\
      6 & 6 & 2.98 & 20 & 20 & 2.94 \\ 
    \hline 
    \end{tabular}
    \label{table_number_of_topk_points}
    \end{center}
\end{table}
\begin{table}[htb!]
    \begin{center}
      \caption{Ablation study on the effect of different grid size settings.}
    \begin{tabular}{|c|c|}
    \hline 
    Grid size & BER $\downarrow$\\
    \hline 
    \hline 
      12 & 2.90 \\
      16 & \textbf{2.75} \\
      24 & 2.81 \\  
      32 & 2.91 \\
    \hline 
    \end{tabular}
    \label{table_number_of_grid_points}
    \end{center}
\end{table}
Our positive and negative points are selected from the predicted coarse shadow mask based on probability values. Intuitively, selecting too many or too few points are both problematic. Too many points may involve non-shadow points, while too few points may miss valuable positive or negative prompts. Thus, we choose different number of points as shown in \tabref{table_number_of_topk_points}. As can be seen, our model achieves the best performance when using 8 positive and 8 negative points. However, we found one interesting phenomenon when using top-k strategy to select point prompt. The selected positive or negative points tend to cluster into two groups, which means that most of the areas will have no point prompts. This is why we propose the grid sampling strategy that can effectively cover the entire area. As shown in \tabref{table_number_of_grid_points}, four different grid size settings are used. When the grid size is equal to 16, our method achieves the best performance.

\subsubsection{Visualization of Selected Point Prompts}
\begin{figure}[!htb]
	\centering
	\vspace*{1.0mm}
	\begin{subfigure}{0.14\textwidth}
		\includegraphics[width=\textwidth]{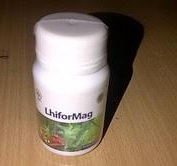}
		\captionsetup{justification=centering}
        \vspace{-5.5mm} \caption{\scriptsize{RGB input}}
	\end{subfigure}
	\begin{subfigure}{0.14\textwidth}
		\includegraphics[width=\textwidth]{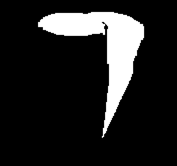}
		\captionsetup{justification=centering}
        \vspace{-5.5mm} \caption{\scriptsize{Ground truth}}
	\end{subfigure}
	
	\vspace*{1.0mm}
	\begin{subfigure}{0.14\textwidth}
		\includegraphics[width=\textwidth]{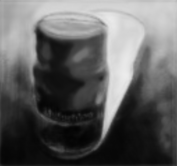}
		\captionsetup{justification=centering}
        \vspace{-5.5mm} \caption{\scriptsize{Coarse result}}
	\end{subfigure}
	\begin{subfigure}{0.14\textwidth}
		\includegraphics[width=\textwidth]{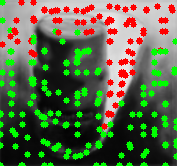}
		\captionsetup{justification=centering}
        \vspace{-5.5mm} \caption{\scriptsize{Point prompts}}
	\end{subfigure}
	\begin{subfigure}{0.14\textwidth}
		\includegraphics[width=\textwidth]{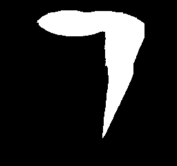}
		\captionsetup{justification=centering}
        \vspace{-5.5mm} \caption{\scriptsize{Final result}}
	\end{subfigure}

	\caption{Visualiaztion of the proposed grid point prompts. In (d), the red and green points indicate shadows and backgrounds, respectively.}
	\label{figure_vis_grid_prompts}
	
\end{figure}

To better understand how the grid point sampling works, we visualize the input image, the corresponding ground truth, the coarse shadow mask, the colored grid points and the final prediction in \figref{figure_vis_grid_prompts}. As it shows, the shadow (red) prompts distribute in most true shadow areas, while the non-shadow or background points are also mostly correct. Despite there exist some error prompts, our method can still recognize them and produce satisfactory results.

\subsubsection{Model Size and Speed}
\begin{table}[htbp!]
    \begin{center}
      \caption{Comparison of model size and speed. }
    \begin{tabular}{|c|c|c|c|}
    \hline 
    Name & \makecell{Trainable \\Parameters} & \makecell{Inference\\Speed (s)} & BER $\downarrow$ \\
    \hline 
    \hline 
      DSC~\cite{hu18_cvpr}~\tablefootnote[1]{We use PyTorch version~\url{https://github.com/stevewongv/DSC-PyTorch}.} & 122,492,659 & 0.041 & 5.59 \\
      BDRAR~\cite{zhu18_eccv} & 42,459,867 & 0.028 & 3.64 \\
      MTMT-Net~\cite{chen20_cvpr} & 44,129,316 & 0.022 & 3.15 \\ 
      FDRNet~\cite{Zhu_2021_ICCV} & \textbf{10,768,970} & \textbf{0.021} & 3.04\\
      SDCM ~\cite{zhu_mm2022} & 10,947,626 & 0.022 & 3.02\\
      RMLANet~\cite{jie2022icme, jie2023rmlanet} & 54,973,371 & 0.026 & 2.97 \\
      \hline 
      \hline 
      Ours & 11,564,121 & 0.070 & \textbf{2.74} \\ 
    \hline 
    \end{tabular}
    \label{table_size_speed}
    \end{center}
\end{table}
Considering that our network is based on the large foundation model SAM, we further analyze the model size and inference speed in \tabref{table_size_speed}. Although our model has $107,748,217$ parameters, only $11,564,121$ of them are trainable, since we freeze the image encoder in SAM and the EfficientNet backbone for prompt generation. Specifically, our model has only slightly more trainable parameters than FDRNet~\cite{Zhu_2021_ICCV} and SDCM~\cite{zhu_mm2022}, while owes much less than DSC~\cite{hu18_cvpr}, BDRAR~\cite{zhu18_eccv}, MTMT-Net~\cite{chen20_cvpr} and RMLANet~\cite{jie2022icme, jie2023rmlanet}. However, the inference speed of our model is slower than others, due to the large size of the foundation model. To speed up, MobileSAM~\cite{zhang2023faster} and FastSAM~\cite{zhang2023faster} can be adopted.
\section{Conclusion}
In this paper, we leverage the segment anything model (SAM) and propose a novel network named AdapterShadow to detect shadows. To reduce the training afford and adapt the pretrained SAM for shadow detection, we propose to insert adapters into the transformer block of SAM's image encoder. To generate high-quality point prompts automatically, we propose an auxiliary network which produces coarse shadow mask and samples grid point prompts. Comprehensive experiments are conducted on four widely used benchmarks to demonstrate the superiority of our AdapterShadow, both qualitatively and quantitatively.

\bibliographystyle{IEEEtran}
\bibliography{reference}

\begin{thebibliography}{10}
\providecommand{\url}[1]{#1}
\csname url@samestyle\endcsname
\providecommand{\newblock}{\relax}
\providecommand{\bibinfo}[2]{#2}
\providecommand{\BIBentrySTDinterwordspacing}{\spaceskip=0pt\relax}
\providecommand{\BIBentryALTinterwordstretchfactor}{4}
\providecommand{\BIBentryALTinterwordspacing}{\spaceskip=\fontdimen2\font plus
\BIBentryALTinterwordstretchfactor\fontdimen3\font minus \fontdimen4\font\relax}
\providecommand{\BIBforeignlanguage}[2]{{%
\expandafter\ifx\csname l@#1\endcsname\relax
\typeout{** WARNING: IEEEtran.bst: No hyphenation pattern has been}%
\typeout{** loaded for the language `#1'. Using the pattern for}%
\typeout{** the default language instead.}%
\else
\language=\csname l@#1\endcsname
\fi
#2}}
\providecommand{\BIBdecl}{\relax}
\BIBdecl

\bibitem{finlayson2009entropy}
G.~D. Finlayson, M.~S. Drew, and C.~Lu, ``Entropy minimization for shadow removal,'' \emph{International Journal of Computer Vision}, vol.~85, no.~1, pp. 35--57, 2009.

\bibitem{huang2009moving}
J.-B. Huang and C.-S. Chen, ``Moving cast shadow detection using physics-based features,'' in \emph{Proc. IEEE Conf. Comput. Vis. Pattern Recognit}, 2009, pp. 2310--2317.

\bibitem{sbu_dataset}
T.~F.~Y. Vicente, L.~Hou, C.-P. Yu, M.~Hoai, and D.~Samaras, ``Large-scale training of shadow detectors with noisily-annotated shadow examples,'' in \emph{Proc. Eur. Conf. Comput. Vis.}, 2016, pp. 816--832.

\bibitem{ucf_dataset}
J.~Zhu, K.~G.~G. Samuel, S.~Z. Masood, and M.~F. Tappen, ``Learning to recognize shadows in monochromatic natural images,'' in \emph{Proc. IEEE Conf. Comput. Vis. Pattern Recognit}, 2010, pp. 223--230.

\bibitem{istd_dataset}
J.~Wang, X.~Li, and J.~Yang, ``Stacked conditional generative adversarial networks for jointly learning shadow detection and shadow removal,'' in \emph{Proc. IEEE Conf. Comput. Vis. Pattern Recognit}, 2018, pp. 1788--1797.

\bibitem{gy_tip2021}
X.~Hu, T.~Wang, C.-W. Fu, Y.~Jiang, Q.~Wang, and P.-A. Heng, ``Revisiting shadow detection: A new benchmark dataset for complex world,'' \emph{IEEE Trans. Image Process.}, vol.~30, pp. 1925--1934, 2021.

\bibitem{deng2009imagenet}
J.~Deng, W.~Dong, R.~Socher, L.-J. Li, K.~Li, and L.~Fei-Fei, ``Imagenet: A large-scale hierarchical image database,'' in \emph{Proc. IEEE Int. Conf. Comput. Vis.}, 2009, pp. 248--255.

\bibitem{jie2023sam}
L.~Jie and H.~Zhang, ``When sam meets shadow detection,'' \emph{arXiv preprint arXiv:2305.11513}, 2023.

\bibitem{he2023accuracy}
S.~He, R.~Bao, J.~Li, P.~E. Grant, and Y.~Ou, ``Accuracy of segment-anything model (sam) in medical image segmentation tasks,'' \emph{arXiv:2304.09324}, 2023.

\bibitem{cucchiara2003detecting}
R.~Cucchiara, C.~Grana, M.~Piccardi, and A.~Prati, ``Detecting moving objects, ghosts, and shadows in video streams,'' \emph{IEEE Trans. Pattern Anal. Mach. Intell.}, vol.~25, no.~10, pp. 1337--1342, 2003.

\bibitem{chen2010enhanced}
C.-T. Chen, C.-Y. Su, and W.-C. Kao, ``An enhanced segmentation on vision-based shadow removal for vehicle detection,'' in \emph{The 2010 International Conference on Green Circuits and Systems}, 2010, pp. 679--682.

\bibitem{huang2011characterizes}
X.~Huang, G.~Hua, J.~Tumblin, and L.~Williams, ``What characterizes a shadow boundary under the sun and sky?'' in \emph{Proc. IEEE Int. Conf. Comput. Vis.}, 2011, pp. 898--905.

\bibitem{martel2005moving}
N.~Martel-Brisson and A.~Zaccarin, ``Moving cast shadow detection from a gaussian mixture shadow model,'' in \emph{Proc. IEEE Conf. Comput. Vis. Pattern Recognit}, vol.~2, 2005, pp. 643--648.

\bibitem{zhu2010learning}
J.~Zhu, K.~G. Samuel, S.~Z. Masood, and M.~F. Tappen, ``Learning to recognize shadows in monochromatic natural images,'' in \emph{Proc. IEEE Conf. Comput. Vis. Pattern Recognit}, 2010, pp. 223--230.

\bibitem{lalonde2010detecting}
J.-F. Lalonde, A.~A. Efros, and S.~G. Narasimhan, ``Detecting ground shadows in outdoor consumer photographs,'' in \emph{Proc. Eur. Conf. Comput. Vis.}, 2010, pp. 322--335.

\bibitem{khan14_cvpr}
S.~H. Khan, M.~Bennamoun, F.~Sohel, and R.~Togneri, ``Automatic feature learning for robust shadow detection,'' in \emph{Proc. IEEE Conf. Comput. Vis. Pattern Recognit}, 2014, pp. 1939--1946.

\bibitem{shen16_cvpr}
L.~Shen, T.~W. Chua, and K.~Leman, ``Shadow optimization from structured deep edge detection,'' in \emph{Proc. IEEE Conf. Comput. Vis. Pattern Recognit}, 2015, pp. 2067--2074.

\bibitem{chen_cvpr2021}
Z.~Chen, L.~Wan, L.~Zhu, J.~Shen, H.~Fu, W.~Liu, and J.~Qin, ``Triple-cooperative video shadow detection,'' in \emph{Proc. IEEE Conf. Comput. Vis. Pattern Recognit}, 2021, pp. 2715--2724.

\bibitem{Lu_2022_CVPR}
X.~Lu, Y.~Cao, S.~Liu, C.~Long, Z.~Chen, X.~Zhou, Y.~Yang, and C.~Xiao, ``Video shadow detection via spatio-temporal interpolation consistency training,'' in \emph{Proc. IEEE Conf. Comput. Vis. Pattern Recognit}, 2022, pp. 3116--3125.

\bibitem{ding_eccv2022}
X.~Ding, J.~Yang, X.~Hu, and X.~Li, ``Learning shadow correspondence for video shadow detection,'' in \emph{Proc. Eur. Conf. Comput. Vis.}, 2022, pp. 705--722.

\bibitem{Wang_2020_CVPR}
T.~Wang, X.~Hu, Q.~Wang, P.-A. Heng, and C.-W. Fu, ``Instance shadow detection,'' in \emph{Proc. IEEE Conf. Comput. Vis. Pattern Recognit}, 2020, pp. 1880--1889.

\bibitem{wang_tpami2022}
T.~Wang, X.~Hu, P.-A. Heng, and C.-W. Fu, ``Instance shadow detection with a single-stage detector,'' \emph{IEEE Trans. Pattern Anal. Mach. Intell.}, pp. 1--14, 2022.

\bibitem{hu18_cvpr}
X.~Hu, L.~Zhu, C.-W. Fu, J.~Qin, and P.-A. Heng, ``Direction-aware spatial context features for shadow detection,'' in \emph{Proc. IEEE Conf. Comput. Vis. Pattern Recognit}, 2018, pp. 7454--7462.

\bibitem{hu18_tpami}
X.~Hu, C.-W. Fu, L.~Zhu, J.~Qin, and P.-A. Heng, ``Direction-aware spatial context features for shadow detection and removal,'' \emph{IEEE Trans. Pattern Anal. Mach. Intell.}, vol.~42, no.~11, pp. 2795--2808, 2020.

\bibitem{zhu18_eccv}
L.~Zhu, Z.~Deng, X.~Hu, C.-W. Fu, X.~Xu, J.~Qin, and P.-A. Heng, ``Bidirectional feature pyramid network with recurrent attention residual modules for shadow detection,'' in \emph{Proc. Eur. Conf. Comput. Vis.}, 2018, pp. 121--136.

\bibitem{zheng19_cvpr}
Q.~Zheng, X.~Qiao, Y.~Cao, and R.~W. Lau, ``Distraction-aware shadow detection,'' in \emph{Proc. IEEE Conf. Comput. Vis. Pattern Recognit}, 2019, pp. 5167--5176.

\bibitem{chen20_cvpr}
Z.~Chen, L.~Zhu, L.~Wan, S.~Wang, W.~Feng, and P.-A. Heng, ``A multi-task mean teacher for semi-supervised shadow detection,'' in \emph{Proc. IEEE Conf. Comput. Vis. Pattern Recognit}, 2020, pp. 5611--5620.

\bibitem{naoto20_tcsvt}
N.~Inoue and T.~Yamasaki, ``Learning from synthetic shadows for shadow detection and removal,'' \emph{IEEE Trans. Circuits Syst. Video Technol.}, vol.~31, no.~11, pp. 4187--4197, 2021.

\bibitem{dosovitskiy20_iclr}
A.~Dosovitskiy, L.~Beyer, A.~Kolesnikov, D.~Weissenborn, X.~Zhai, T.~Unterthiner, M.~Dehghani, M.~Minderer, G.~Heigold, S.~Gelly, J.~Uszkoreit, and N.~Houlsby, ``An image is worth 16x16 words: Transformers for image recognition at scale,'' in \emph{Proc. Int. Conf. Learn. Representations.}, 2021.

\bibitem{jie2022icassp}
L.~Jie and H.~Zhang, ``A fast and efficient network for single image shadow detection,'' in \emph{Proc. IEEE Int. Conf. Acoust. Speech Signal Process.}, 2022, pp. 2634--2638.

\bibitem{jie2022icme}
------, ``Rmlanet: Random multi-level attention network for shadow detection,'' in \emph{Proc. IEEE Int. Conf. Multimedia Expo.}, 2022, pp. 1--6.

\bibitem{jie2023rmlanet}
------, ``Rmlanet: Random multi-level attention network for shadow detection and removal,'' \emph{IEEE Trans. Circuits Syst. Video Technol.}, Early Access, 2023.

\bibitem{liao2021mm}
J.~Liao, Y.~Liu, G.~Xing, H.~Wei, J.~Chen, and S.~Xu, ``Shadow detection via predicting the confidence maps of shadow detection methods,'' in \emph{Proc. ACM Int. Conf. Multimedia.}, 2021, pp. 704--712.

\bibitem{kirillov2023segment}
A.~Kirillov, E.~Mintun, N.~Ravi, H.~Mao, C.~Rolland, L.~Gustafson, T.~Xiao, S.~Whitehead, A.~C. Berg, W.-Y. Lo, P.~Doll{\'a}r, and R.~Girshick, ``Segment anything,'' \emph{arXiv:2304.02643}, 2023.

\bibitem{tang2023can}
L.~Tang, H.~Xiao, and B.~Li, ``Can sam segment anything? when sam meets camouflaged object detection,'' \emph{arXiv:2304.04709}, 2023.

\bibitem{ji2023sam}
G.-P. Ji, D.-P. Fan, P.~Xu, M.-M. Cheng, B.~Zhou, and L.~Van~Gool, ``Sam struggles in concealed scenes--empirical study on" segment anything",'' \emph{arXiv:2304.06022}, 2023.

\bibitem{chen2023sam}
T.~Chen, L.~Zhu, C.~Ding, R.~Cao, S.~Zhang, Y.~Wang, Z.~Li, L.~Sun, P.~Mao, and Y.~Zang, ``Sam fails to segment anything?--sam-adapter: Adapting sam in underperformed scenes: Camouflage, shadow, and more,'' \emph{arXiv:2304.09148}, 2023.

\bibitem{hu2023sam}
C.~Hu and X.~Li, ``When sam meets medical images: An investigation of segment anything model (sam) on multi-phase liver tumor segmentation,'' \emph{arXiv:2304.08506}, 2023.

\bibitem{ma2023segment}
J.~Ma and B.~Wang, ``Segment anything in medical images,'' \emph{arXiv:2304.12306}, 2023.

\bibitem{cheng2023sam}
D.~Cheng, Z.~Qin, Z.~Jiang, S.~Zhang, Q.~Lao, and K.~Li, ``Sam on medical images: A comprehensive study on three prompt modes,'' \emph{arXiv:2305.00035}, 2023.

\bibitem{han2023segment}
D.~Han, C.~Zhang, Y.~Qiao, M.~Qamar, Y.~Jung, S.~Lee, S.-H. Bae, and C.~S. Hong, ``Segment anything model (sam) meets glass: Mirror and transparent objects cannot be easily detected,'' \emph{arXiv:2305.00278}, 2023.

\bibitem{zhang2023customized}
K.~Zhang and D.~Liu, ``Customized segment anything model for medical image segmentation,'' \emph{arXiv:2304.13785}, 2023.

\bibitem{dai2023samaug}
H.~Dai, C.~Ma, Z.~Liu, Y.~Li, P.~Shu, X.~Wei, L.~Zhao, Z.~Wu, D.~Zhu, W.~Liu \emph{et~al.}, ``Samaug: Point prompt augmentation for segment anything model,'' \emph{arXiv:2307.01187}, 2023.

\bibitem{wu2023medical}
J.~Wu, R.~Fu, H.~Fang, Y.~Liu, Z.~Wang, Y.~Xu, Y.~Jin, and T.~Arbel, ``Medical sam adapter: Adapting segment anything model for medical image segmentation,'' \emph{arXiv:2304.12620}, 2023.

\bibitem{shaharabany2023autosam}
T.~Shaharabany, A.~Dahan, R.~Giryes, and L.~Wolf, ``Autosam: Adapting sam to medical images by overloading the prompt encoder,'' \emph{arXiv:2306.06370}, 2023.

\bibitem{he2022masked}
K.~He, X.~Chen, S.~Xie, Y.~Li, P.~Doll{\'a}r, and R.~Girshick, ``Masked autoencoders are scalable vision learners,'' in \emph{Proc. IEEE Conf. Comput. Vis. Pattern Recognit}, 2022, pp. 16\,000--16\,009.

\bibitem{clip2021icml}
A.~Radford, J.~W. Kim, C.~Hallacy, A.~Ramesh, G.~Goh, S.~Agarwal, G.~Sastry, A.~Askell, P.~Mishkin, J.~Clark, G.~Krueger, and I.~Sutskever, ``Learning transferable visual models from natural language supervision,'' in \emph{Proc. Int. Conf. Mach. Learn.}, 2021, pp. 8748--8763.

\bibitem{kuznetsova2020open}
A.~Kuznetsova, H.~Rom, N.~Alldrin, J.~Uijlings, I.~Krasin, J.~Pont-Tuset, S.~Kamali, S.~Popov, M.~Malloci, A.~Kolesnikov \emph{et~al.}, ``The open images dataset v4: Unified image classification, object detection, and visual relationship detection at scale,'' \emph{Int. J. Comput. Vis.}, vol. 128, no.~7, pp. 1956--1981, 2020.

\bibitem{ba2016layer}
J.~L. Ba, J.~R. Kiros, and G.~E. Hinton, ``Layer normalization,'' \emph{arXiv preprint arXiv:1607.06450}, 2016.

\bibitem{hendrycks2016gaussian}
D.~Hendrycks and K.~Gimpel, ``Gaussian error linear units (gelus),'' \emph{arXiv preprint arXiv:1606.08415}, 2016.

\bibitem{pfeiffer2020adapterfusion}
J.~Pfeiffer, A.~Kamath, A.~R{\"u}ckl{\'e}, K.~Cho, and I.~Gurevych, ``Adapterfusion: Non-destructive task composition for transfer learning,'' \emph{arXiv preprint arXiv:2005.00247}, 2020.

\bibitem{tan19_icml}
M.~Tan and Q.~Le, ``Efficientnet: Rethinking model scaling for convolutional neural networks,'' in \emph{Proc. Int. Conf. Mach. Learn.}, 2019, pp. 6105--6114.

\bibitem{xie2017aggregated}
S.~Xie, R.~Girshick, P.~Doll{\'a}r, Z.~Tu, and K.~He, ``Aggregated residual transformations for deep neural networks,'' in \emph{Proceedings of the IEEE conference on computer vision and pattern recognition}, 2017, pp. 1492--1500.

\bibitem{lin2017focal}
T.-Y. Lin, P.~Goyal, R.~Girshick, K.~He, and P.~Doll{\'a}r, ``Focal loss for dense object detection,'' in \emph{Proc. IEEE Int. Conf. Comput. Vis.}, 2017, pp. 2980--2988.

\bibitem{guo11_cvpr}
R.~Guo, Q.~Dai, and D.~Hoiem, ``Single-image shadow detection and removal using paired regions,'' in \emph{Proc. IEEE Conf. Comput. Vis. Pattern Recognit}, 2011, pp. 2033--2040.

\bibitem{nguyen17_iccv}
V.~Nguyen, T.~F.~Y. Vicente, M.~Zhao, M.~Hoai, and D.~Samaras, ``Shadow detection with conditional generative adversarial networks,'' in \emph{Proc. IEEE Int. Conf. Comput. Vis.}, 2017, pp. 4510--4518.

\bibitem{hs18_iros}
S.~Hosseinzadeh, M.~Shakeri, and H.~Zhang, ``Fast shadow detection from a single image using a patched convolutional neural network,'' in \emph{Proc. IEEE Int. Conf. Intell. Rob. Syst.}, 2018, pp. 3124--3129.

\bibitem{le18_eccv}
H.~Le, T.~F.~Y. Vicente, V.~Nguyen, M.~Hoai, and D.~Samaras, ``{A+D Net}: Training a shadow detector with adversarial shadow attenuation,'' in \emph{Proc. Eur. Conf. Comput. Vis.}, 2018, pp. 662--678.

\bibitem{wang18_ijcai}
Y.~Wang, X.~Zhao, Y.~Li, X.~Hu, and K.~Huang, ``Densely cascaded shadow detection network via deeply supervised parallel fusion,'' in \emph{Proc. Int. Joint Conf. Artif. Intell.}, 2018, pp. 1007--1013.

\bibitem{Zhu_2021_ICCV}
L.~Zhu, K.~Xu, Z.~Ke, and R.~W. Lau, ``Mitigating intensity bias in shadow detection via feature decomposition and reweighting,'' in \emph{Proc. IEEE Int. Conf. Comput. Vis.}, 2021, pp. 4702--4711.

\bibitem{zhu_mm2022}
Y.~Zhu, X.~Fu, C.~Cao, X.~Wang, Q.~Sun, and Z.-J. Zha, ``Single image shadow detection via complementary mechanism,'' in \emph{Proc. ACM Int. Conf. Multimedia.}, 2022, pp. 6717--6726.

\bibitem{jose_wacv2023}
J.~M.~J. Valanarasu and V.~M. Patel, ``Fine-context shadow detection using shadow removal,'' in \emph{Proc. IEEE Winter Conf. Appl. Comput. Vis}, 2023, pp. 1705--1714.

\bibitem{cong2023sddnet}
R.~Cong, Y.~Guan, J.~Chen, W.~Zhang, Y.~Zhao, and S.~Kwong, ``Sddnet: Style-guided dual-layer disentanglement network for shadow detection,'' \emph{arXiv preprint arXiv:2308.08935}, 2023.

\bibitem{Sun_2023_ICCV}
J.~Sun, K.~Xu, Y.~Pang, L.~Zhang, H.~Lu, G.~Hancke, and R.~Lau, ``Adaptive illumination mapping for shadow detection in raw images,'' in \emph{Proc. IEEE Int. Conf. Comput. Vis.}, October 2023, pp. 12\,709--12\,718.

\bibitem{Yang_2023_ICCV}
H.~Yang, T.~Wang, X.~Hu, and C.-W. Fu, ``Silt: Shadow-aware iterative label tuning for learning to detect shadows from noisy labels,'' in \emph{Proc. IEEE Int. Conf. Comput. Vis.}, 2023, pp. 12\,687--12\,698.

\bibitem{Paszke_PyTorch_An_Imperative_2019}
A.~Paszke, S.~Gross, F.~Massa, A.~Lerer, J.~Bradbury, G.~Chanan, T.~Killeen, Z.~Lin, N.~Gimelshein, L.~Antiga, A.~Desmaison, A.~Kopf, E.~Yang, Z.~DeVito, M.~Raison, A.~Tejani, S.~Chilamkurthy, B.~Steiner, L.~Fang, J.~Bai, and S.~Chintala, ``{PyTorch: An Imperative Style, High-Performance Deep Learning Library},'' in \emph{Advances in Neural Information Processing Systems 32}, 2019, pp. 8024--8035.

\bibitem{zhang2023faster}
C.~Zhang, D.~Han, Y.~Qiao, J.~U. Kim, S.-H. Bae, S.~Lee, and C.~S. Hong, ``Faster segment anything: Towards lightweight sam for mobile applications,'' \emph{arXiv preprint arXiv:2306.14289}, 2023.

\end{thebibliography}

    
    




\end{document}